\documentclass{article}

\PassOptionsToPackage{numbers,compress}{natbib}
\usepackage[main,final]{neurips_2026}

\usepackage[utf8]{inputenc}
\usepackage[T1]{fontenc}
\usepackage{hyperref}
\usepackage{url}
\usepackage{booktabs}
\usepackage{amsfonts}
\usepackage{nicefrac}
\usepackage{microtype}
\usepackage{xcolor}

\usepackage{graphicx}
\usepackage{subcaption}
\usepackage{multirow}
\usepackage{enumitem}
\usepackage{tikz-cd}
\usepackage{tikz}
\usepackage{pgfplots}
\pgfplotsset{compat=1.18}
\usepackage{wrapfig}

\usepackage{amsmath}
\usepackage{amssymb}
\usepackage{mathtools}
\usepackage{amsthm}
\usepackage{scalerel}

\usepackage[capitalize,noabbrev]{cleveref}

\theoremstyle{plain}
\newtheorem{theorem}{Theorem}[section]

\newtheorem{lemma}[theorem]{Lemma}

\theoremstyle{definition}
\newtheorem{definition}[theorem]{Definition}

\theoremstyle{remark}
\newtheorem{remark}[theorem]{Remark}

\title{Learning Pseudo-Riemannian Manifolds for Heterophilic Graphs via Graph Signature}

\author{%
  Yun Young Choi\thanks{Equal contribution. Yun Young Choi, Asung Kil, and Sun Woo Park contributed equally.}%
  \kern0.45em\thanks{Corresponding author: \texttt{young@solverx.ai}.} \\
  SolverX \\
  Seoul, Republic of Korea \\
  \texttt{young@solverx.ai}
  \And
  Asung Kil\footnotemark[1] \\
  SolverX \\
  Seoul, Republic of Korea \\
  \texttt{askil@solverx.ai}
  \And
  Sun Woo Park\footnotemark[1] \\
  Max Planck Institute for Mathematics \\
  Bonn, Germany \\
  \texttt{s.park@mpim-bonn.mpg.de}
  \AND
  Minho Lee \\
  SolverX \\
  Seoul, Republic of Korea \\
  \texttt{mhlee@solverx.ai}
  \And
  Seokhwan Kim \\
  SolverX \\
  Seoul, Republic of Korea \\
  \texttt{kshsh0405@solverx.ai}
}

\begin{document}

\maketitle

\begin{abstract}
  Several widely used families of Graph Neural Networks carry a positive-definite geometric prior. Graph convolution read as heat diffusion needs a positive semi-definite operator, hyperbolic models place features on positive-definite Riemannian manifolds, and dot-product and GAT-style attention scores use positive-definite inner products. On heterophilic graphs, where edges often signify dissimilarity or structural repulsion, that prior is a mismatch: no positive-definite form can assign a negative squared distance to a direction the labels separate. To measure the mismatch, we introduce the \textit{Graph Signature Index} ($s_{\mathcal{G}}$), the inertia of a class-compatibility matrix built from label topology alone, and we prove that no value of the edge or node homophily index in $(0,1)$ rules out a negative eigenvalue.
  We then propose the \textit{Pseudo-Riemannian Attention Network} (PRAT), whose attention logit is a signed squared pseudo-distance under a learnable diagonal metric initialized from $s_{\mathcal{G}}$. The design is diagnose-then-learn: the signature supplies a sign structure that unconstrained training keeps outside near-zero directions, and the projection learns the rest of the metric around it.
  Across fifteen node-classification benchmarks PRAT outperforms or matches the strongest heterophily-adapted models. A controlled ablation ties that performance to the sign structure itself: with architecture and protocol fixed, forbidding the metric to be indefinite costs accuracy on the graphs $s_{\mathcal{G}}$ marks as indefinite and leaves the others unchanged.
\end{abstract}

\section{Introduction}
Graph Neural Networks (GNNs) are the standard tool for relational learning~\cite{7974879}.
Several of the most widely used families rest on a geometric prior rooted in Riemannian manifolds, typically Euclidean space, in which the metric tensor is strictly positive definite and connected nodes are therefore ``close'' and expected to share representations.
The prior is explicit in three places. Graph convolution read as heat diffusion requires a positive semi-definite operator~\cite{chamberlain2021grand}; hyperbolic architectures such as HGCN~\cite{chami2019hyperbolic} and HAT~\cite{zhang2021hyperbolic} embed features on positive-definite Riemannian manifolds; and dot-product~\cite{vaswani2017attention} and GAT-style~\cite{vel2018graph} attention use Euclidean inner products of projected features, which admit no negative squared distance.
The assumption is not universal. Signed attention~\cite{fagcn2021} and sheaf models~\cite{bodnar2022neural} are natural counterexamples, and precisely because they work, the geometry deserves to be treated as a modeling choice rather than a default.

While this Riemannian inductive bias serves well for homophilic graphs, it creates a fundamental \textit{geometric mismatch} when applied to heterophilic graphs, where edges often denote dissimilarity or structural repulsion~\cite{NEURIPS2020_58ae23d8, Pei2020Geom-GCN:}.
Heterophily-adapted GNNs address this architecturally with high-frequency spectral filters~\cite{gasteiger2022predictpropagategraphneural, chien2021adaptive, he2021bernnet} or signed aggregation weights~\cite{fagcn2021, yan2022, wang2025adaptiveriemanniangraphneural}.
These are empirically effective, but the space they act in stays positive definite, so repulsion is expressed by distorting an embedding built for attraction~\cite{topping2022understandingoversquashingbottlenecksgraphs}.

We read heterophily geometrically, whereas the prevailing view is statistical. Non-Euclidean GNNs~\cite{chami2019hyperbolic, Xiong2021Pseudo} adapt curvature to hierarchical structure. We argue that some heterophilic graphs call instead for an indefinite metric, which admits \textit{negative squared distances}~\cite{NEURIPS2020_123b7f02} and so can sharpen across one class pair while smoothing across another. We diagnose the geometry the labels induce and then learn the metric inside it. The diagnostic is the \textit{Graph Signature Index} ($s_{\mathcal{G}}$), a spectral invariant of the label topology. The model is the \textit{Pseudo-Riemannian Attention Network} (PRAT), whose attention logit is a signed squared pseudo-distance under a diagonal metric initialized from $s_{\mathcal{G}}$ and learned thereafter.

\section{Theoretical Framework: The Geometry of Heterophily}

\subsection{Graph Signature Index}
\label{graphsignatureindex}

Previous studies proposed measures of graph homophily, including the node homophily index $h_V$ \cite{Pei2020Geom-GCN:} and edge homophily index $h_E$ \cite{Zhu21} described in Appendix \ref{homophilyindex}. Their values, ranging between $0$ and $1$, quantify the likelihood that nodes with identical labels are connected by edges (see Appendix \ref{AppendixA:Refinement}). The graph signature index refines these indices by recording the signature of the eigenvalues of a real symmetric matrix constructed from a graph with node labels.

\begin{definition}
\label{def:graph_signature_matrix}
Let $\mathcal{G} := (\mathcal{V}, \mathcal{E})$ be an undirected graph without self-loops. Let $l: \mathcal{V} \to L = \{1,2, \cdots, |L|\}$ be the ground-truth node label function assigning each node to one of $|L|$ classes. The graph signature matrix $S_{G}$ is the $|L| \times |L|$ real symmetric matrix whose $(i,j)$-th entry is given as
\begin{align}
    s_{i,j} &:= \frac{|\{(v,w) \in \mathcal{E} \mid l(v) = i, l(w) = j \}|}{
        |\{v \in \mathcal{V} \mid l(v) = i\}| \cdot |\{v \in \mathcal{V} \mid l(v) = j\}|
    }
\end{align}
The graph signature index of $\mathcal{G}$, denoted as $s_{\mathcal{G}}$, is the 3-tuple consisting of the number of positive, zero, and negative eigenvalues of $S_{\mathcal{G}}$.
\end{definition}

Because $S_{\mathcal{G}}$ is a real symmetric matrix, every eigenvalue of $S_{\mathcal{G}}$ is real. Sylvester's law of inertia shows that the graph signature index is invariant under any invertible linear change of coordinates on $\mathbb{R}^{|L|}$.
\begin{theorem}[\cite{S52}]
\label{sylvester}
Let $S \in \mathbb{R}^{n \times n}$ be a real symmetric matrix. Then there exists a diagonal matrix $D \in \mathbb{R}^{n \times n}$ and an invertible matrix $W \in \mathbb{R}^{n \times n}$ such that $S = W^T D W$, where $( \cdot )^T$ denotes the transpose of a square matrix. The numbers of positive, zero, and negative eigenvalues of $S$ are invariant under the change of coordinates.
\end{theorem}

Each $(i,j)$-th entry of $S_{\mathcal{G}}$ is the expected probability that a node with a preassigned label $j$ lies in the depth $1$ unfolding tree rooted at a node with label $i$; see Appendix \ref{AppendixA:S_G-Interpretation}. The signature of the eigenvalues of $S_{\mathcal{G}}$ can be used to determine whether graphs with discrete labels are heterophilic or homophilic.
For example, graphs with two discrete node labels and $\mathcal{E} \neq \emptyset$ have three types of $s_{\mathcal{G}}$; see Appendix \ref{AppendixA:twolabel}. Appendix \ref{AppendixA:heuristic} provides heuristics on the density of these three graph types.

The following theorem relates the edge homophily level $h_E$ to the graph signature index $s_{\mathcal{G}}$.

\begin{theorem}[Critical Homophily Threshold]\label{theorem:critical_threshold}
    Assume class $i$ contains $n_i$ nodes and $N = \sum_{i=1}^{|L|} n_i$. Under a proportional allocation model, the expected graph signature matrix $S_{\mathcal{G}}$ has eigenvalues:
    \begin{align}
        \lambda_1 > 0, \hspace{15pt} \lambda_2 = \cdots = \lambda_{|L|} &= \frac{2|\mathcal{E}|\left(h_E N^2 - \sum_{k=1}^{|L|} n_k^2\right)}{\sum_{k=1}^{|L|} n_k^2 \cdot \left(N^2 - \sum_{k=1}^{|L|} n_k^2\right)}.
    \end{align}
    The critical homophily threshold is given by: $h_E^* = \frac{\sum_{k=1}^{|L|} n_k^2}{N^2}$. For balanced classes where $n_i = n$ for all $i$, this reduces to $h_E^* = 1/|L|$.
\end{theorem}

While Theorem \ref{theorem:critical_threshold} reveals a correlation between edge homophily and graph signature under idealized conditions, homophily indices cannot fully characterize the spectral properties of $S_{\mathcal{G}}$. We formalize this limitation by showing that no value of $h_E$ or $h_V$ in $(0,1)$ rules out a negative eigenvalue of $S_{\mathcal{G}}$.

\begin{theorem}[Limitation of Homophily Indices]\label{theorem:refinement}
    For every rational $h \in (0,1)$, there exists a graph $\mathcal{G} := (\mathcal{V}, \mathcal{E})$ with discrete node labels such that $h_E = h_V = h$ and $S_{\mathcal{G}}$ has a negative eigenvalue.
\end{theorem}

\subsection{Limitations of Riemannian GNNs}
\label{Riemannianlimit}

We reformulate the problem of analyzing heterophilic graphs
using pseudo-Riemannian manifolds. A pseudo-Riemannian manifold $(\mathcal{M},g)$ is a smooth manifold endowed with a metric tensor $g_p: T_p\mathcal{M} \times T_p\mathcal{M} \to \mathbb{R}$ which defines a non-degenerate symmetric bilinear form over the tangent space $T_p\mathcal{M}$ at each point $p \in \mathcal{M}$. A pseudo-Riemannian manifold is Riemannian if the tensor $g_p$ is positive-definite for every point $p \in \mathcal{M}$.
We keep this terminology for a symmetric bilinear form that is allowed to be degenerate. This is deliberate: the form on the hidden space is a pullback of rank at most $|L|$, hence degenerate whenever the hidden width exceeds the number of classes, and it depends on a hidden vector only through its image in $\mathcal{L}$. Non-degeneracy returns on $\mathcal{L}$ itself whenever $s_{\mathcal{G}}$ has no zero eigenvalue, and \cref{theorem:main_theorem0} counts only negative eigenvalues, so degeneracy does not affect it.
We focus on the $h$-dimensional manifold $\mathcal{M}$ endowed with a metric tensor $g_p := S \in \mathbb{R}^{h \times h}$ for some fixed real symmetric matrix $S$ at every point $p \in \mathcal{M}$, which defines the following quadratic form on tangent vectors $w_1, w_2 \in T_p\mathcal{M}$: $g_p(w_1,w_2) = w_1^T S w_2$~\citep{M09, Tu17}.

Let $\mathcal{G}:=(\mathcal{V}, \mathcal{E})$ be a graph with continuous node attributes $X:\mathcal{V} \to \mathbb{R}^{h}$ and embedded node labels $l:\mathcal{V} \to \mathbb{R}^{|L|}$.
We set the metric tensor over $\mathbb{R}^{|L|}$ to be the graph signature matrix $S_{\mathcal{G}}$ of the given graph $\mathcal{G}$. By doing so, we regard the graph $\mathcal{G}$ as a subspace of some smooth manifold $\mathcal{M}$ admitting a smooth function $f:\mathcal{M} \to \mathcal{L}\cong\mathbb{R}^{|L|}$, whose metric tensor is the pullback of metric tensor $f^* S_{\mathcal{G}}$. Given a metric tensor $g$ on $\mathcal{L}$, its pullback $f^* g$ on $\mathcal{M}$ is defined as
\begin{align*}
    (f^*g)_{p}(v,w) := g_{f(p)}(T_p f(v), T_p f(w)).
\end{align*}
Because $S_{\mathcal{G}}$ can have negative eigenvalues, $\mathcal{M}$ is a pseudo-Riemannian manifold, thereby covering a class of manifolds which are not Riemannian. The theorem below, proved in Appendix \ref{AppendixA:main_theorem0}, makes this observation explicit.

\begin{theorem}[Geometric Mismatch]\label{theorem:main_theorem0}
    Let $\mathcal{M}$ and $\mathcal{L}$ be differentiable manifolds and $f: \mathcal{M} \to \mathcal{L}$ be a smooth map. If $S_{\mathcal{G}}$ defined on $\mathcal{L}$ is indefinite and the pushforward $T_pf: T_p \mathcal{M} \to T_{f(p)}\mathcal{L}$ is surjective for every $p \in \mathcal{M}$, then there exists no diffeomorphism $\phi: \mathcal{L} \to \mathcal{L}$ such that the pullback of the Euclidean metric $I$, the identity form, satisfies $(\phi \circ f)^* I = f^* S_{\mathcal{G}}$.
\end{theorem}
\cref{theorem:main_theorem0} establishes a geometric mismatch between the Riemannian prior and graphs whose $S_{\mathcal{G}}$ is indefinite. It proves that this geometric mismatch is intrinsic and cannot be bridged by any smooth transformation of the Riemannian space.
Surjectivity of the pushforward is what carries the indefiniteness through to the pullback: the image of $T_pf$ is then all of $T_{f(p)}\mathcal{L}$, so a negative direction of $S_{\mathcal{G}}$ has a preimage. Injectivity alone would not suffice, since an injective map can pull an indefinite form back to a degenerate one. In our architecture $T_pf$ is the projection $\mathbf{W}$ of \cref{pullback_metric}, surjective exactly when $\mathrm{rank}\,\mathbf{W} = |L|$. Every configuration we report has $h' > |L|$, so the hypothesis holds for almost every $\mathbf{W}$.
Appendix \ref{AppendixA:main_theorem0} discusses further implications for heterophilic graphs.

\section{Method: Pseudo-Riemannian Attention Network (PRAT)}

\subsection{Pseudo-Riemannian Edge Attention using continuous features}

We define the tangent attention scores of a finite graph $\mathcal{G}$, which generalize the notion of attention scores used in graph attention networks (GAT) and attention-transformers.
\begin{figure}[t]
\centering
\begin{subfigure}{.40\textwidth}
\includegraphics[width=.95\linewidth]{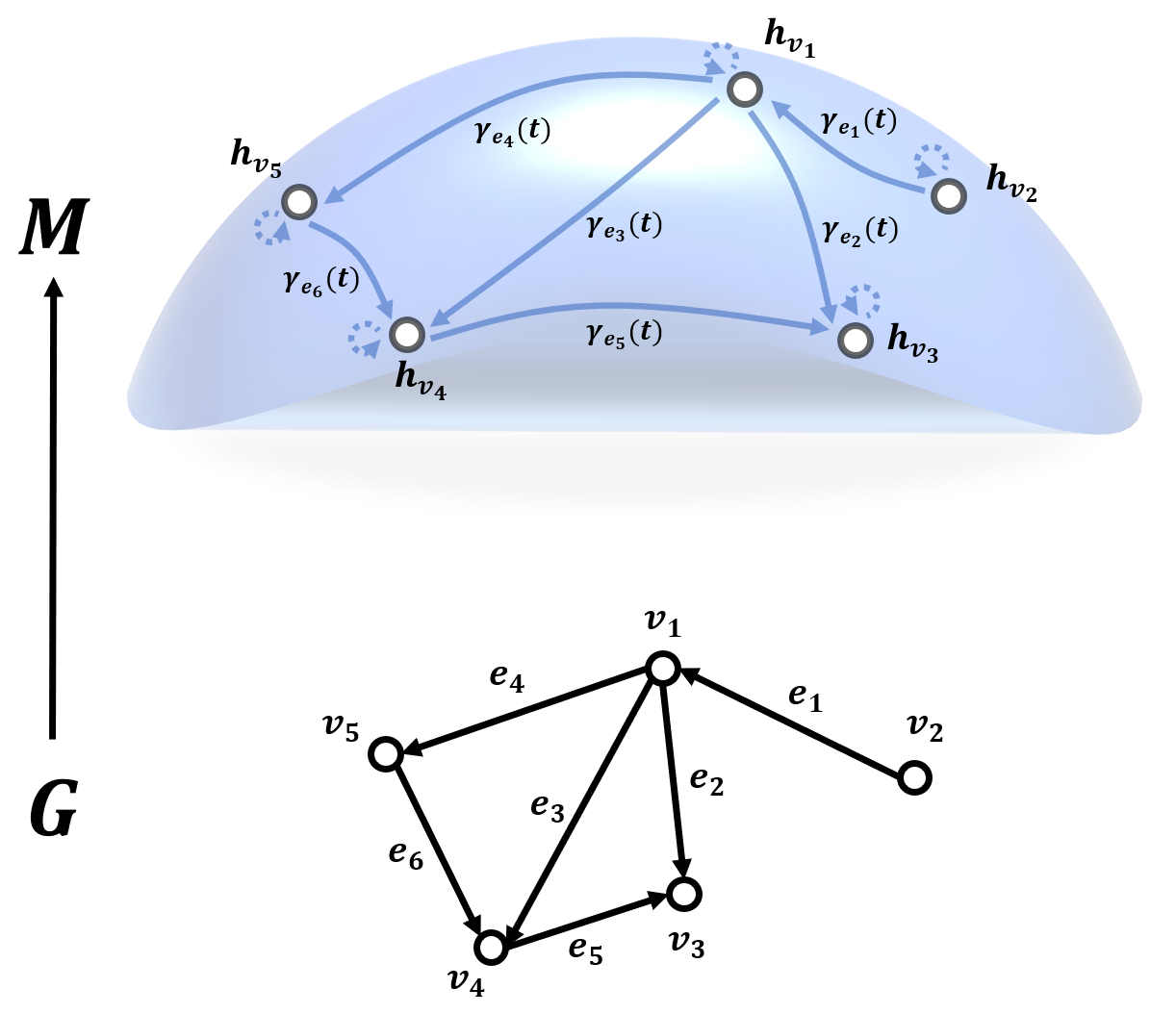}
\caption{}
\label{fig:manifold1}
\end{subfigure}
\begin{subfigure}{.55\textwidth}
\includegraphics[width=.95\linewidth]{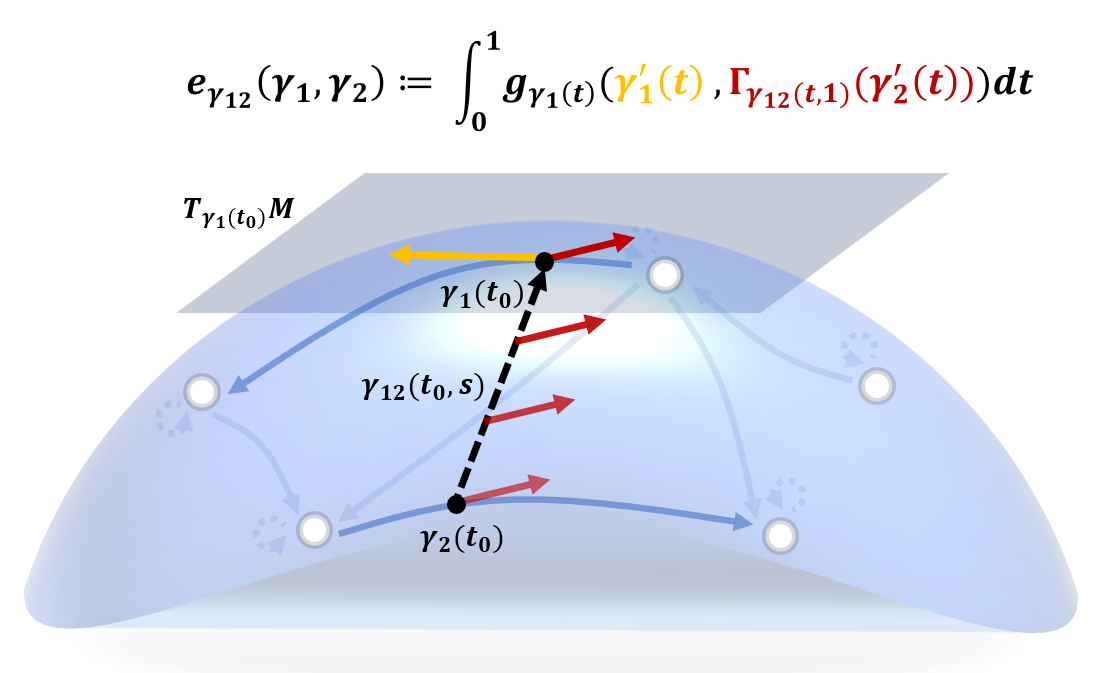}
\caption{}
\label{fig:manifold2}
\end{subfigure}
\caption{Figure 1(a) illustrates how one can map a directed graph $\mathcal{G}$ to a subspace of a pseudo-Riemannian manifold $(\mathcal{M},g)$. The nodes of $\mathcal{G}$ correspond to points on $\mathcal{M}$, whereas the edges of $\mathcal{G}$ correspond to curves on $\mathcal{M}$. Figure 1(b) demonstrates how one may compute tangent attention scores between two edges of $\mathcal{G}$ using the pseudo-Riemannian metric $g$ defined over the tangent space $T_{\gamma_1(t)}\mathcal{M}$ at each point of the curve representing an edge.}
\label{fig:manifold}
\end{figure}
The geometric structure endowed by the pseudo-metric tensor $g$ allows us to compute the numerical relation between two edges of a finite directed graph $\mathcal{G}$ by comparing their derivatives. Given a curve $\gamma:[0,1] \to \mathcal{M}$, the derivative of the curve at time $t = t_0$, denoted as $\gamma'(t_0)$, is an element of the tangent space $T_{\gamma(t_0)} \mathcal{M}$. Given two curves $\gamma_1, \gamma_2:[0,1] \to \mathcal{M}$, the two derivatives $\gamma_1'(t_0)$ and $\gamma_2'
(t_0)$ may not lie in the same tangent space. To overcome this, we define a function $\gamma_{12}: [0,1] \times [0,1] \to \mathcal{M}$ which satisfies $\gamma_{12}(t,0) = \gamma_2(t)$ and $\gamma_{12}(t,1) = \gamma_1(t)$. For each time $t = t_0$, the parallel transport operator $\Gamma_{\gamma_{12}(t_0,1)}: T_{\gamma_2(t_0)}\mathcal{M} \to T_{\gamma_1(t_0)}\mathcal{M}$ moves any tangent vector $w_2 \in T_{\gamma_2(t_0)}\mathcal{M}$ to the tangent vector $\Gamma_{\gamma_{12}(t_0,1)}(w_2) \in T_{\gamma_1(t_0)}\mathcal{M}$ along the curve $\gamma_{12}(t_0,s):\{t_0\} \times [0,1] \to \mathcal{M}$. After performing parallel transport,
we define the tangent attention score between $\gamma_1$ and $\gamma_2$.
Figure \ref{fig:manifold2} illustrates how to compute a tangent attention score.

\begin{definition} \label{def:tas_curve}
The tangent attention score between two curves $\gamma_1, \gamma_2:[0,1] \to \mathcal{M}$ with a choice of a function $\gamma_{12}:[0,1] \times [0,1] \to \mathcal{M}$ is
\begin{equation*}
    e_{\gamma_{12}}(\gamma_1, \gamma_2) := \int_{0}^1 g_{\gamma_1(t)} \left( \gamma_1'(t), \Gamma_{\gamma_{12}(t,1)}(\gamma_2'(t)) \right) dt.
\end{equation*}
\end{definition}

By imposing suitable topological structures on finite directed graphs, we can compute a tangent attention score between two edges of a directed graph $\mathcal{G} := (\mathcal{V}, \mathcal{E})$ (including self-loops) as a function of node features $X: \mathcal{V} \to \mathbb{R}^h$. See Appendix \ref{AppendixA:Topology} for details.
We also choose a continuous function $X_\gamma: \mathcal{G} \to \mathcal{M}$ which extends the node feature function $X$. Given two edges $e_1,e_2 \in \mathcal{E}$ with the corresponding paths $\gamma_{e_1},\gamma_{e_2}:[0,1] \to \mathbb{R}^h$, we define $\gamma_{e_1,e_2}(t,s) := s \cdot \gamma_{e_1}(t) + (1-s) \cdot \gamma_{e_2}(t)$.
Tangent attention scores are symmetric in their two edges. \cref{AppendixA:Topology} gives the argument.
We can quantify the differences among node features of $\mathcal{G}$ using these scores.
\begin{definition} \label{def:tas_graph}
The tangent attention score matrix of $\mathcal{G}$, denoted as $e_{\mathcal{G},\gamma}$, is the $|\mathcal{E}| \times |\mathcal{E}|$ real symmetric matrix whose $(e_1,e_2)$-th entry for $e_1, e_2 \in \mathcal{E}$ is the tangent attention score $e_{\gamma_{e_1,e_2}}(\gamma_{e_1}, \gamma_{e_2})$.
\end{definition}

Using $e_{\mathcal{G},\gamma}$, we recover several classes of previously studied attention functions. The detailed version of the theorem stated below can be found in Appendix \ref{AppendixA:main_theorem}.

\begin{theorem} \label{theorem:main_theorem}
    There exist continuous extensions $X_\gamma: \mathcal{G} \to \mathcal{M}$ such that $e_{\mathcal{G},\gamma}$ respectively recovers dot-product attention, single-layer feedforward linear layer, and dynamic graph attention~\citep{vaswani2017attention, vel2018graph, brody2022how}.
\end{theorem}

\subsection{Learnable Indefinite Metric}
Because tangent attention scores encompass various types of previously studied attention scores, a wide class of novel graph attention networks can be constructed by carefully choosing the curves representing the edges of $\mathcal{G}$, choosing the symmetric matrix $S$, and choosing which subsets of entries of the tangent attention score matrix to use.

Building on this, our proposed \textit{Pseudo-Riemannian Attention Network} (PRAT) architecture reformulates classical attention networks with tangent attention scores defined from the curve $\gamma_e^{[1]}:[0,1] \to \mathcal{M}$ as shown in Theorem \ref{theorem:main_theorem} and Appendix \ref{AppendixA:main_theorem}. The following equation governs the PRAT algorithm.
\begin{equation} \label{equation:PRAT}
\begin{split}
    h^{(i+1)}_v := \bigoplus_{k=1}^K\sigma \left( \sum_{w \in \mathcal{N}_v \cup \{v\}} \alpha_{v,w}^{[i,k]} \mathbf{W_v}^{[i,k]} h^{(i)}_w \right), \hspace{15pt} h^{(0)}_v := \psi_{enc}(X(v))
\end{split}
\end{equation}
For each $1 \leq k \leq K$,
\begin{equation*} \label{equation:PRATattention}
\begin{split}
    \alpha_{v,w}^{[i,k]} := \frac{\exp(e(h^{(i)}_v,h^{(i)}_w)^{[k]})}{\sum_{u \in \mathcal{N}_v \cup \{v\}} \exp(e(h^{(i)}_v,h^{(i)}_u)^{[k]})}, \;
    e(h^{(i)}_v,h^{(i)}_u)^{[k]} := \begin{cases}
    \Delta(h_{v,u}^{(i)})^\top \mathbf{D}^{[i,k]} \Delta(h_{v,u}^{(i)}) & \text{if}\,\substack{u \in \mathcal{N}_v} \\
    a^{[i,k]} (h^{(i)}_v)^\top \mathbf{D}^{[i,k]} (h^{(i)}_v) & \text{if}\,\substack{v = u}
    \end{cases}
\end{split}
\end{equation*}
where $\mathcal{N}_v$ is the set of nodes adjacent to the node $v$, $\Delta(h_{v,u}^{(i)})$ is equal to $h_v^{(i)} - h_u^{(i)}$ given $u \in \mathcal{N}_v$, and $\sigma$ is the nonlinear activation function. $\psi_{enc}:\mathbb{R}^{h}\rightarrow\mathbb{R}^{h'}$ is a learnable function (e.g., multilayer perceptron) that maps node features to hidden states of dimension $h'$. The $a^{[i,k]}$ are learnable self-loop weights playing the role of $a_v^2$ in \cref{eqn:gamma_result1}, calibrated in Appendix \ref{selfloopscaling}, and $\mathbf{D}^{[i,k]}\in\mathbb{R}^{h'\times h'}$ are symmetric matrices that serve as the metric tensor.

In Appendix \ref{PRATframework}, we conceptualize the learning process of PRAT as a sequence of geometric transformations, from which we derive $\mathbf{D}^{[l,k]}$ for the $k$-th head at layer $l$ as follows, where $\mathbf{W}^{[l,k]}\in\mathbb{R}^{|L|\times h'}$ is a learnable weight matrix, and $S^{|L|}\in\mathbb{R}^{|L|\times|L|}$ is a pseudo-metric on the latent manifold.
\begin{align}
    \mathbf{D}^{[l,k]} = (\mathbf{W}^{[l,k]})^\top S^{|L|} \mathbf{W}^{[l,k]}.
\label{pullback_metric}
\end{align}
Following the theoretical motivation in \cref{Riemannianlimit}, we instantiate the pseudo-metric $S^{|L|}$ as the graph signature matrix $S_{\mathcal{G}}$.
The pullback then carries the label topology into the attention score. Substituting $S^{|L|} = S_{\mathcal{G}}$ into \cref{pullback_metric} and performing the eigendecomposition $S_{\mathcal{G}} = Q^\top\Lambda_{\mathcal{G}} Q$, we can reformulate the induced metric tensor. By absorbing the orthogonal eigenvector matrix $Q$ into the learnable weight matrix, i.e., defining $\mathbf{W}'^{[l,k]} := Q \mathbf{W}^{[l,k]}$, we arrive at:
\begin{align}
    \mathbf{D}^{[l,k]} &= (\mathbf{W}^{[l,k]})^\top (Q^\top \Lambda_{\mathcal{G}} Q) \mathbf{W}^{[l,k]} = (Q \mathbf{W}^{[l,k]})^\top \Lambda_{\mathcal{G}} (Q \mathbf{W}^{[l,k]}) = (\mathbf{W}'^{[l,k]})^\top \Lambda_{\mathcal{G}} \mathbf{W}'^{[l,k]}. \label{eq:factorized_metric}
\end{align}
In \cref{eq:factorized_metric}, $\Lambda_{\mathcal{G}}$ is a diagonalized graph signature matrix which directly encodes the homophilic or heterophilic character of the graph. In practical semi-supervised settings, however, the ground-truth labels for the entire graph are inaccessible, so we compute the empirical graph signature $S_{\mathcal{G},\text{train}}$ and its eigenvalues $\Lambda_{\mathcal{G},\text{train}}$ on the subgraph induced by the training nodes. We initialize the diagonal matrix $\Lambda$ using a normalized version of $\Lambda_{\mathcal{G},\text{train}}$, i.e. $\Lambda_{\text{init}} = \frac{\Lambda_{\mathcal{G},\text{train}}}{\text{std}(\Lambda_{\mathcal{G},\text{train}})} \cdot \frac{1}{\sqrt{|L|}}$, where $\text{std}(\cdot)$ is the standard deviation, set to $1$ if zero. We leave these values free and treat the diagonal matrix $\Lambda$ as a learnable parameter. This design allows the model to fine-tune the geometric prior derived from the training set and generalize it to the unseen test nodes.
Labels enter only here, and the forward pass never reads one. What the initialization has to supply is an informed indefinite prior, not ground truth: \cref{app:labelfree} shows that a signature estimated from $k$-means pseudo-labels on node features recovers the label-based accuracy on both indefinite benchmarks, and \cref{app:sparse} shows how far the estimate and the accuracy survive when the training labels are few or wrong.

\section{Experiments}
\subsection{Datasets and Baselines}

\paragraph{Standard Benchmark Datasets.} We first evaluate PRAT on node classification across eight standard benchmarks covering a broad spectrum of homophily levels. We use Texas and Cornell from WebKB \footnote{\url{http://www.cs.cmu.edu/afs/cs.cmu.edu/project/theo-11/www/wwkb}}, Chameleon and Squirrel \cite{rozemberczki2021multi}, and Actor \cite{tang2009social} as heterophilic datasets, and Cora, Citeseer, and PubMed \cite{sen2008collective, yang2016revisiting} as homophilic counterparts. Detailed statistics, including edge homophily levels ($h_E$) and graph signature indices ($s_{\mathcal{G}}$), are summarized in \cref{benchmark_stats}. The homophilic graphs here carry strictly positive signatures and the heterophilic ones other than Actor carry negative components. That association only motivates the design; the ablation in \cref{ablation_result}, not this correlation, shows that the sign structure produces the gain.

\paragraph{Platonov Benchmark Datasets.} We additionally evaluate PRAT on seven datasets curated by~\citet{platonov2023critical}: (i) the \textit{filtered} versions of Chameleon and Squirrel, in which duplicate nodes causing severe train--test leakage in the original benchmarks are removed; and (ii) five large-scale heterophilic datasets, \textit{roman-empire}, \textit{amazon-ratings}, \textit{minesweeper}, \textit{tolokers} and \textit{questions}, spanning 10K to 49K nodes % and a wider range of edge homophily ($h_E \in [0.05,\,0.84]$).
They test PRAT at scale without the data-quality artifacts of the original Wikipedia networks. Statistics are in \cref{benchmark_stats_platonov} and descriptions in Appendix \ref{benchmarkdatasets}.

\begin{table}[t]
\centering
\setlength{\tabcolsep}{2pt}
\scriptsize{}
\begin{minipage}[t]{0.49\textwidth}
\centering
\caption{Standard benchmark statistics ($h_E$, $s_{\mathcal{G}}$).}
\label{benchmark_stats}
\resizebox{\linewidth}{!}{%
\begin{tabular}{ccccccccc}
\toprule
\textbf{Dataset} & {Cora} & {PubMed} & {Citeseer} & {Chameleon} & {Squirrel} & {Actor} & {Cornell} & {Texas} \\
\midrule
Nodes   & 2,708 & 19,717 & 3,327 & 2,277 & 5,201 & 7,600 & 183 & 183 \\
Edges   & 5,278 & 44,324 & 4,552 & 36,101 & 217,073 & 30,019 & 298 & 325 \\
Classes & 7 & 3 & 6 & 5 & 5 & 5 & 5 & 5 \\
\midrule
$h_E$ & 0.810 & 0.802 & 0.736 & 0.230 & 0.222 & 0.217 & 0.123 & 0.061 \\
$s_{\mathcal{G}}$ & (7,0,0) & (3,0,0) & (6,0,0) & (3,0,2) & (2,2,1) & (1,4,0) & (3,0,2) & (2,0,3) \\
\bottomrule
\end{tabular}}
\end{minipage}\hfill
\begin{minipage}[t]{0.49\textwidth}
\centering
\caption{Platonov benchmark statistics~\citep{platonov2023critical}.}
\label{benchmark_stats_platonov}
\resizebox{\linewidth}{!}{%
\begin{tabular}{cccccccc}
\toprule
\textbf{Dataset} & {Cham.-filt.} & {Sq.-filt.} & {roman-empire} & {amazon-ratings} & {minesweeper} & {tolokers} & {questions} \\
\midrule
Nodes    & 890   & 2,223  & 22,662 & 24,492 & 10,000 & 11,758 & 48,921 \\
Edges    & 8,854 & 57,134 & 32,927 & 93,050 & 39,402 & 519,000 & 153,540 \\
Classes  & 5     & 5      & 18     & 5      & 2      & 2      & 2 \\
\midrule
$h_E$    & 0.235 & 0.220  & 0.046  & 0.380  & 0.683  & 0.595  & 0.840 \\
$s_{\mathcal{G}}$ & (3,0,2) & (2,1,2) & (6,7,5) & (3,2,0) & (1,1,0) & (2,0,0) & (1,1,0) \\
\bottomrule
\end{tabular}}
\end{minipage}
\end{table}

\paragraph{Baselines on Standard Benchmarks.} We compare PRAT against a diverse set of architectures spanning multiple paradigms:
(1) \textbf{Standard GNNs}: GCN \cite{kipf2017semisupervised}, GIN \cite{xu2018how}, and GraphSAGE \cite{hamilton2017inductive};
(2) \textbf{Attention-based models}: GAT \cite{vel2018graph}, GATv2 \cite{brody2022how}, Scaled Dot-Product (SDP) Attention \cite{vaswani2017attention, shi2021masked}, and Graph Transformer with and without ego-neighbor separation (GT, GT-sep) \cite{shi2021masked, NEURIPS2020_58ae23d8};
(3) \textbf{Heterophily-adapted GNNs}: CPGNN \cite{Zhu21}, FAGCN \cite{fagcn2021}, Geom-GCN \cite{Pei2020Geom-GCN:}, GPR-GNN \cite{chien2021adaptive}, BernNet \cite{he2021bernnet}, and H$_2$GCN \cite{NEURIPS2020_58ae23d8};
(4) \textbf{Geometric models}: Q-GCN \cite{Xiong2021Pseudo} on a pseudo-hyperboloid and HAT \cite{zhang2021hyperbolic} on the Poincar\'e ball.

\paragraph{Baselines on Platonov Benchmarks.} On the Platonov benchmarks, we compare PRAT against the best of three ResNet variants and the 14 GNN baselines reported in~\citet{platonov2023critical}.
(1) \textbf{Standard GNNs}: GCN~\cite{kipf2017semisupervised} and GraphSAGE~\cite{hamilton2017inductive};
(2) \textbf{Attention-based models}: GAT~\cite{vel2018graph}, GAT-sep~\cite{NEURIPS2020_58ae23d8}, Graph Transformer (GT)~\cite{shi2021masked}, and GT-sep~\cite{NEURIPS2020_58ae23d8}, where the ``-sep'' suffix denotes the ego- and neighbor-embedding separation trick of~\citet{NEURIPS2020_58ae23d8};
(3) \textbf{Heterophily-adapted GNNs}: H$_2$GCN~\cite{NEURIPS2020_58ae23d8}, CPGNN~\cite{Zhu21}, GPR-GNN~\cite{chien2021adaptive}, FSGNN~\cite{maurya2022simplifying}, GloGNN~\cite{li2022finding}, FAGCN~\cite{fagcn2021}, GBK-GNN~\cite{du2022gbk}, and JacobiConv~\cite{wang2022how}.

\subsection{Experimental Setup}

\paragraph{Standard Benchmarks.} We evaluate performance using 10 splits. For heterophilic datasets, we adopt the fixed 48\%/32\%/20\% train/validation/test splits provided by \citet{Pei2020Geom-GCN:}. For homophilic datasets, we generate random 60\%/20\%/20\% splits, avoiding the selection bias of a fixed set. All graphs are undirected with self-loops added, and the seed is fixed across splits. Baselines and PRAT share one sweep over a predefined range (hidden dimension, learning rate, dropout), and we report mean test accuracy with standard deviation at the best-validation step.\footnote{For Geom-GCN, we report results from~\citet{yan2022}.} Detailed configurations are provided in Appendix \ref{hyperparameters}.

\paragraph{Platonov Benchmarks.} For the seven Platonov benchmarks, we use the official training pipeline of~\citet{platonov2023critical}\footnote{\url{https://github.com/yandex-research/heterophilous-graphs}} as our experimental harness, where PRAT is implemented as a drop-in attention replacement inside the Platonov backbone.
% (input linear $\rightarrow$ residual blocks $\rightarrow$ output linear, with LayerNorm and dropout).
Data loading, the ten predefined splits, validation-based model selection and 10-split mean $\pm$ std reporting all follow~\citet{platonov2023critical} exactly, so every deviation we report is across splits at a fixed seed. \cref{app:seeds} adds three seeds and finds seed variance the smaller of the two.
PRAT is hyperparameter-tuned inside the same harness published by~\citet{platonov2023critical}.
% This guarantees that the reported numbers are directly comparable to their Table~4 baselines, which we quote verbatim. Following the protocol used for our filtered Chameleon/Squirrel evaluation, only PRAT is hyperparameter-tuned (via a sweep over the search range in \cref{tab:hyperparams_platonov}); baseline numbers are kept as published.
Detailed configurations are provided in Appendix \ref{hyperparameters}.

\paragraph{Cost.} A PRAT layer has GAT's asymptotic cost, since the metric it adds is diagonal and acts on a latent space whose dimension is the number of classes (\cref{appendix:complexity} gives the expressions). Measured on the two largest graphs at equal backbone, width and depth, a training step takes $88.7$\,ms on questions against $91.2$ for GAT and $175.2$ for GT, and $26.7$\,ms on tolokers against $22.8$ and $32.9$ (\cref{app:cost}). At this scale, PRAT costs as much as GAT.

\subsection{Main Results}

\begin{table}[t]
    \caption{Performance of PRAT and baselines on benchmark datasets. The top 3 performances are highlighted in \textbf{First}, \textcolor{blue}{Second}, and \textcolor{red}{Third}.}
	\centering
    \scriptsize
	\setlength{\tabcolsep}{2pt}
	\begin{tabular}{@{}lccc@{\hspace{6pt}}ccccc@{}}
		\toprule
		{} & \multicolumn{3}{c}{\textbf{Homophilic}} & \multicolumn{5}{c}{\textbf{Heterophilic}} \\
		\cmidrule(lr){2-4}\cmidrule(lr){5-9}
		{} & {Cora} & {PubMed} & {Citeseer} & {Chameleon} & {Squirrel} & {Actor} & {Cornell} & {Texas} \\
		$h_E$ & 0.810 & 0.802 & 0.736 & 0.230 & 0.222 & 0.217 & 0.123 & 0.061 \\ \midrule
		GCN & 88.32\scriptsize{$\pm$1.02} & 88.75\scriptsize{$\pm$0.43} & 77.69\scriptsize{$\pm$1.06} & 67.43\scriptsize{$\pm$2.45} & 54.95\scriptsize{$\pm$1.19} & 30.56\scriptsize{$\pm$0.70} & 44.32\scriptsize{$\pm$6.75} & 58.38\scriptsize{$\pm$4.86} \\
		GIN & 87.38\scriptsize{$\pm$1.56} & 89.20\scriptsize{$\pm$0.51} & 76.82\scriptsize{$\pm$1.29} & 62.52\scriptsize{$\pm$2.61} & 53.56\scriptsize{$\pm$1.66} & 30.97\scriptsize{$\pm$1.56} & 41.62\scriptsize{$\pm$6.86} & 59.73\scriptsize{$\pm$3.51} \\
		SAGE & 88.80\scriptsize{$\pm$1.04} & 89.77\scriptsize{$\pm$0.45} & 77.66\scriptsize{$\pm$0.81} & 64.91\scriptsize{$\pm$1.81} & 43.92\scriptsize{$\pm$1.87} & \textcolor{red}{36.17\scriptsize{$\pm$1.26}} & 69.46\scriptsize{$\pm$8.64} & 81.35\scriptsize{$\pm$3.72} \\
		\midrule
		GAT & 88.32\scriptsize{$\pm$1.09} & 88.87\scriptsize{$\pm$0.41} & 77.10\scriptsize{$\pm$1.27} & \textcolor{blue}{69.67\scriptsize{$\pm$1.67}} & 55.18\scriptsize{$\pm$2.13} & 31.61\scriptsize{$\pm$1.99} & 50.54\scriptsize{$\pm$7.55} & 64.59\scriptsize{$\pm$6.67} \\
		GATv2 & 87.99\scriptsize{$\pm$0.93} & 89.63\scriptsize{$\pm$0.47} & 76.97\scriptsize{$\pm$0.96} & 67.94\scriptsize{$\pm$2.34} & \textbf{59.79\scriptsize{$\pm$1.83}} & 32.97\scriptsize{$\pm$1.37} & 54.32\scriptsize{$\pm$9.70} & 66.76\scriptsize{$\pm$6.63} \\
		SDP & 87.99\scriptsize{$\pm$1.30} & 89.45\scriptsize{$\pm$0.55} & 76.85\scriptsize{$\pm$0.97} & 67.06\scriptsize{$\pm$1.53} & 56.42\scriptsize{$\pm$3.30} & 34.19\scriptsize{$\pm$0.82} & 52.97\scriptsize{$\pm$8.04} & 65.68\scriptsize{$\pm$5.28} \\
		GT & 87.64\scriptsize{$\pm$0.94} & 89.10\scriptsize{$\pm$0.50} & 76.76\scriptsize{$\pm$1.31} & 68.71\scriptsize{$\pm$2.35} & \textcolor{red}{56.83\scriptsize{$\pm$1.40}} & 35.70\scriptsize{$\pm$0.97} & 66.22\scriptsize{$\pm$4.81} & 76.76\scriptsize{$\pm$6.14} \\
		GT-sep & 87.92\scriptsize{$\pm$1.31} & 89.13\scriptsize{$\pm$0.45} & 77.69\scriptsize{$\pm$1.27} & 60.59\scriptsize{$\pm$1.78} & 44.07\scriptsize{$\pm$1.29} & 35.86\scriptsize{$\pm$0.67} & 70.00\scriptsize{$\pm$2.97} & 77.03\scriptsize{$\pm$5.59} \\
		\midrule
		CPGNN & 87.24\scriptsize{$\pm$1.07} & 89.78\scriptsize{$\pm$0.48} & 76.83\scriptsize{$\pm$1.00} & 56.18\scriptsize{$\pm$2.93} & 42.34\scriptsize{$\pm$1.07} & 33.95\scriptsize{$\pm$1.08} & 64.05\scriptsize{$\pm$6.63} & 77.57\scriptsize{$\pm$7.55} \\
		FAGCN & 88.75\scriptsize{$\pm$1.38} & 89.03\scriptsize{$\pm$0.55} & 77.46\scriptsize{$\pm$1.11} & 65.07\scriptsize{$\pm$1.56} & 49.55\scriptsize{$\pm$1.92} & \textcolor{blue}{36.51\scriptsize{$\pm$1.02}} & 60.27\scriptsize{$\pm$6.40} & 71.89\scriptsize{$\pm$7.17} \\
		Geom-GCN & 85.35\scriptsize{$\pm$1.57} & 89.95\scriptsize{$\pm$0.47} & \textcolor{blue}{78.02\scriptsize{$\pm$1.15}} & 60.00\scriptsize{$\pm$2.81} & 38.15\scriptsize{$\pm$0.92} & 31.59\scriptsize{$\pm$1.15} & 60.54\scriptsize{$\pm$3.67} & 66.76\scriptsize{$\pm$2.72} \\
		GPR-GNN & \textcolor{red}{89.43\scriptsize{$\pm$1.06}} & 90.01\scriptsize{$\pm$0.60} & 77.76\scriptsize{$\pm$0.83} & 68.18\scriptsize{$\pm$1.92} & 51.46\scriptsize{$\pm$1.34} & 35.38\scriptsize{$\pm$0.98} & \textcolor{red}{74.59\scriptsize{$\pm$4.12}} & \textcolor{red}{83.51\scriptsize{$\pm$5.78}} \\
		BernNet & \textbf{89.67\scriptsize{$\pm$1.04}} & \textcolor{blue}{90.13\scriptsize{$\pm$0.51}} & 77.79\scriptsize{$\pm$1.38} & \textbf{70.77\scriptsize{$\pm$1.63}} & 55.38\scriptsize{$\pm$1.46} & 35.20\scriptsize{$\pm$1.14} & \textbf{76.76\scriptsize{$\pm$3.68}} & \textcolor{blue}{83.78\scriptsize{$\pm$4.27}} \\
		H$_2$GCN & \textcolor{blue}{89.61\scriptsize{$\pm$1.15}} & \textbf{90.14\scriptsize{$\pm$0.51}} & 76.86\scriptsize{$\pm$1.35} & 59.30\scriptsize{$\pm$2.12} & 35.55\scriptsize{$\pm$1.58} & 34.72\scriptsize{$\pm$1.18} & \textcolor{blue}{74.86\scriptsize{$\pm$4.68}} & 82.43\scriptsize{$\pm$5.28} \\
		\midrule
		Q-GCN & 87.70\scriptsize{$\pm$1.19} & 88.22\scriptsize{$\pm$0.40} & 77.07\scriptsize{$\pm$1.50} & 59.67\scriptsize{$\pm$2.09} & 46.27\scriptsize{$\pm$1.22} & 29.65\scriptsize{$\pm$0.70} & 45.68\scriptsize{$\pm$4.90} & 61.62\scriptsize{$\pm$5.10} \\
		HAT & 88.36\scriptsize{$\pm$1.09} & 89.89\scriptsize{$\pm$0.53} & \textcolor{red}{77.90\scriptsize{$\pm$1.35}} & 57.48\scriptsize{$\pm$2.03} & 38.87\scriptsize{$\pm$1.40} & 34.34\scriptsize{$\pm$1.03} & 66.49\scriptsize{$\pm$4.39} & 78.65\scriptsize{$\pm$5.05} \\
		\midrule
		\textbf{PRAT (ours)} & 88.42\scriptsize{$\pm$1.17} & \textcolor{red}{90.03\scriptsize{$\pm$0.47}} & \textbf{78.06\scriptsize{$\pm$1.37}} & \textcolor{red}{68.73\scriptsize{$\pm$1.74}} & \textcolor{blue}{56.96\scriptsize{$\pm$1.51}} & \textbf{37.26\scriptsize{$\pm$1.44}} & 71.89\scriptsize{$\pm$4.71} & \textbf{84.32\scriptsize{$\pm$5.10}} \\
		\bottomrule
	\end{tabular}
	\label{benchmark_results}
\end{table}

\cref{benchmark_results} reports node classification accuracy on the eight standard benchmarks. PRAT is best on Actor and Texas, second on Squirrel, third on Chameleon and fourth on Cornell, and among the homophilic datasets it is best on Citeseer and third on PubMed; \cref{app:significance} reports the pairwise tests, four of which are ties. PRAT attends over \emph{1-hop} neighborhoods only, yet it exceeds the multi-hop spectral filters BernNet and GPR-GNN on Texas, Squirrel and Actor, so the margin there does not come from a larger receptive field.
\cref{app:attention_geometric} reads the GT and geometric rows.

\begin{table}[t]
    \caption{Performance on \citet{platonov2023critical} benchmarks. Accuracy ($\uparrow$) is reported for roman-empire, amazon-ratings, Cham.-filt., Sq.-filt.; ROC AUC ($\uparrow$) for minesweeper, tolokers, questions. $^{\dagger}$Best per dataset of the graph-agnostic ResNet of~\citet{platonov2023critical} and its graph-aware variants ResNet+SGC and ResNet+adj.}
    \centering
    \scriptsize
    \setlength{\tabcolsep}{2pt}
    \begin{tabular}{lcc@{\hspace{6pt}}ccccc}
        \toprule
        & \multicolumn{2}{c}{\textbf{Filtered Wikipedia}} & \multicolumn{5}{c}{\textbf{Large-scale heterophilic}} \\
        \cmidrule(lr){2-3}\cmidrule(lr){4-8}
         & Cham.-filt. & Sq.-filt. & roman-empire & amazon-ratings & minesweeper & tolokers & questions \\
        $h_E$ & 0.235 & 0.220 & 0.046 & 0.380 & 0.683 & 0.595 & 0.840 \\
        \midrule
        ResNet$^{\dagger}$ & \textcolor{red}{41.01\scriptsize{$\pm$4.54}} & 38.37\scriptsize{$\pm$1.99} & 73.90\scriptsize{$\pm$0.51} & 51.83\scriptsize{$\pm$0.57} & 70.88\scriptsize{$\pm$0.90} & 80.70\scriptsize{$\pm$0.97} & 75.81\scriptsize{$\pm$0.96} \\
        \midrule
        GCN & 40.89\scriptsize{$\pm$4.12} & \textcolor{red}{39.47\scriptsize{$\pm$1.47}} & 73.69\scriptsize{$\pm$0.74} & 48.70\scriptsize{$\pm$0.63} & 89.75\scriptsize{$\pm$0.52} & 83.64\scriptsize{$\pm$0.67} & 76.09\scriptsize{$\pm$1.27} \\
        SAGE & 37.77\scriptsize{$\pm$4.14} & 36.09\scriptsize{$\pm$1.99} & 85.74\scriptsize{$\pm$0.67} & \textcolor{blue}{53.63\scriptsize{$\pm$0.39}} & \textcolor{blue}{93.51\scriptsize{$\pm$0.57}} & 82.43\scriptsize{$\pm$0.44} & 76.44\scriptsize{$\pm$0.62} \\
        \midrule
        GAT & 39.21\scriptsize{$\pm$3.08} & 35.62\scriptsize{$\pm$2.06} & 80.87\scriptsize{$\pm$0.30} & 49.09\scriptsize{$\pm$0.63} & 92.01\scriptsize{$\pm$0.68} & \textcolor{red}{83.70\scriptsize{$\pm$0.47}} & 77.43\scriptsize{$\pm$1.20} \\
        GAT-sep & 39.26\scriptsize{$\pm$2.50} & 35.46\scriptsize{$\pm$3.10} & \textcolor{blue}{88.75\scriptsize{$\pm$0.41}} & 52.70\scriptsize{$\pm$0.62} & \textbf{93.91\scriptsize{$\pm$0.35}} & \textcolor{blue}{83.78\scriptsize{$\pm$0.43}} & 76.79\scriptsize{$\pm$0.71} \\
        GT & 38.87\scriptsize{$\pm$3.66} & 36.30\scriptsize{$\pm$1.98} & 86.51\scriptsize{$\pm$0.73} & 51.17\scriptsize{$\pm$0.66} & 91.85\scriptsize{$\pm$0.76} & 83.23\scriptsize{$\pm$0.64} & 77.95\scriptsize{$\pm$0.68} \\
        GT-sep & 40.31\scriptsize{$\pm$3.01} & 36.66\scriptsize{$\pm$1.63} & \textcolor{red}{87.32\scriptsize{$\pm$0.39}} & 52.18\scriptsize{$\pm$0.80} & \textcolor{red}{92.29\scriptsize{$\pm$0.47}} & 82.52\scriptsize{$\pm$0.92} & \textcolor{blue}{78.05\scriptsize{$\pm$0.93}} \\
        \midrule
        H$_2$GCN & 26.75\scriptsize{$\pm$3.64} & 35.10\scriptsize{$\pm$1.15} & 60.11\scriptsize{$\pm$0.52} & 36.47\scriptsize{$\pm$0.23} & 89.71\scriptsize{$\pm$0.31} & 73.35\scriptsize{$\pm$1.01} & 63.59\scriptsize{$\pm$1.46} \\
        CPGNN & 33.00\scriptsize{$\pm$3.15} & 30.04\scriptsize{$\pm$2.03} & 63.96\scriptsize{$\pm$0.62} & 39.79\scriptsize{$\pm$0.77} & 52.03\scriptsize{$\pm$5.46} & 73.36\scriptsize{$\pm$1.01} & 65.96\scriptsize{$\pm$1.95} \\
        GPR-GNN & 39.93\scriptsize{$\pm$3.30} & 38.95\scriptsize{$\pm$1.99} & 64.85\scriptsize{$\pm$0.27} & 44.88\scriptsize{$\pm$0.34} & 86.24\scriptsize{$\pm$0.61} & 72.94\scriptsize{$\pm$0.97} & 55.48\scriptsize{$\pm$0.91} \\
        FSGNN & 40.61\scriptsize{$\pm$2.97} & 35.92\scriptsize{$\pm$1.32} & 79.92\scriptsize{$\pm$0.56} & \textcolor{red}{52.74\scriptsize{$\pm$0.83}} & 90.08\scriptsize{$\pm$0.70} & 82.76\scriptsize{$\pm$0.61} & \textbf{78.86\scriptsize{$\pm$0.92}} \\
        GloGNN & 25.90\scriptsize{$\pm$3.58} & 35.11\scriptsize{$\pm$1.24} & 59.63\scriptsize{$\pm$0.69} & 36.89\scriptsize{$\pm$0.14} & 51.08\scriptsize{$\pm$1.23} & 73.39\scriptsize{$\pm$1.17} & 65.74\scriptsize{$\pm$1.19} \\
        FAGCN & \textcolor{blue}{41.90\scriptsize{$\pm$2.72}} & \textbf{41.08\scriptsize{$\pm$2.27}} & 65.22\scriptsize{$\pm$0.56} & 44.12\scriptsize{$\pm$0.30} & 88.17\scriptsize{$\pm$0.73} & 77.75\scriptsize{$\pm$1.05} & 77.24\scriptsize{$\pm$1.26} \\
        GBK-GNN & 39.61\scriptsize{$\pm$2.60} & 35.51\scriptsize{$\pm$1.65} & 74.57\scriptsize{$\pm$0.47} & 45.98\scriptsize{$\pm$0.71} & 90.85\scriptsize{$\pm$0.58} & 81.01\scriptsize{$\pm$0.67} & 74.47\scriptsize{$\pm$0.86} \\
        JacobiConv & 39.00\scriptsize{$\pm$4.20} & 29.71\scriptsize{$\pm$1.66} & 71.14\scriptsize{$\pm$0.42} & 43.55\scriptsize{$\pm$0.48} & 89.66\scriptsize{$\pm$0.40} & 68.66\scriptsize{$\pm$0.65} & 73.88\scriptsize{$\pm$1.16} \\
        \midrule
        \textbf{PRAT (ours)} & \textbf{42.69\scriptsize{$\pm$2.66}} & \textcolor{blue}{39.60\scriptsize{$\pm$2.25}} & \textbf{90.08\scriptsize{$\pm$0.62}} & \textbf{53.78\scriptsize{$\pm$0.66}} & 92.24\scriptsize{$\pm$0.68} & \textbf{85.04\scriptsize{$\pm$0.48}} & \textcolor{red}{78.04\scriptsize{$\pm$1.19}} \\
        \bottomrule
    \end{tabular}
    \label{tab:platonov_results}
\end{table}

\cref{tab:platonov_results} extends the comparison to \citet{platonov2023critical} benchmarks. PRAT is best or statistically tied on four of the seven datasets (Cham.-filt., roman-empire, amazon-ratings, tolokers), ranks 2nd on Sq.-filt., and sits $0.05$pp below the third-placed model on minesweeper and $0.01$pp below GT-sep at rank 3 on questions. Only two of those four are wins; the other two are ties: against SAGE on amazon-ratings a Welch test gives $p = 0.55$, and against FAGCN on filtered Chameleon $p = 0.52$ (\cref{app:significance}). Across the seven benchmarks PRAT therefore outperforms or remains competitive with all standard and heterophily-adapted GNNs reported by~\citet{platonov2023critical}. The ResNet row is graph-aware on every dataset (ResNet+SGC or ResNet+adj) and outscores GCN on filtered Chameleon, yet PRAT leads it on all seven, by $1.23$ to $21.36$ points. On filtered Chameleon, where the duplicate nodes responsible for train--test leakage in the original benchmark are removed, the gain survives; removing that leakage shows the gain is not an artifact of it, though it does not by itself rule out memorization of other kinds. These results support our central thesis: the geometry of the latent space, not the depth of aggregation, is the dominant factor for learning under heterophily.

\subsection{Validation of Geometric Hypothesis Based on Synthetic Data}
Generated graphs let us vary one structural property at a time. Nodes, edges, features and classes are held fixed across all of them, and every cell is five random $60/20/20$ splits. \cref{syntheticdatagen} gives the generator.

\begin{figure}[t]
    \centering
    \begin{minipage}[t]{0.59\columnwidth}
        \centering
        \begin{tikzpicture}
            \begin{axis}[
                width=\columnwidth,
                height=4.1cm,
                xlabel={Homophily Level},
                ylabel={Accuracy (\%)},
                xmin=0, xmax=1.0,
                ymin=15, ymax=105,
                xtick={0.0, 0.2, 0.4, 0.6, 0.8, 1.0},
                ytick={20, 40, 60, 80, 100},
                legend style={
                    at={(0.50,0.3)},
                    anchor=north west,
                    legend columns=2,
                    draw=none,
                    fill=white,
                    fill opacity=0.8,
                    font=\scriptsize,
                    column sep=0.1cm
                },
                ymajorgrids=true,
                grid style=dashed,
                tick label style={font=\footnotesize},
                label style={font=\small},
                cycle list name=color list
            ]
            \addplot[
                color=red,
                line width=1.0pt,
                mark=*,
                mark size=0.0pt
            ] coordinates {
                (0.00,51.5) (0.05,70.2) (0.10,80.3) (0.15,81.7) (0.20,82.3)
                (0.25,80.7) (0.30,81.0) (0.35,81.4) (0.40,82.6) (0.45,81.7)
                (0.50,82.8) (0.55,84.5) (0.60,83.6) (0.65,83.6) (0.70,86.3)
                (0.75,85.5) (0.80,87.8) (0.85,90.5) (0.90,90.6) (0.95,95.1) (1.00,98.8)
            };
            \addlegendentry{\tiny{\textbf{PRAT}}}

            \addplot[
                color=blue,
                line width=1.0pt,
                mark=square,
                mark size=0.0pt,
                dashed
            ] coordinates {
                (0.00,30.2) (0.05,32.4) (0.10,28.6) (0.15,35.4) (0.20,34.2)
                (0.25,38.9) (0.30,41.9) (0.35,45.7) (0.40,51.6) (0.45,54.5)
                (0.50,58.6) (0.55,65.2) (0.60,68.7) (0.65,72.6) (0.70,78.2)
                (0.75,81.6) (0.80,84.8) (0.85,88.8) (0.90,90.2) (0.95,96.0) (1.00,98.5)
            };
            \addlegendentry{\tiny{GAT}}

            \addplot[
                color=violet,
                line width=1.0pt,
                mark=triangle,
                mark size=0.0pt,
                dashed
            ] coordinates {
                (0.00,37.2) (0.05,29.1) (0.10,32.6) (0.15,33.7) (0.20,36.5)
                (0.25,37.8) (0.30,42.3) (0.35,45.0) (0.40,52.0) (0.45,55.6)
                (0.50,59.8) (0.55,66.4) (0.60,69.3) (0.65,73.2) (0.70,78.3)
                (0.75,80.6) (0.80,84.7) (0.85,87.4) (0.90,90.6) (0.95,95.7) (1.00,98.7)
            };
            \addlegendentry{\tiny{GATv2}}

            \addplot[
                color=orange,
                line width=1.0pt,
                mark=diamond,
                mark size=0.0pt,
                dashdotted
            ] coordinates {
                (0.00,25.3) (0.05,26.9) (0.10,30.2) (0.15,33.1) (0.20,34.0)
                (0.25,38.2) (0.30,40.1) (0.35,45.8) (0.40,52.7) (0.45,52.5)
                (0.50,59.9) (0.55,67.3) (0.60,69.0) (0.65,70.6) (0.70,75.1)
                (0.75,77.7) (0.80,84.8) (0.85,88.5) (0.90,91.0) (0.95,96.5) (1.00,98.3)
            };
            \addlegendentry{\tiny{FAGCN}}
            \end{axis}
        \end{tikzpicture}
        \captionof{figure}{Accuracy across edge homophily levels, within the attention family.}
        \label{fig:homophily_trend}
    \end{minipage}
    \hfill
    \begin{minipage}[t]{0.39\columnwidth}
        \centering
        \begin{tikzpicture}
            \begin{axis}[
                width=\columnwidth,
                height=4.1cm,
                xlabel={Homophily Level},
                ylabel={Ratio (\%)},
                xmin=-0.02, xmax=1.02,
                ymin=15, ymax=105,
                xtick={0, 0.2, 0.4, 0.6, 0.8, 1.0},
                ytick={20, 40, 60, 80, 100},
                grid=both,
                grid style={line width=0.3pt, draw=gray!30},
                tick label style={font=\footnotesize},
                label style={font=\small},
            ]

            \draw[gray, dashed, line width=0.8pt] (axis cs:-0.02,50) -- (axis cs:1.02,50);

            \addplot[
                color=red,
                mark=*,
                mark size=0.0pt,
                line width=1.0pt,
            ] coordinates {
                (0.00, 26.50) (0.05, 31.50) (0.10, 39.00) (0.15, 48.00) (0.20, 53.00)
                (0.25, 62.00) (0.30, 70.50) (0.35, 71.50) (0.40, 80.00) (0.45, 84.00)
                (0.50, 91.50) (0.55, 97.00) (0.60, 99.50) (0.65, 99.00) (0.70, 100.00)
                (0.75, 100.00) (0.80, 100.00) (0.85, 100.00) (0.90, 100.00) (0.95, 100.00)
                (1.00, 100.00)
            };

            \end{axis}
        \end{tikzpicture}
        \captionof{figure}{Ratio of positive eigenvalues of the learned metric in PRAT across varying homophily levels.}
        \label{fig:positive_ratio}
    \end{minipage}
\end{figure}

\paragraph{Varying Homophily level} We generated synthetic graphs with edge homophily $h_E \in \{0, 0.05, \dots, 1.0\}$ and evaluated classification accuracy. The comparison stays inside the attention family, so the receptive field is fixed and only the metric differs: GAT and GATv2 keep it positive definite, FAGCN signs the aggregation weights without touching the geometry, and PRAT makes the metric itself indefinite. In \cref{fig:homophily_trend}, PRAT holds its accuracy across the sweep while the other three fall away together below $h_E = 0.5$. The gap to FAGCN averages $40.6$ points over the nine levels at $h_E \leq 0.4$. The curves converge above $h_E = 0.8$, where a positive-definite metric is no longer a mismatch. Signed weights alone do not substitute for an indefinite metric.

\cref{fig:positive_ratio} tracks the ratio of positive eigenvalues in the learned metric: as homophily falls the negative share rises, so the signature the model settles on is not imposed but follows the data. On real graphs the learned $\Lambda$ instead keeps the initial sign of every entry with $|\lambda| > 10^{-4}$ in $\Lambda_{\mathcal{G},\text{train}}$, which \cref{app:init_controls} reports.
%This phenomenon confirms that the observed eigenvalue evolution is an emergent property driven by the injected geometric prior, rather than a handcrafted heuristic.
\begin{figure}
    \centering
    \begin{subfigure}[t]{0.46\textwidth}
        \centering
        \includegraphics[width=\columnwidth]{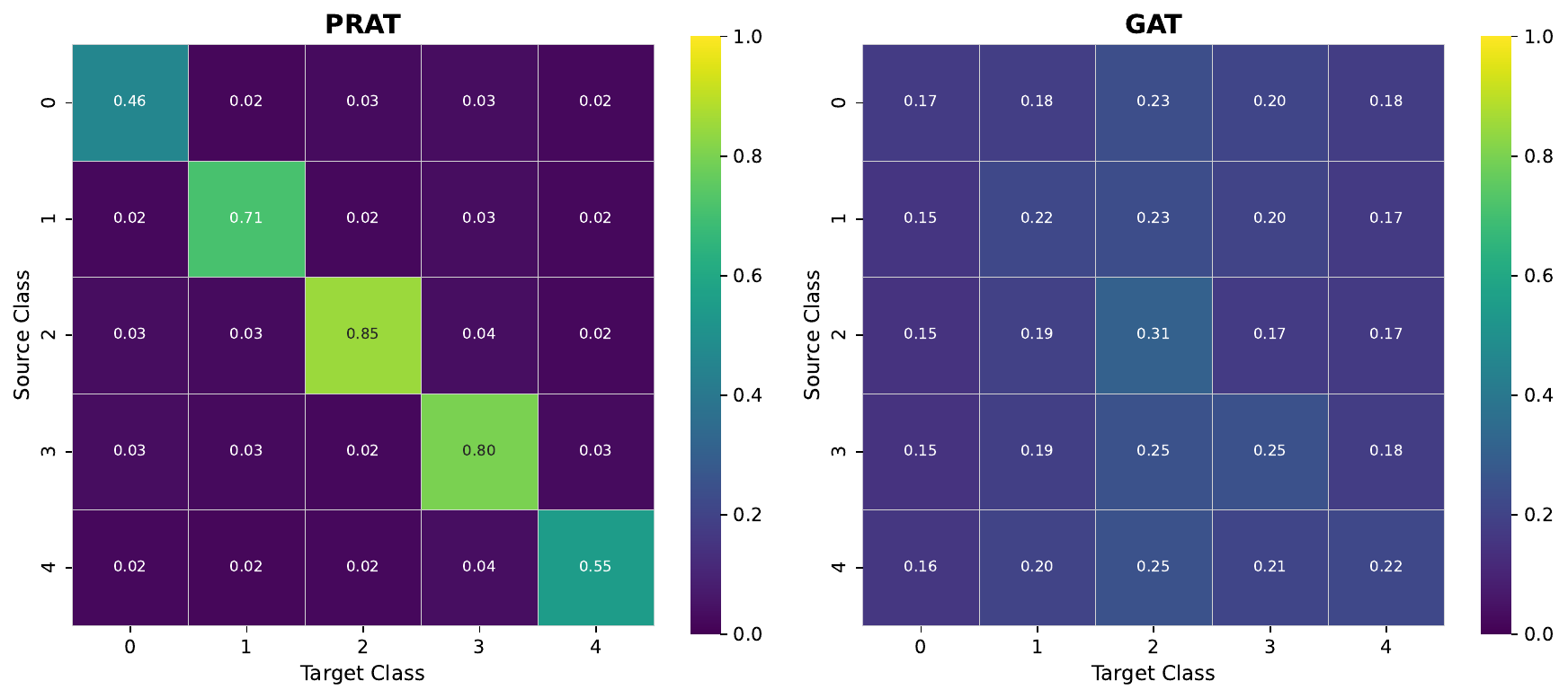}
        \caption{$h_E = 0.1$ (heterophilic)}
        \label{fig:attention_comparison_h01}
    \end{subfigure}
    \begin{subfigure}[t]{0.46\textwidth}
        \centering
        \includegraphics[width=\columnwidth]{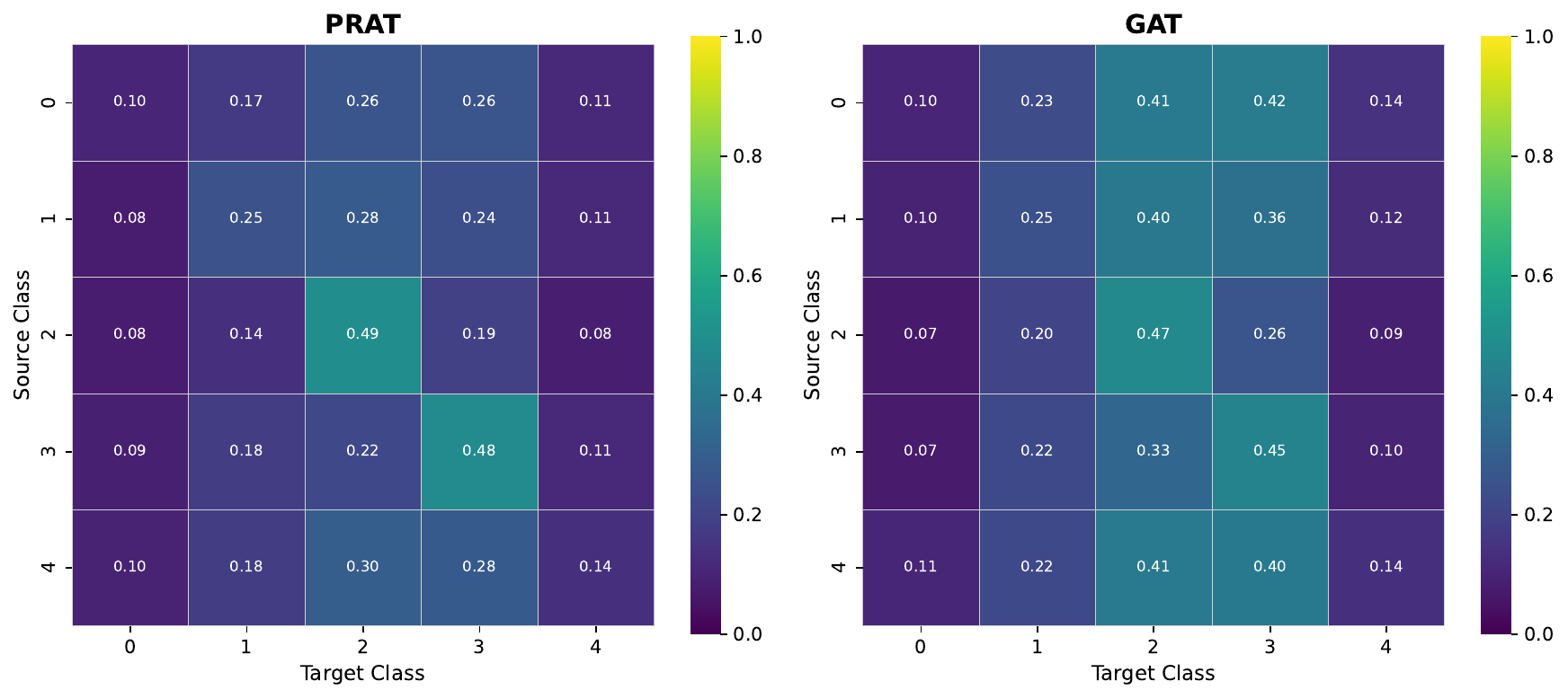}
        \caption{$h_E = 0.9$ (homophilic)}
        \label{fig:attention_comparison_h09}
    \end{subfigure}

    \caption{Attention score distributions of PRAT and GAT. Each heatmap gives mean attention from source class (row) to target class (column).}
    \label{fig:attention_heatmap}
\end{figure}

\cref{fig:attention_heatmap} visualizes the attention scores of PRAT and GAT, aggregated by source-target label pairs across all edges. Unlike GAT, %shows no significant variation in its score distribution at different homophily levels, 
PRAT concentrates attention on same-class connections (the diagonal) at $h_E = 0.1$, suppressing dissimilar neighbors while keeping the node's own signal, but does not do so at $h_E = 0.9$. The metric changes how attention uses the same neighborhood. %distinction from conventional GAT, which lacks the capacity to modulate attention based on the underlying graph structure.

\paragraph{Varying Graph Signature Index} To investigate the influence of higher-order structural patterns beyond simple homophily, we generated synthetic graphs with a fixed homophily level ($h_E=0.5$) but varying graph signature indices $s_{\mathcal{G}} \in \{(5,0,0), (4,0,1), (3,0,2)\}$.

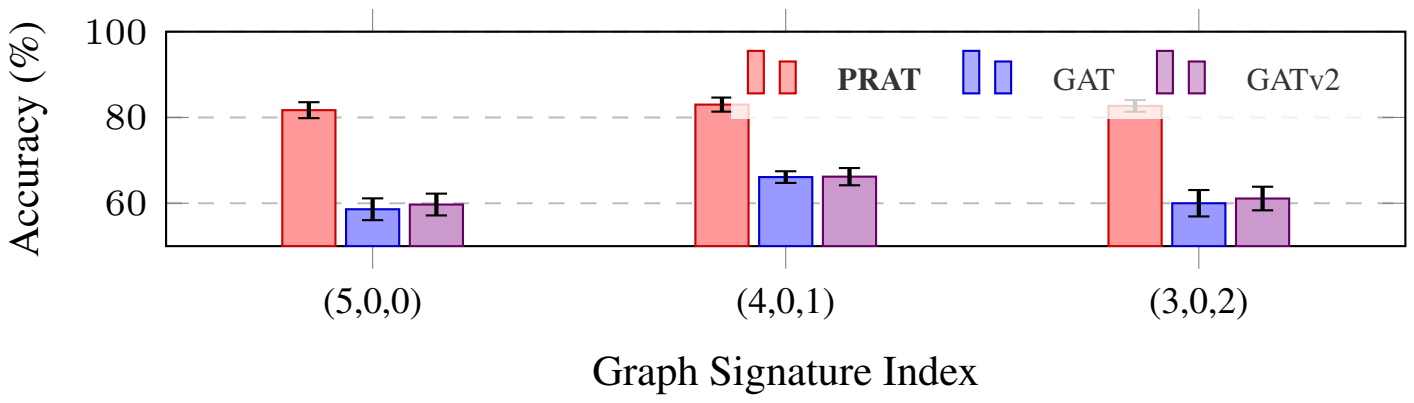
\begin{figure}
    \centering
    \begin{tikzpicture}
        \begin{axis}[
            ybar,
            bar width=10pt,
            width=0.7\columnwidth,
            height=3.0cm,
            enlarge x limits=0.25,
            legend style={
                at={(0.97,0.97)},
                anchor=north east,
                legend columns=3,
                draw=none,
                fill=white,
                fill opacity=0.8,
                font=\tiny,
                column sep=0.2cm,
                row sep=1pt
            },
            ylabel={Accuracy (\%)},
            xlabel={Graph Signature Index},
            symbolic x coords={5p-0n, 4p-1n, 3p-2n},
            xticklabels={(5,0,0), (4,0,1), (3,0,2)},
            xtick=data,
            ymin=50, ymax=100,
            ymajorgrids=true,
            grid style=dashed,
            tick label style={font=\scriptsize},
            label style={font=\footnotesize}
        ]
        \addplot[
            fill=red!40,
            draw=red!80!black,
            error bars/.cd,
            y dir=both, y explicit,
            error bar style={line width=0.8pt, black}
        ] coordinates {
            (5p-0n, 81.70) +- (0, 1.86)
            (4p-1n, 83.00) +- (0, 1.64)
            (3p-2n, 82.70) +- (0, 1.36)
        };
        \addlegendentry{\textbf{PRAT}}

        \addplot[
            fill=blue!40,
            draw=blue!80!black,
            error bars/.cd,
            y dir=both, y explicit,
            error bar style={line width=0.8pt, black}
        ] coordinates {
            (5p-0n, 58.60) +- (0, 2.54)
            (4p-1n, 66.10) +- (0, 1.36)
            (3p-2n, 60.00) +- (0, 3.08)
        };
        \addlegendentry{GAT}

        \addplot[
            fill=violet!40,
            draw=violet!80!black,
            error bars/.cd,
            y dir=both, y explicit,
            error bar style={line width=0.8pt, black}
        ] coordinates {
            (5p-0n, 59.70) +- (0, 2.54)
            (4p-1n, 66.20) +- (0, 2.01)
            (3p-2n, 61.10) +- (0, 2.75)
        };
        \addlegendentry{GATv2}
        \end{axis}
    \end{tikzpicture}
    \caption{Performance comparison under fixed $h_E=0.5$ across different graph signature indices.}
    \label{fig:varying_signature}
\end{figure}

In \cref{fig:varying_signature}, baseline performance fluctuates significantly with the graph signature index. Specifically, the standard deviation of accuracy across different signatures is 3.99 for GAT and 3.42 for GATv2. PRAT varies by $0.68$. Two readings follow: PRAT is the less sensitive of the three to latent structure, and $s_{\mathcal{G}}$ separates graphs that a scalar $h_E$ cannot, since these three share an $h_E$ and behave differently. %This validates our theoretical motivation from \cref{graphsignatureindex} for introducing the graph signature as a generalized measure of class compatibility.

\subsection{Ablation Study}

\begin{wraptable}{r}{0.45\textwidth}
    \caption{Ablation study results.}
    \centering
    \setlength{\tabcolsep}{2pt}
    \scriptsize
    \begin{tabular}{@{}lccc@{}}
        \toprule
        \textbf{Dataset} & \textbf{PRAT} & $\text{PRAT}_{\text{fixed}}$ & $\text{PRAT}_{\text{pos}}$ \\
        \midrule
        Cora      & 88.42\scriptsize{$\pm$1.17} & \textbf{88.49\scriptsize{$\pm$1.27}} & 88.18\scriptsize{$\pm$1.10} \\
        PubMed    & \textbf{90.03\scriptsize{$\pm$0.47}} & 89.94\scriptsize{$\pm$0.46} & 89.87\scriptsize{$\pm$0.45} \\
        Citeseer  & 78.06\scriptsize{$\pm$1.37} & 77.93\scriptsize{$\pm$1.38} & \textbf{78.09\scriptsize{$\pm$1.45}} \\
        Chameleon & \textbf{68.73\scriptsize{$\pm$1.74}} & 68.40\scriptsize{$\pm$2.42} & 65.64\scriptsize{$\pm$1.60} \\
        Squirrel  & \textbf{56.96\scriptsize{$\pm$1.51}} & 54.01\scriptsize{$\pm$2.08} & 53.47\scriptsize{$\pm$1.43} \\
        Actor     & \textbf{37.26\scriptsize{$\pm$1.44}} & 36.93\scriptsize{$\pm$1.21} & 36.69\scriptsize{$\pm$0.99} \\
        Cornell   & \textbf{71.89\scriptsize{$\pm$4.71}} & 55.15\scriptsize{$\pm$8.97} & 57.58\scriptsize{$\pm$8.87} \\
        Texas     & \textbf{84.32\scriptsize{$\pm$5.10}} & 74.89\scriptsize{$\pm$5.70} & 76.78\scriptsize{$\pm$2.90} \\
        \midrule
        \textbf{Average} & \textbf{71.96} & 68.22 & 68.29 \\
        \bottomrule
    \end{tabular}
    \label{ablation_result}
\end{wraptable}
To evaluate the effectiveness of graph-aware metric initialization in PRAT, we ran an ablation with two variants: (1) $\text{PRAT}_{\text{pos}}$, whose learnable $\Lambda$ is constrained to be positive definite, and (2) $\text{PRAT}_{\text{fixed}}$, whose $\Lambda$ is frozen at its signature initialization.

In \cref{ablation_result} the gap between PRAT and $\text{PRAT}_{\text{pos}}$ falls on the graphs whose signature has negative eigenvalues: Chameleon, Squirrel, Cornell and Texas all lose under the constraint, while Cora, PubMed and Citeseer, whose signatures are strictly positive, are unchanged. \cref{app:platonov_ablation} repeats the comparison on the Platonov suite with an exact paired test: both significant losses are on indefinite graphs and no graph without a negative eigenvalue changes significantly, while filtered Squirrel, whose negative eigenvalues sit near zero, ties. A negative signature is therefore necessary for a significant loss, and \cref{app:correlation} finds no significant rank correlation between $n_-$ and what signed or high-pass baselines gain, so we offer $s_{\mathcal{G}}$ as a diagnostic and not as a predictor.

The two variants isolate different things. $\text{PRAT}_{\text{pos}}$ removes the freedom to be indefinite and starts $\Lambda$ at a scaled identity, so it measures the sign structure. $\text{PRAT}_{\text{fixed}}$ freezes $\Lambda$ but not the metric, since $\mathbf{D} = (\mathbf{W}')^\top \Lambda \mathbf{W}'$ keeps adapting through the projection, so it measures whether the inertia has to be supplied at initialization. Its drop sits on the small graphs, $16.7$ points on Cornell and $9.4$ on Texas, where the training-split signature varies most; on the large Platonov graphs the two are within noise. Learning $\Lambda$ earns its place where the signature is noisy, not where it is clean.

Which part of the signature carries the effect differs by graph. In \cref{app:init_controls} the only control that loses decisively on roman-empire deletes the negative directions and keeps the magnitudes, $2.26$ points at $p = 0.002$, while sign-only, sign-flipped and three random draws stay within $0.7$. On filtered Chameleon the magnitudes matter too, since every random draw loses $1.63$ to $2.62$ points.

%Consequently, metric learning is an integral component of PRAT, with the graph signature matrix offering an effective initialization point. This approach enhances model performance by injecting global structural information, which can hardly be captured by local message passing, directly into the metric.

\section{Related Works}
\textbf{GNNs for Heterophily.}
Standard GNNs assume homophily, acting as low-pass filters that smooth features across edges. Heterophily-adapted models answer this architecturally: Geom-GCN~\cite{Pei2020Geom-GCN:} and $H_2$GCN~\cite{NEURIPS2020_58ae23d8} separate aggregation from the immediate neighborhood, spectral methods such as GPR-GNN~\cite{chien2021adaptive} and BernNet~\cite{he2021bernnet} learn polynomial filters that mix high- and low-frequency signals, and attention-based models use signed edge weights~\cite{fagcn2021, yan2022} to represent repulsion. All of these operate within a Euclidean space. %, effectively ``simulating'' repulsion through weight engineering rather than addressing the underlying geometric mismatch.

\textbf{Pseudo-Riemannian graph models.} We are not the first to bring an indefinite geometry to graphs. Q-GCN~\cite{Xiong2021Pseudo} embeds node features on a pseudo-hyperboloid $\mathcal{Q}^{s,t}_\beta$ and convolves there through diffeomorphic exponential and logarithmic maps; Q-GT~\cite{le2025pseudo} carries the same ambient space into a transformer with linear attention. Three differences separate them from PRAT. First, where the indefinite form lives: in both it is the canonical scalar product of the manifold the nodes are embedded on, whereas in PRAT it is the attention logit itself, a signed squared pseudo-distance $\Delta^\top \mathbf{D} \Delta$. Q-GT's attention weights come from a positive-definite kernel, so its indefinite product never reaches a score. Second, how the signature is chosen: Q-GCN picks the number of time dimensions by a curvature heuristic and a sweep, which its own conclusion names as an open limitation, and Q-GT selects its own signature $(p,q)$, the counts of positive and negative dimensions, offline by matching a sectional-curvature distribution. Neither criterion reads the label topology, so neither can see heterophily, while $s_{\mathcal{G}}$ is built from that topology and $\Lambda$ is learned from the spectrum of $S_{\mathcal{G},\text{train}}$. Third, coverage: Q-GCN's labeled evaluation is citation networks and Airport, Q-GT adds two non-homophilous social graphs, and neither reports the WebKB, Wikipedia or Platonov suites used here. We claim a learnable label-initialized indefinite metric inside attention together with a diagnostic for when it should help, not priority for pseudo-Riemannian geometry on graphs.
ACM-GCN~\cite{luan2022revisiting} is closest on the channel side, mixing aggregation, diversification and identity channels node-wise, which puts the sign in the channel domain rather than in a metric. Its published numbers use different splits, so we cite it as a counterpart, not a comparable row.

\textbf{Geometric Graph Learning.} Hyperbolic GNNs~\cite{chami2019hyperbolic,zhang2021hyperbolic} use constant negative curvature for hierarchical structure, and ARGNN~\cite{wang2025adaptiveriemanniangraphneural} makes the metric node-wise and learnable. Both stay Riemannian ($g > 0$); only an indefinite metric admits a negative squared distance~\cite{NEURIPS2020_123b7f02}.

\textbf{Graph Attention.}
Standard $L_2$ attention can be reformulated as an SDP and is biased accordingly %due to the non-negative nature of Euclidean distances
~\cite{ma2025plaintransformerspowerfulgraph}. DGAT~\cite{lu2024representationlearningheterophilicgraph} uses spectral distances for rewiring in a separate preprocessing step, and global Graph Transformers~\cite{ying2021transformers,rampavsek2022recipe,choi2024topology} reach long range at quadratic cost while remaining Euclidean. PRAT instead reads attention as a tangent-space interaction on a learnable manifold.

\textbf{Statistical Foundations.}
The Generalised Random Dot Product Graph (GRDPG)~\cite{rubindelanchy2022grdpg} shows statistically that indefinite inner products are needed to model disassortative networks. 
% Under a stochastic block model with heterophilic connectivity, the standard positive-definite RDPG cannot represent the block matrix, requiring a generalised model with signature $I_{p,q}$. 
Theorem~\ref{theorem:main_theorem0} reaches a complementary conclusion from the geometric side. GRDPG gives a fixed spectral embedding. PRAT learns the indefinite metric end-to-end, so attention moves with it at every layer.

\section*{Conclusion}
We examined the positive-definite geometric prior carried by diffusion, hyperbolic and dot-product attention models, and asked what it costs when the labels induce an indefinite form. Theorem~\ref{theorem:main_theorem0} answers in one direction: where $S_{\mathcal{G}}$ is indefinite and the pushforward is surjective, no diffeomorphism of the target turns the Riemannian pullback into $f^{*}S_{\mathcal{G}}$. That is a statement about those model families and about graphs with a negative signature, not about heterophily as a whole.

The empirical picture matches that scope. The Graph Signature Index says which graphs carry a negative signature. In the paired Platonov ablation every significant loss from constraining the metric falls on such a graph, and no other graph changes significantly. The condition is necessary for a detectable loss. Several of our heterophilic benchmarks do not meet it, so the diagnosis, not heterophily as such, is a precondition of the indefinite metric's advantage.

\paragraph{Limitations and future work.}
The signature is an initialization prior rather than an architectural requirement, since the forward pass never reads labels and $\Lambda$ is an ordinary learnable parameter. Even so, four limits are worth stating.
It is computed from training-node labels, which ties the scheme as presented to (semi-)supervised node classification. \cref{app:labelfree} weakens that, but a full self-supervised treatment, objective as well as initialization, is outside our scope.
Inductive use needs no recomputation in the forward pass, yet signature shift between training and test graphs is open.
Graph-level tasks are out of reach as the method stands, since $S_{\mathcal{G}}$ is undefined without node labels.
Finally, $\mathrm{rank}(\mathbf{D}) \leq |L|$ bounds the structure the metric can carry when classes are few, and on the binary benchmarks the inertia reduces to a single bit. Dynamic curvature and directed graphs are natural next steps.

\begin{ack}
This research was supported by the “Advanced GPU Utilization Support Program” funded by the Government of the Republic of Korea (Ministry of Science and ICT). Sun Woo Park would like to sincerely thank Max Planck Institute for Mathematics for its kind and generous support. The authors declare no competing interests.
\end{ack}

\bibliographystyle{plainnat}
\bibliography{refs}

@ARTICLE{S52,
    author = {Sylvester, J. J.},
    title = {XIX. A demonstration of the theorem that every homogeneous quadratic polynomial is reducible by real orthogonal substitutions to the form of a sum of positive and negative squares},
    journal = {The London, Edinburgh, and Dublin Philosophical Magazine and Journal of Science},
    volume = {4},
    number = {23},
    pages = {138--142},
    year = {1852}
}

@book{Tu17,
    author = {Tu, Loring},
    title = {Differential Geometry: Connections, Curvature, and Characteristic Classes},
    journal = {Graduate Texts in Mathematics},
    publisher = {Springer},
    year = {2017}
}

@inproceedings{
shi2021masked,
title={Masked Label Prediction: Unified Message Passing Model for Semi-Supervised Classification},
author={Shi, Yunsheng and Huang, Zhengjie and Feng, Shikun and Zhong, Hui and Wang, Wenjing and Sun, Yu},
booktitle={Proceedings of the Thirtieth International Joint Conference on Artificial Intelligence, {IJCAI-21}},
publisher={International Joint Conferences on Artificial Intelligence Organization},
pages={1548--1554},
year={2021},
doi={10.24963/ijcai.2021/214}
}

@misc{M09,
    author = {Moore, John Douglas},
    Title = {Lectures on Differential Geometry},
    howpublished = {Lecture notes, University of California, Santa Barbara. \url{https://web.math.ucsb.edu/~moore/riemanniangeometry.pdf}},
    year = {2009}
}

@article{chamberlain2021grand,
  title={GRAND: Graph Neural Diffusion},
  author={Chamberlain, Benjamin Paul and Rowbottom, James and Goronova, Maria and Webb, Stefan and Rossi, 
  Emanuele and Bronstein, Michael M},
  journal={Proceedings of the 38th International Conference on Machine Learning,
               (ICML) 2021, 18-24 July 2021, Virtual Event},
  year={2021}
}

@inproceedings{
bodnar2022neural,
title={Neural Sheaf Diffusion: A Topological Perspective on Heterophily and Oversmoothing in {GNN}s},
author={Cristian Bodnar and Francesco Di Giovanni and Benjamin Paul Chamberlain and Pietro Liò and Michael M. Bronstein},
booktitle={Advances in Neural Information Processing Systems},
volume={35},
pages={18527--18541},
year={2022}
}

@article{Zhu21, title={Graph Neural Networks with Heterophily}, volume={35}, url={https://ojs.aaai.org/index.php/AAAI/article/view/17332}, number={12}, journal={Proceedings of the AAAI Conference on Artificial Intelligence}, author={Zhu, Jiong and Rossi, Ryan A. and Rao, Anup and Mai, Tung and Lipka, Nedim and Ahmed, Nesreen K. and Koutra, Danai}, year={2021}, month={May}, pages={11168-11176} }

@inproceedings{yan2022,
  title={Two sides of the same coin: Heterophily and oversmoothing in graph convolutional neural networks},
  author={Yan, Yujun and Hashemi, Milad and Swersky, Kevin and Yang, Yaoqing and Koutra, Danai},
  booktitle={2022 IEEE International Conference on Data Mining (ICDM)},
  pages={1287--1292},
  year={2022},
  organization={IEEE}
}

@inproceedings{Pei2020Geom-GCN:,
title={Geom-GCN: Geometric Graph Convolutional Networks},
author={Hongbin Pei and Bingzhe Wei and Kevin Chen-Chuan Chang and Yu Lei and Bo Yang},
booktitle={International Conference on Learning Representations},
year={2020},
url={https://openreview.net/forum?id=S1e2agrFvS}
}

@Book{Ha02,
    title={Algebraic Topology},
    author={Hatcher, Allen},
    publisher = {Cambridge University Press},
    pages={1--96},
    year={2002}
}

@inproceedings{
ji2025rethinking,
title={Rethinking Graph Neural Networks From A Geometric Perspective Of Node Features},
author={Feng Ji and Yanan Zhao and Kai Zhao and Hanyang Meng and Jielong Yang and Wee Peng Tay},
booktitle={The Thirteenth International Conference on Learning Representations},
year={2025},
url={https://openreview.net/forum?id=lBMRmw59Lk}
}

@inproceedings{
xu2018how,
title={How Powerful are Graph Neural Networks?},
author={Keyulu Xu and Weihua Hu and Jure Leskovec and Stefanie Jegelka},
booktitle={International Conference on Learning Representations},
year={2019},
url={https://openreview.net/forum?id=ryGs6iA5Km},
}

@inproceedings{
vel2018graph,
title={Graph Attention Networks},
author={Petar Veličković and Guillem Cucurull and Arantxa Casanova and Adriana Romero and Pietro Liò and Yoshua Bengio},
booktitle={International Conference on Learning Representations},
year={2018},
url={https://openreview.net/forum?id=rJXMpikCZ},
}

@inproceedings{
brody2022how,
title={How Attentive are Graph Attention Networks? },
author={Shaked Brody and Uri Alon and Eran Yahav},
booktitle={International Conference on Learning Representations},
year={2022},
url={https://openreview.net/forum?id=F72ximsx7C1}
}

@article{vaswani2017attention,
  title={Attention is all you need},
  author={Vaswani, Ashish and Shazeer, Noam and Parmar, Niki and Uszkoreit, Jakob and Jones, Llion and Gomez, Aidan N and Kaiser, {\L}ukasz and Polosukhin, Illia},
  journal={Advances in neural information processing systems},
  volume={30},
  year={2017}
}

@inproceedings{
kipf2017semisupervised,
title={Semi-Supervised Classification with Graph Convolutional Networks},
author={Thomas N. Kipf and Max Welling},
booktitle={International Conference on Learning Representations},
year={2017},
url={https://openreview.net/forum?id=SJU4ayYgl}
}

@inproceedings{park2022pwlr,
  title={The pwlr graph representation: A persistent weisfeiler-lehman scheme with random walks for graph classification},
  author={Park, Sun Woo and Choi, Yun Young and Joe, Dosang and Choi, U Jin and Woo, Youngho},
  booktitle={Topological, Algebraic and Geometric Learning Workshops 2022},
  pages={287--297},
  year={2022},
  organization={PMLR}
}

@article{park2026functorial,
  title={A Functorial Formulation of Neighborhood Aggregating Deep Learning},
  author={Park, Sun Woo and Choi, Yun Young and Choi, U Jin and Woo, Youngho},
  journal={arXiv preprint arXiv:2604.24672},
  year={2026}
}

@inproceedings{choi2022cycle,
  title={Cycle to clique (cy2c) graph neural network: A sight to see beyond neighborhood aggregation},
  author={Choi, Yun Young and Park, Sun Woo and Woo, Youngho and Choi, U Jin},
  booktitle={The eleventh international conference on learning representations},
  year={2022}
}

@inproceedings{Xiong2021Pseudo,
  title={Pseudo-Riemannian Graph Convolutional Networks},
  author={Bo Xiong and Shichao Zhu and Nico Potyka and Shirui Pan and Chuan Zhou and Steffen Staab},
  booktitle={Advances in Neural Information Processing Systems},
  year={2022}
}

@inproceedings{fagcn2021,
  title={Beyond Low-frequency Information in Graph Convolutional Networks},
  author={Deyu Bo and Xiao Wang and Chuan Shi and Huawei Shen},
  booktitle = {{AAAI}},
  publisher = {{AAAI} Press},
  year      = {2021}
}

@article{rozemberczki2021multi,
  title={Multi-scale attributed node embedding},
  author={Rozemberczki, Benedek and Allen, Carl and Sarkar, Rik},
  journal={Journal of Complex Networks},
  volume={9},
  number={2},
  pages={cnab014},
  year={2021},
  publisher={Oxford University Press}
}

@article{sen2008collective,
  title={Collective classification in network data},
  author={Sen, Prithviraj and Namata, Galileo and Bilgic, Mustafa and Getoor, Lise and Galligher, Brian and Eliassi-Rad, Tina},
  journal={AI magazine},
  volume={29},
  number={3},
  pages={93--93},
  year={2008}
}

@inproceedings{yang2016revisiting,
  title={Revisiting semi-supervised learning with graph embeddings},
  author={Yang, Zhilin and Cohen, William and Salakhudinov, Ruslan},
  booktitle={International conference on machine learning},
  pages={40--48},
  year={2016},
  organization={PMLR}
}

@inproceedings{tang2009social,
  title={Social influence analysis in large-scale networks},
  author={Tang, Jie and Sun, Jimeng and Wang, Chi and Yang, Zi},
  booktitle={Proceedings of the 15th ACM SIGKDD international conference on Knowledge discovery and data mining},
  pages={807--816},
  year={2009}
}

@inproceedings{
wang2025adaptiveriemanniangraphneural,
title={Adaptive Riemannian Graph Neural Networks},
author={Xudong Wang and Tongxin Li and Chris Ding and Jicong Fan},
booktitle={Proceedings of the AAAI Conference on Artificial Intelligence},
year={2026}
}

@misc{lu2024representationlearningheterophilicgraph,
      title={Representation Learning on Heterophilic Graph with Directional Neighborhood Attention}, 
      author={Qincheng Lu and Jiaqi Zhu and Sitao Luan and Xiao-Wen Chang},
      year={2024},
      eprint={2403.01475},
      archivePrefix={arXiv},
      primaryClass={cs.LG},
      url={https://arxiv.org/abs/2403.01475}, 
}

@article{
ma2025plaintransformerspowerfulgraph,
title={Plain Transformers Can be Powerful Graph Learners},
author={Liheng Ma and Soumyasundar Pal and Yingxue Zhang and Philip H. S. Torr and Mark Coates},
journal={Transactions on Machine Learning Research},
year={2026},
url={https://openreview.net/forum?id=bEmDvP0fdv}
}

@article{zhang2021hyperbolic,
  title={Hyperbolic graph attention network},
  author={Zhang, Yiding and Wang, Xiao and Shi, Chuan and Jiang, Xunqiang and Ye, Yanfang Fanny},
  journal={IEEE Transactions on Big Data},
  year={2021},
  publisher={IEEE}
}

@inproceedings{hamilton2017inductive,
     author = {Hamilton, William L. and Ying, Rex and Leskovec, Jure},
     title = {Inductive Representation Learning on Large Graphs},
     booktitle = {NIPS},
     year = {2017}
   }

@ARTICLE{7974879,
  author={Bronstein, Michael M. and Bruna, Joan and LeCun, Yann and Szlam, Arthur and Vandergheynst, Pierre},
  journal={IEEE Signal Processing Magazine}, 
  title={Geometric Deep Learning: Going beyond Euclidean data}, 
  year={2017},
  volume={34},
  number={4},
  pages={18-42},
  doi={10.1109/MSP.2017.2693418}}

@inproceedings{NEURIPS2020_58ae23d8,
 author = {Zhu, Jiong and Yan, Yujun and Zhao, Lingxiao and Heimann, Mark and Akoglu, Leman and Koutra, Danai},
 booktitle = {Advances in Neural Information Processing Systems},
 editor = {H. Larochelle and M. Ranzato and R. Hadsell and M.F. Balcan and H. Lin},
 pages = {7793--7804},
 publisher = {Curran Associates, Inc.},
 title = {Beyond Homophily in Graph Neural Networks: Current Limitations and Effective Designs},
 url = {https://proceedings.neurips.cc/paper_files/paper/2020/file/58ae23d878a47004366189884c2f8440-Paper.pdf},
 volume = {33},
 year = {2020}
}

@inproceedings{he2021bernnet,
  title={BernNet: Learning Arbitrary Graph Spectral Filters via Bernstein Approximation},
  author={He, Mingguo and Wei, Zhewei and Huang, Zengfeng and Xu, Hongteng},
  booktitle={NeurIPS},
  year={2021}
}

@inproceedings{
chien2021adaptive,
title={Adaptive Universal Generalized PageRank Graph Neural Network},
author={Eli Chien and Jianhao Peng and Pan Li and Olgica Milenkovic},
booktitle={International Conference on Learning Representations},
year={2021},
url={https://openreview.net/forum?id=n6jl7fLxrP}
}

@inproceedings{
gasteiger2022predictpropagategraphneural,
title={Predict then Propagate: Graph Neural Networks meet Personalized PageRank},
author={Johannes Gasteiger and Aleksandar Bojchevski and Stephan Günnemann},
booktitle={International Conference on Learning Representations},
year={2019},
url={https://openreview.net/forum?id=H1gL-2A9Ym}
}

@inproceedings{choi2024topology,
  title={Topology-Informed Graph Transformer},
  author={Choi, Yun Young and Park, Sun Woo and Lee, Minho and Woo, Youngho},
  booktitle={Proceedings of the Geometry-grounded Representation Learning and Generative Modeling Workshop (GRaM)},
  series={Proceedings of Machine Learning Research},
  volume={251},
  pages={20--34},
  publisher={PMLR},
  year={2024}
}

@article{ying2021transformers,
  title={Do transformers really perform badly for graph representation?},
  author={Ying, Chengxuan and Cai, Tianle and Luo, Shengjie and Zheng, Shuxin and Ke, Guolin and He, Di and Shen, Yanming and Liu, Tie-Yan},
  journal={Advances in neural information processing systems},
  volume={34},
  pages={28877--28888},
  year={2021}
}

@article{rampavsek2022recipe,
  title={Recipe for a general, powerful, scalable graph transformer},
  author={Ramp{\'a}{\v{s}}ek, Ladislav and Galkin, Michael and Dwivedi, Vijay Prakash and Luu, Anh Tuan and Wolf, Guy and Beaini, Dominique},
  journal={Advances in Neural Information Processing Systems},
  volume={35},
  pages={14501--14515},
  year={2022}
}

@inproceedings{
topping2022understandingoversquashingbottlenecksgraphs,
title={Understanding over-squashing and bottlenecks on graphs via curvature},
author={Jake Topping and Francesco Di Giovanni and Benjamin Paul Chamberlain and Xiaowen Dong and Michael M. Bronstein},
booktitle={International Conference on Learning Representations},
year={2022},
url={https://openreview.net/forum?id=7UmjRGzp-A}
}

@inproceedings{chami2019hyperbolic,
  title={Hyperbolic graph convolutional neural networks},
  author={Chami, Ines and Ying, Zhitao and R{\'e}, Christopher and Leskovec, Jure},
  booktitle={Advances in Neural Information Processing Systems},
  pages={4869--4880},
  year={2019}
}

@inproceedings{NEURIPS2020_123b7f02,
 author = {Law, Marc and Stam, Jos},
 booktitle = {Advances in Neural Information Processing Systems},
 editor = {H. Larochelle and M. Ranzato and R. Hadsell and M.F. Balcan and H. Lin},
 pages = {1668--1678},
 publisher = {Curran Associates, Inc.},
 title = {Ultrahyperbolic Representation Learning},
 url = {https://proceedings.neurips.cc/paper_files/paper/2020/file/123b7f02433572a0a560e620311a469c-Paper.pdf},
 volume = {33},
 year = {2020}
}

@inproceedings{platonov2023critical,
  title={A Critical Look at the Evaluation of {GNN}s under Heterophily: Are We Really Making Progress?},
  author={Platonov, Oleg and Kuznedelev, Denis and Diskin, Michael and Babenko, Artem and Prokhorenkova, Liudmila},
  booktitle={International Conference on Learning Representations},
  year={2023}
}

@article{rubindelanchy2022grdpg,
  title={A statistical interpretation of spectral embedding: the generalised random dot product graph},
  author={Rubin-Delanchy, Patrick and Cape, Joshua and Tang, Minh and Priebe, Carey E.},
  journal={Journal of the Royal Statistical Society Series B},
  volume={84},
  number={4},
  pages={1446--1473},
  year={2022}
}

@article{maurya2022simplifying,
  title   = {Simplifying approach to node classification in {G}raph {N}eural {N}etworks},
  author  = {Maurya, Sunil Kumar and Liu, Xin and Murata, Tsuyoshi},
  journal = {Journal of Computational Science},
  volume  = {62},
  pages   = {101695},
  year    = {2022},
  publisher = {Elsevier}
}

@inproceedings{li2022finding,
  title     = {Finding Global Homophily in Graph Neural Networks When Meeting Heterophily},
  author    = {Li, Xiang and Zhu, Renyu and Cheng, Yao and Shan, Caihua and Luo, Siqiang and Li, Dongsheng and Qian, Weining},
  booktitle = {International Conference on Machine Learning (ICML)},
  year      = {2022}
}

@inproceedings{du2022gbk,
  title     = {{GBK-GNN}: Gated Bi-Kernel Graph Neural Networks for Modeling Both Homophily and Heterophily},
  author    = {Du, Lun and Shi, Xiaozhou and Fu, Qiang and Ma, Xiaojun and Liu, Hengyu and Han, Shi and Zhang, Dongmei},
  booktitle = {Proceedings of the ACM Web Conference 2022 (WWW)},
  year      = {2022}
}

@inproceedings{wang2022how,
  title     = {How Powerful are Spectral Graph Neural Networks},
  author    = {Wang, Xiyuan and Zhang, Muhan},
  booktitle = {International Conference on Machine Learning (ICML)},
  year      = {2022}
}

@inproceedings{
le2025pseudo,
title={Pseudo-Riemannian Graph Transformer},
author={Le, Viet Quan and Ta, Cuong Van},
booktitle={Advances in Neural Information Processing Systems},
volume={38},
pages={15951--15983},
year={2025}
}

@inproceedings{
luan2022revisiting,
title={Revisiting Heterophily for Graph Neural Networks},
author={Luan, Sitao and Hua, Chenqing and Lu, Qincheng and Zhu, Jiaqi and Zhao, Mingde and Zhang, Shuyuan and Chang, Xiao-Wen and Precup, Doina},
booktitle={Advances in Neural Information Processing Systems},
year={2022}
}

@inproceedings{
zheng2024missing,
title={What Is Missing For Graph Homophily? Disentangling Graph Homophily For Graph Neural Networks},
author={Zheng, Yilun and Luan, Sitao and Chen, Lihui},
booktitle={Advances in Neural Information Processing Systems},
volume={37},
pages={68406--68452},
year={2024}
}

%%%%%%%%%%%%%%%%%%%%%%%%%%%%%%%%%%%%%%%%%%%%%%%%%%%%%%%%%%%%%%%%%%%%%%%%%%%%%%%
%%%%%%%%%%%%%%%%%%%%%%%%%%%%%%%%%%%%%%%%%%%%%%%%%%%%%%%%%%%%%%%%%%%%%%%%%%%%%%%
% APPENDIX
%%%%%%%%%%%%%%%%%%%%%%%%%%%%%%%%%%%%%%%%%%%%%%%%%%%%%%%%%%%%%%%%%%%%%%%%%%%%%%%
%%%%%%%%%%%%%%%%%%%%%%%%%%%%%%%%%%%%%%%%%%%%%%%%%%%%%%%%%%%%%%%%%%%%%%%%%%%%%%%
\appendix
\section{Theoretical analysis and proofs}

\subsection{Normalization of the Graph Signature Matrix} \label{AppendixA:Normalization}

The $(i,j)$-th entry of $S_{\mathcal{G}}$ is defined as $s_{ij} = |E_{ij}| / (n_i \cdot n_j)$, where $|E_{ij}|$ is the number of ordered pairs $(v,w)$ of adjacent nodes with $l(v) = i$ and $l(w) = j$ (so an edge inside class $i$ is counted twice, which is the factor $2$ of the diagonal entries in Appendix~\ref{AppendixA:critical_threshold}), and $n_i, n_j$ are the class sizes. This quantity is the \emph{edge density} between classes $i$ and $j$: the probability that a randomly chosen pair $(v, w)$ with $l(v) = i$, $l(w) = j$ is connected. This normalization is standard in stochastic block models (SBMs) and carries a clear physical interpretation: positive eigenvalues correspond to over-connected class pairs (homophilic tendency), while negative eigenvalues indicate under-connected pairs (heterophilic tendency).

The signature does not depend on the normalization by $n_i \cdot n_j$. Write $C$ for the matrix of raw counts $|E_{ij}|$ and $N_L := \mathrm{diag}(n_1, \ldots, n_{|L|})$. Since every class is nonempty, $N_L$ is invertible, so $S_{\mathcal{G}} = N_L^{-1} C N_L^{-1}$ is congruent to $C$ and has the same inertia by \cref{sylvester}. The normalization acts on the eigenvalues instead. Those of $C$ scale with class sizes, so class imbalance enters their magnitudes, while those of $S_{\mathcal{G}}$ are the spectrum of the connection-probability matrix above. Rescaled, they initialize $\Lambda$ (\cref{eq:factorized_metric}), and \cref{app:init_controls} shows that on filtered Chameleon their magnitudes matter beyond their signs. The critical homophily threshold in Theorem~\ref{theorem:critical_threshold} is derived under this normalization, where the expected matrix takes the form $aI + b(J - I)$ with eigenvalues in closed form, and the resulting $h_E^* = \sum n_k^2 / N^2$ explicitly accounts for class imbalance.

\subsection{A different characterization of \texorpdfstring{$S_{\mathcal{G}}$}{S_G}} \label{AppendixA:S_G-Interpretation}

Using the unfolding-tree perspective on local graph structure~\cite{park2022pwlr}, we interpret entries of the graph signature matrix through depth 1 unfolding trees rooted at each node of $\mathcal{G}$. %We interpret entries of $S_{\mathcal{G}}$ using depth 1 unfolding trees rooted at each node of $\mathcal{G}$.

\begin{definition}
For any node $v \in \mathcal{V}$, we define the probability that a node with a given node label $j$ is connected to a directed edge starting from node $v$ as
\begin{equation}
\begin{split}
    p_{v,j} &:= \mathbb{P} \biggl[ (v,w) \in \mathcal{E} \; \bigg| \; \substack{w \in \mathcal{V} \\ l(w) = j} \biggr] \\
    &= \frac{| \{w \in \mathcal{V} \mid l(w) = j, (v,w) \in \mathcal{E}\}|}{| \{w \in \mathcal{V} \mid l(w) = j\}|}
\end{split}
\end{equation}
\end{definition}

\begin{lemma}
Given a graph $\mathcal{G}$, each matrix entry of $S_{\mathcal{G}}$ satisfies
\begin{equation}
    p_{i,j} = \frac{1}{| \{v \in \mathcal{V} \; | \; l(v) = i\}|} \sum_{\substack{v \in \mathcal{V} \\ l(v) = i}} p_{v,j}
\end{equation}
\end{lemma}
\begin{proof}
We note that
\begin{align}
    \begin{split}
        p_{i,j} &:= \frac{1}{|\{v \in \mathcal{V} \; | \; l(v) = i\}|} \frac{| \{(v,w) \in \mathcal{E} \; | \; l(v) = i, l(w) = j\}|}{| \{v \in \mathcal{V} \; | \; l(v) = j\}|} \\
        &= \frac{1}{|\{v \in \mathcal{V} \; | \; l(v) = i\}|} \frac{\sum_{\substack{v \in \mathcal{V} \\ l(v) = i}} | \{(v,w) \in \mathcal{E} \; | \; l(w) = j\}|}{| \{w \in \mathcal{V} \; | \; l(w) = j\}|} \\
        &= \frac{1}{|\{v \in \mathcal{V} \; | \; l(v) = i\}|} \sum_{\substack{v \in \mathcal{V} \\ l(v) = i}} p_{v,j}
    \end{split}
\end{align}
\end{proof}
In other words, each $(i,j)$-th entry of $S_{\mathcal{G}}$ is the expected value of the probability that a node with a preassigned label $j$ lies in the depth 1 unfolding tree rooted at a node with label $i$.

\subsection{An example of graph signature index for graphs with two discrete node labels} \label{AppendixA:twolabel}

\begin{theorem}[2 Discrete Node Labels] \label{theorem:basic_example}
Let $\mathcal{G} := (\mathcal{V}, \mathcal{E})$ be an undirected graph without self-loops and with $\mathcal{E} \neq \emptyset$. Let $l:\mathcal{V} \to \{1,2\}$ be the discrete node label function over the set of nodes of $\mathcal{G}$. Then the following correspondence holds between $s_{\mathcal{G}}$ and entries of $S_\mathcal{G} := (p_{i,j})_{1 \leq i,j \leq 2}$:
\begin{align}
    \begin{split}
        s_{\mathcal{G}} = (2,0,0) \; &\iff \; p_{1,2} < \sqrt{p_{1,1} p_{2,2}}\\
        s_{\mathcal{G}} = (1,1,0) \; &\iff \; p_{1,2} = \sqrt{p_{1,1} p_{2,2}}\\
        s_{\mathcal{G}} = (1,0,1) \; &\iff \; p_{1,2} > \sqrt{p_{1,1} p_{2,2}}
    \end{split}
\end{align}
\end{theorem}
\begin{proof}
Let $S_{\mathcal{G}}$ be the graph signature matrix of $\mathcal{G}$. Its explicit form is given as
\begin{equation}
    S_\mathcal{G} := \begin{pmatrix}
    p_{1,1} & p_{1,2} \\
    p_{2,1} & p_{2,2}
    \end{pmatrix}
\end{equation}
The characteristic polynomial of $S_{\mathcal{G}}$ is given by
\begin{equation}
    (\lambda-p_{1,1})(\lambda-{p_{2,2}}) - p_{1,2}^2 = 0
\end{equation}
The eigenvalues of $S_{\mathcal{G}}$ are given by
\begin{equation}
    \lambda = \frac{(p_{1,1} + p_{2,2}) \pm \sqrt{(p_{1,1} + p_{2,2})^2 - 4(p_{1,1}p_{2,2} - p_{1,2}^2)}}{2}
\end{equation}
Since $\mathcal{E} \neq \emptyset$, the trace $p_{1,1} + p_{2,2}$ is positive whenever $p_{1,1} p_{2,2} \geq p_{1,2}^2$, as otherwise all three entries vanish. Hence, the graph signature $s_{\mathcal{G}}$ is determined by the sign of the term $p_{1,1} p_{2,2} - p_{1,2}^2$.
\end{proof}

\subsection{Heuristics on heterophilic and homophilic graphs with two discrete node labels} \label{AppendixA:heuristic}

\begin{figure}
\centering
\begin{tikzpicture}
  \begin{axis}[
    xlabel=$p_{1,1}$,
    ylabel=$p_{2,2}$,
    zlabel=$p_{1,2}$,
    no markers, grid,
    domain=0:1,
    xmax=1, ymax=1,
    xmin=0, ymin=0]

    \addplot3[surf, color=blue, opacity=0.01,fill opacity=0.25, faceted color=blue] {sqrt(x)*sqrt(y)};
  \end{axis}
\end{tikzpicture}
  \caption{A phase diagram of graph signature index with two types of discrete node labels. The blue surface denotes the region where $F: p_{1,2} = \sqrt{p_{1,1} p_{2,2}}$.}
  \label{fig:phase_diagram}
\end{figure}
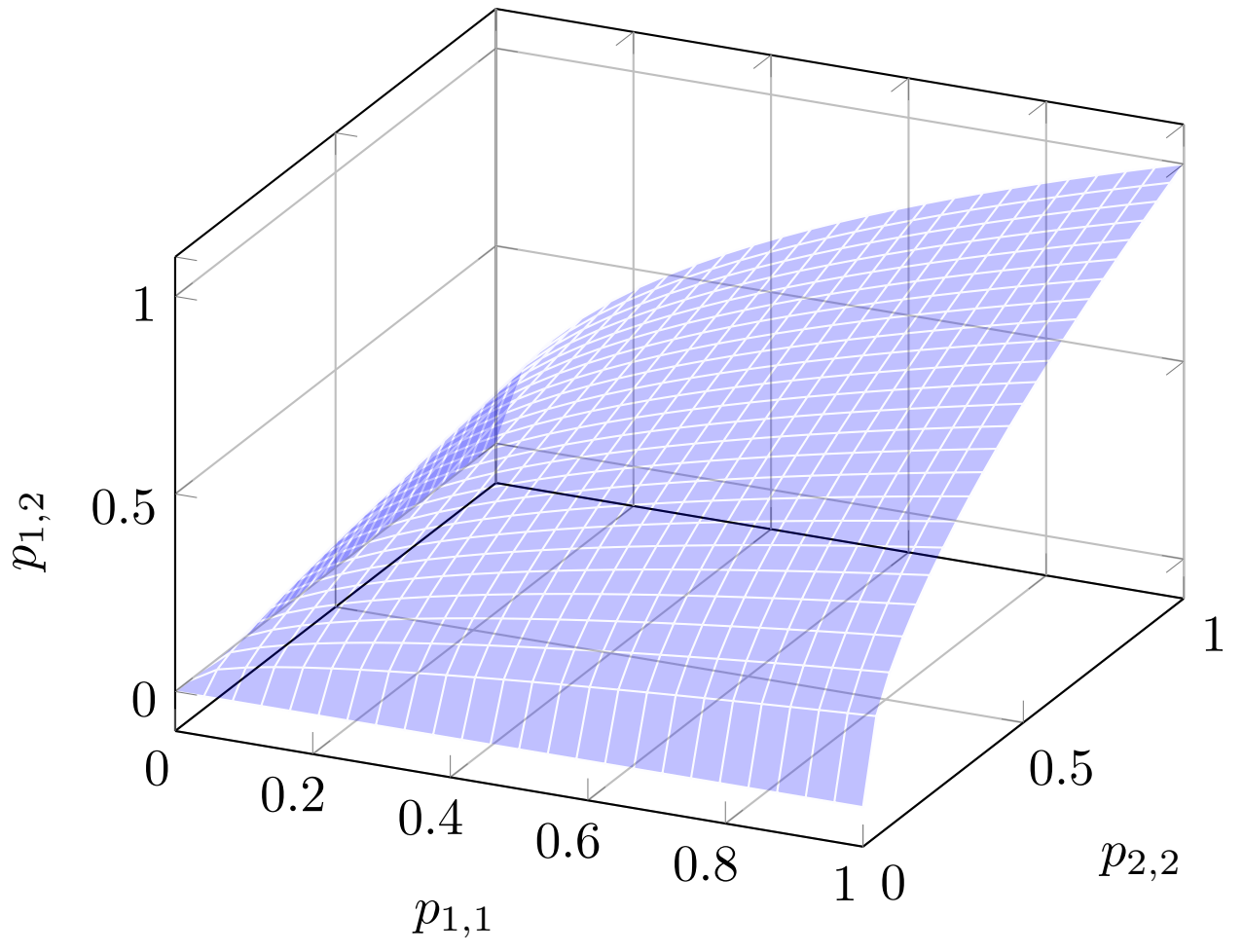

The phase diagram for the graph $F: p_{1,2} - \sqrt{p_{1,1} p_{2,2}} = 0$ is shown in Figure \ref{fig:phase_diagram}, where the region below $F$ characterizes graphs whose nodes with identical labels are more likely to be connected than those with distinct labels (i.e. homophilic graphs), whereas the region above $F$ contains heterophilic graphs. A third region, defined by $F$, contains graphs whose nodes with identical labels are as likely to be connected as those with distinct labels. Note that the volume of the region below $F$ over $0 \leq p_{1,1} \leq 1$ and $0 \leq p_{2,2} \leq 1$ is given by
\begin{equation}
    \int_0^1 \int_0^1 \sqrt{p_{1,1} p_{2,2}} dp_{1,1} dp_{2,2} = \frac{4}{9}.
\end{equation}

Consider the idealized model with 2 discrete node labels whose connectivity parameters $(p_{1,1}, p_{1,2}, p_{2,2})$ are uniformly distributed on $[0,1]^3$, and let $P = (p_{i,j})$ be the symmetric $2 \times 2$ connectivity matrix, with inertia $\operatorname{In}(P)$. The integral computed above shows that
\begin{equation}
    \mathbb{P}\left[ \operatorname{In}(P) = (i,j,k) \right] = \begin{cases}
        \frac{5}{9} & \text{if } (i,j,k) = (1,0,1) \\
        0 & \text{if } (i,j,k) = (1, 1, 0) \\
        \frac{4}{9} & \text{if } (i,j,k) = (2,0,0)
    \end{cases}
\end{equation}

\subsection{Notion of homophily indices}
\label{homophilyindex}

\begin{remark}
Earlier work measured homophily in undirected graphs $\mathcal{G}$ without self-loops by using the node homophily index defined as in \cite{Pei2020Geom-GCN:}
\begin{equation}
    h_V := \frac{1}{|\mathcal{V}|} \sum_{v \in \mathcal{V}} \frac{|\{ w \in \mathcal{V} \; | \; \substack{(v,w) \in \mathcal{E} \\ l(w) = l(v)}\}|}{\deg(v)},
\end{equation}
or by using the edge homophily index defined as in \cite{Zhu21}:
\begin{equation}
    h_E := \frac{| \{ (u,v) \in \mathcal{E} \; | \; l(u) = l(v) \}|}{|\mathcal{E}|}.
\end{equation}
\end{remark}

A common way to distinguish homophilic from heterophilic graphs is to compute the homophily indices $h_V$ and $h_E$ and check whether their values are closer to $0$ or $1$. We demonstrate that the graph signature matrix allows one to closely analyze the relations among different node labels, a trait that numerical homophily indices do not necessarily share.

\subsection{Proof of Theorem \ref{theorem:critical_threshold}}
\label{AppendixA:critical_threshold}

\paragraph{Setup and Notation.}
Consider an undirected graph $\mathcal{G} := (\mathcal{V}, \mathcal{E})$ with $|L| \geq 2$ class labels, where class $i$ contains $n_i \geq 1$ nodes. Let $N = \sum_{i=1}^{|L|} n_i$ denote the total number of nodes. The graph has $|\mathcal{E}| > 0$ edges with edge homophily level $h_E \in [0, 1]$, partitioned into $h_E|\mathcal{E}|$ intra-class edges and $(1-h_E)|\mathcal{E}|$ inter-class edges.

We adopt a \emph{proportional allocation model}:
\begin{itemize}
    \item Intra-class edges are distributed among classes proportionally to $n_i^2$ (the denominator of $s_{i,i}$ in Definition~\ref{def:graph_signature_matrix}).
    \item Inter-class edges are distributed among class pairs $(i, j)$ with $i \neq j$ proportionally to $n_i n_j$.
\end{itemize}
Define the following quantities for notational convenience:
\begin{equation}
    \Sigma_2 := \sum_{k=1}^{|L|} n_k^2, \qquad \Sigma_{\times} := \sum_{i < j} n_i n_j = \frac{N^2 - \Sigma_2}{2}.
\end{equation}

\paragraph{Expected Edge Counts.}
Under the proportional allocation model, the expected number of intra-class edges in class $i$ is
\begin{equation}
    \mathbb{E}[X_i] = h_E|\mathcal{E}| \cdot \frac{n_i^2}{\Sigma_2},
\end{equation}
and the expected number of inter-class edges between classes $i$ and $j$ (where $i \neq j$) is
\begin{equation}
    \mathbb{E}[Y_{ij}] = (1-h_E)|\mathcal{E}| \cdot \frac{n_i n_j}{\Sigma_{\times}} = \frac{2(1-h_E)|\mathcal{E}| \cdot n_i n_j}{N^2 - \Sigma_2}.
\end{equation}

\paragraph{Expected Signature Matrix Entries.}
The expected signature matrix $S_{\mathcal{G}} = \mathbb{E}[\widehat{S}_G]$ has entries given by Definition \ref{def:graph_signature_matrix}. For diagonal entries:
\begin{equation}
    (S_{\mathcal{G}})_{ii} = \frac{2\mathbb{E}[X_i]}{n_i^2} = \frac{2h_E|\mathcal{E}| \cdot n_i^2 / \Sigma_2}{n_i^2} = \frac{2h_E|\mathcal{E}|}{\Sigma_2} =: a.
\end{equation}
For off-diagonal entries:
\begin{equation}
    (S_{\mathcal{G}})_{ij} = \frac{\mathbb{E}[Y_{ij}]}{n_i n_j} = \frac{2(1-h_E)|\mathcal{E}| \cdot n_i n_j / (N^2 - \Sigma_2)}{n_i n_j} = \frac{2(1-h_E)|\mathcal{E}|}{N^2 - \Sigma_2} =: b.
\end{equation}
Both diagonal and off-diagonal entries are constant across all indices, yielding the matrix structure:
\begin{equation}
    S_{\mathcal{G}} = a \cdot I_{|L|} + b \cdot (J_{|L|} - I_{|L|}) = (a - b) I_{|L|} + b J_{|L|},
\end{equation}
where $I_{|L|}$ is the $|L| \times |L|$ identity matrix and $J_{|L|}$ is the $|L| \times |L|$ all-ones matrix.

\paragraph{Eigenvalue Computation.}
The all-ones matrix $J_{|L|}$ has eigenvalue $|L|$ with eigenvector $\mathbf{1}_{|L|} = (1, \ldots, 1)^\top$, and eigenvalue $0$ with multiplicity $|L|-1$ for eigenvectors orthogonal to $\mathbf{1}_{|L|}$. Therefore, the first eigenvalue (with eigenvector $\mathbf{1}_{|L|}$) is
\begin{equation}
    \lambda_1 = (a - b) + |L|b = a + (|L|-1)b,
\end{equation}
and the remaining eigenvalues (with eigenvectors in $\mathbf{1}_{|L|}^\perp$, multiplicity $|L|-1$) are
\begin{equation}
    \lambda_2 = \cdots = \lambda_{|L|} = a - b.
\end{equation}

Substituting the definitions of $a$ and $b$:
\begin{align}
\begin{split}
    a - b &= \frac{2h_E|\mathcal{E}|}{\Sigma_2} - \frac{2(1-h_E)|\mathcal{E}|}{N^2 - \Sigma_2} = 2|\mathcal{E}| \left( \frac{h_E}{\Sigma_2} - \frac{1-h_E}{N^2 - \Sigma_2} \right) \\
    &= 2|\mathcal{E}| \cdot \frac{h_E(N^2 - \Sigma_2) - (1-h_E)\Sigma_2}{\Sigma_2(N^2 - \Sigma_2)}.
\end{split}
\end{align}
Expanding the numerator:
\begin{equation}
    h_E(N^2 - \Sigma_2) - (1-h_E)\Sigma_2 = h_EN^2 - h_E\Sigma_2 - \Sigma_2 + h_E\Sigma_2 = h_EN^2 - \Sigma_2.
\end{equation}
Therefore:
\begin{equation}
    \lambda_2 = \cdots = \lambda_{|L|} = \frac{2|\mathcal{E}|(h_EN^2 - \Sigma_2)}{\Sigma_2(N^2 - \Sigma_2)}.
\end{equation}

\paragraph{Critical Threshold.}
The sign of $\lambda_2, \ldots, \lambda_{|L|}$ is determined by the numerator $(h_EN^2 - \Sigma_2)$:
\begin{equation}
    h_EN^2 - \Sigma_2 \gtreqqless 0 \quad \Longleftrightarrow \quad h_E \gtreqqless \frac{\Sigma_2}{N^2} = \frac{\sum_{k=1}^{|L|} n_k^2}{N^2}.
\end{equation}
Thus, the critical threshold is:
\begin{equation}
    h_E^* = \frac{\sum_{k=1}^{|L|} n_k^2}{N^2} = \sum_{k=1}^{|L|} \left(\frac{n_k}{N}\right)^2.
\end{equation}

\paragraph{Positivity of $\lambda_1$.}
We verify that $\lambda_1 > 0$ for all valid parameter ranges:
\begin{equation}
    \lambda_1 = a + (|L|-1)b = \frac{2h_E|\mathcal{E}|}{\Sigma_2} + \frac{2(|L|-1)(1-h_E)|\mathcal{E}|}{N^2 - \Sigma_2}.
\end{equation}
Since $h_E \in [0, 1]$, $|\mathcal{E}| > 0$, $\Sigma_2 > 0$, and $N^2 - \Sigma_2 = 2\sum_{i < j} n_i n_j > 0$ for $|L| \geq 2$, we have $\lambda_1 > 0$ always.

\paragraph{Signature Classification.}
Combining the above results:
\begin{itemize}
    \item If $h_E > h_E^*$: $\lambda_1 > 0$ and $\lambda_2 = \cdots = \lambda_{|L|} > 0$, so $s_{\mathcal{G}} = (|L|, 0, 0)$.
    \item If $h_E = h_E^*$: $\lambda_1 > 0$ and $\lambda_2 = \cdots = \lambda_{|L|} = 0$, so $s_{\mathcal{G}} = (1, |L|-1, 0)$.
    \item If $h_E < h_E^*$: $\lambda_1 > 0$ and $\lambda_2 = \cdots = \lambda_{|L|} < 0$, so $s_{\mathcal{G}} = (1, 0, |L|-1)$.
\end{itemize}

\paragraph{Reduction to Balanced Case.}
For balanced classes where $n_i = n$ for all $i \in \{1, \ldots, |L|\}$:
\begin{equation}
    h_E^* = \frac{\sum_{k=1}^{|L|} n^2}{(|L|n)^2} = \frac{|L|n^2}{|L|^2 n^2} = \frac{1}{|L|}.
\end{equation}

\paragraph{Probabilistic Interpretation.}
The critical threshold $h_E^* = \sum_{k=1}^{|L|} (n_k/N)^2$ equals the probability that two nodes chosen uniformly at random (with replacement) belong to the same class. This quantity provides a natural baseline:
\begin{itemize}
    \item $h_E > h_E^*$: Edges preferentially connect intra-class nodes relative to random chance (\emph{homophilic}).
    \item $h_E < h_E^*$: Edges preferentially connect inter-class nodes relative to random chance (\emph{heterophilic}).
\end{itemize}

\hfill $\square$

\subsection{Proof of Theorem \ref{theorem:refinement}} \label{AppendixA:Refinement}

Both homophily indices of a finite graph are rational: $h_E$ is a ratio of edge counts and $h_V$ is an average of ratios of integers. The construction below attains every rational number in $(0,1)$, so Theorem \ref{theorem:refinement} covers every value that $h_E$ or $h_V$ takes in $(0,1)$. We prove it in the following stronger form, which also prescribes the number of classes.
\begin{theorem}
For every rational $h \in (0,1)$ and every number of classes $|L| \geq 2$, there exists a graph $\mathcal{G} := (\mathcal{V}, \mathcal{E})$ with node labels $l:\mathcal{V} \to L$ such that $h_E = h_V = h$ and $S_{\mathcal{G}}$ has a negative eigenvalue.
\end{theorem}
\begin{proof}
We first record a sufficient condition that involves neither $h_E$ nor $h_V$. Suppose that some class $j$ has no intra-class edge and at least one edge to a class $i \neq j$, so that $s_{j,j} = 0$ and $s_{i,j} = s_{j,i} > 0$. For the standard basis vectors $e_i, e_j \in \mathbb{R}^{|L|}$ and $x = t e_i - e_j$ with $t > 0$,
\begin{equation}
    x^T S_{\mathcal{G}} x = t^2 s_{i,i} - 2t s_{i,j} + s_{j,j} = t \left( t s_{i,i} - 2 s_{i,j} \right),
\end{equation}
which is negative for every sufficiently small $t > 0$. The smallest eigenvalue of $S_{\mathcal{G}}$ is at most the Rayleigh quotient $x^T S_{\mathcal{G}} x / x^T x$, hence negative.

Write $h = p/q$ with integers $1 \leq p < q$, and let $k = |L|$ and $m = k-1$. Let $\mathcal{G}$ be the disjoint union of $mq$ edges: $mp$ of them join two nodes of class $1$, and each of the other $m(q-p) \geq k-1$ edges joins a node of class $1$ to a node of a class $j \in \{2, \dots, k\}$, with every such $j$ used at least once. Every class is nonempty, so $S_{\mathcal{G}}$ is well defined, and $h_E = mp/(mq) = h$. Every node has degree $1$: each of the $2mp$ endpoints of an intra-class edge has its only neighbor in its own class, and each of the other $2m(q-p)$ nodes has its only neighbor in another class, so $h_V = 2mp/(2mq) = h$. Class $2$ has no intra-class edge and an edge to class $1$, so $S_{\mathcal{G}}$ has a negative eigenvalue by the sufficient condition.

The signature is explicit. Class $1$ has $n_1 = m(q+p)$ nodes and $mp \geq 1$ intra-class edges, so $s_{1,1} > 0$. Each node of a class $j \geq 2$ has its single edge to class $1$, so $s_{j,1} = 1/n_1$ and $s_{j,j'} = 0$ for all $j, j' \geq 2$. Rows $2, \dots, k$ of $S_{\mathcal{G}}$ therefore all equal $(1/n_1, 0, \dots, 0)$, while row $1$ has the entry $1/n_1$ in columns $2, \dots, k$, so $S_{\mathcal{G}}$ has rank $2$. One of its two nonzero eigenvalues is negative and their sum is $\operatorname{tr} S_{\mathcal{G}} = s_{1,1} > 0$, so $s_{\mathcal{G}} = (1, |L|-2, 1)$. For $|L| = 2$,
\begin{equation}
    S_{\mathcal{G}} = \begin{pmatrix} s_{1,1} & \frac{1}{q+p} \\ \frac{1}{q+p} & 0 \end{pmatrix}, \qquad s_{1,1} > 0, \qquad \det S_{\mathcal{G}} = -\frac{1}{(q+p)^2} < 0.
\end{equation}
\end{proof}

\subsection{Proof and interpretations of Theorem \ref{theorem:main_theorem0}} \label{AppendixA:main_theorem0}

For a smooth map $f:\mathcal{M} \to \mathcal{L}$ between differentiable manifolds $\mathcal{M}$ and $\mathcal{L}$, the signature of the metric $g$ over $\mathcal{L}$ and that of the pullback $f^*g$ over $\mathcal{M}$ satisfy the following conditions:
\begin{align}
    \begin{split}
         \text{Number of positive eigenvalues of } (f^*g)_{p} &\leq  \text{Number of positive eigenvalues of } g_{f(p)} \\
         \text{Number of negative eigenvalues of } (f^*g)_{p} &\leq  \text{Number of negative eigenvalues of } g_{f(p)} \\
         \text{Number of zero eigenvalues of } (f^*g)_{p} &= \dim_{\mathbb{R}} \text{Ker}(T_p f) + \dim_{\mathbb{R}} \left( \text{Im}(T_p f) \cap \text{Im}(T_p f)^{\perp_g} \right)
    \end{split}
\end{align}
The second term in the last line is needed whenever $g$ is indefinite. A linear map can be injective and still pull an indefinite form back to a degenerate one: for $A = (1,1)^\top : \mathbb{R} \to \mathbb{R}^{1,1}$ with $g = \mathrm{diag}(1,-1)$ we have $\ker A = 0$ yet $A^\top g A = 0$, so the nullity is $1$. The image of $A$ is a null line, contained in its own $g$-orthogonal complement, which is exactly what the second term counts. When $g$ is positive definite that intersection is trivial and the identity reduces to the familiar $\dim \ker$. The two inequalities on the positive and negative counts hold regardless.

By Sylvester's law of inertia (\cref{sylvester}), $\phi^*I$ is positive-definite. Hence, for any pushforward $T_pf: T_p \mathcal{M} \to T_{f(p)} \mathcal{L}$, the eigenvalues of the pullback $f^*(\phi^*I)$ satisfy:
\begin{align}
     \text{Number of negative eigenvalues of } (f^*(\phi^*I))_{p} = 0
\end{align}
On the other hand, if $T_pf$ is surjective then $\text{Im}(T_pf) = T_{f(p)}\mathcal{L}$, so a negative direction of $S_{\mathcal{G}}$ has a preimage and the eigenvalues of the pullback $f^*S_{\mathcal{G}}$ satisfy:
\begin{align}
     \text{Number of negative eigenvalues of } (f^*S_{\mathcal{G}})_{p} > 0
\end{align}
Surjectivity is what this step needs, and it is strictly stronger than full rank when $\dim \mathcal{M} < \dim \mathcal{L}$: an injective map into a higher-dimensional indefinite space can miss every negative direction, as the example above shows. A linear map is surjective exactly when its rank equals $\dim \mathcal{L}$, which requires $\dim \mathcal{M} \geq \dim \mathcal{L}$ and is not implied by it. In our setting $T_pf$ is the projection $\mathbf{W} \in \mathbb{R}^{|L| \times h'}$ of \cref{pullback_metric}, and every configuration we report has $h' > |L|$. The matrices of rank below $|L|$ are the common zero set of the $|L| \times |L|$ minors, a proper algebraic subset of $\mathbb{R}^{|L| \times h'}$ of measure zero, so the hypothesis holds for almost every $\mathbf{W}$, and almost surely at the random initialization.
Therefore, there does not exist a diffeomorphism $\phi:\mathcal{L} \to \mathcal{L}$ which induces $(\phi \circ f)^* I = f^* S_{\mathcal{G}}$. This concludes the proof of Theorem \ref{theorem:main_theorem0}.

We can also deduce the following three observations from Theorem \ref{theorem:main_theorem0}.
\begin{itemize}
    \item Discretization of graph diffusion processes \cite{chamberlain2021grand}, which inherently assumes that the underlying topological space with graphs as subspaces is a Riemannian manifold, may have limited capabilities in analyzing relations among nodes of graphs whose $S_{\mathcal{G}}$ has a negative eigenvalue.
    \item Continuous node attributes of graphs may not be beneficial in classifying node labels of heterophilic graph data sets,
    implying that utilizing only the connectivity structure between node labels could be advantageous. Empirical evidence also supports this observation, as shown in \cite{Zhu21}.
    \item Using adjacency matrices other than positive-definite symmetric matrices, such as graph laplacians, may boost results in classifying node labels of heterophilic graph data sets. Examples include using adjacency matrices multiplied with cosine similarities of node labels \cite{yan2022} or using sheaf graph laplacian operator, a type of block matrix which is not necessarily positive-definite (but positive-definite as a block matrix) \cite{bodnar2022neural}.
\end{itemize}

\subsection{Topological structure on directed finite graphs in preparation for Definition \ref{def:tas_graph}} \label{AppendixA:Topology}

The parallel transport $\Gamma_{\gamma_{e_1,e_2}(t_0,s)}$ for each $t = t_0$ is taken with respect to the flat connection of $\mathbb{R}^h$, so it translates vectors, preserving both their directions and lengths. Because the metric $g = S$ is constant, this connection is compatible with $g$ even when $S$ is degenerate, and it is the Levi-Civita connection of $g$ when $S$ is non-degenerate. Since $\Gamma$ acts as the identity on $\mathbb{R}^h$ and $S$ is symmetric, $e_{\gamma_{e_1,e_2}}(\gamma_{e_1},\gamma_{e_2}) = e_{\gamma_{e_2,e_1}}(\gamma_{e_2},\gamma_{e_1})$, which proves the commutative property of tangent attention scores.

This subsection describes suitable topological structures to impose on directed finite graphs so that we can rigorously define the tangent attention score matrix of a graph. We treat graphs with $h$-dimensional node features as subspaces of an $h$-dimensional connected pseudo-Riemannian manifold. Without loss of generality, let $\mathcal{G} := (\mathcal{V}, \mathcal{E})$ be a finite directed graph with a node feature function $X:\mathcal{V} \to \mathbb{R}^h$. We may view undirected graphs as directed graphs by regarding each undirected edge between the node $v$ and $w$ as two reciprocal directed edges between them. Any graph $\mathcal{G}$ can be viewed as a 1-dimensional topological space with nodes as discrete points and edges as unit intervals $[0,1]$~\cite{park2026functorial,choi2022cycle}; see Chapter 0 of~\cite{Ha02} for further discussion. This presupposition allows us to extend a function of node features $X:\mathcal{V} \to \mathbb{R}^h$ to a continuous function $X_\gamma:\mathcal{G} \to \mathcal{M}$ for some $h$-dimensional pseudo-Riemannian manifold $(\mathcal{M},g)$, where each directed edge $e$ from the source node $v$ to the target node $w$, denoted as $e := (v,w)$, is represented by a smooth curve of our choice $\gamma_e: [0,1] \to \mathcal{M}$ satisfying the constraint that $\gamma_e(0) = X(v)$ and $\gamma_e(1) = X(w)$. Figure \ref{fig:manifold1} illustrates how one may regard a directed graph $\mathcal{G}$ as a subspace of $\mathcal{M}$.

\subsection{Detailed description and proof of Theorem \ref{theorem:main_theorem}} \label{AppendixA:main_theorem}

This subsection shows, with proof, which choice of extension $X_\gamma$ allows us to reconstruct a wide class of attention functions, including the dot-product attention from \cite{vaswani2017attention}, the single-layer feedforward linear layer from GAT~\cite{vel2018graph}, and the dynamic graph attention from GATv2~\cite{brody2022how}. Provided below is a detailed description of Theorem \ref{theorem:main_theorem}.
\begin{theorem}
Let $\mathcal{G} := (\mathcal{V}, \mathcal{E})$ be a finite directed complete graph with self-loops, along with an $h$-dimensional node feature function $X: \mathcal{V} \to \mathbb{R}^h$. Fix a symmetric matrix $S \in \mathbb{R}^{h \times h}$ which endows $\mathbb{R}^h$ a pseudo-metric structure.

Suppose that for each edge $e := (v,w) \in \mathcal{E}$, the curves $\gamma_e^{[i]}:[0,1] \to \mathcal{M}\cong\mathbb{R}^h$ are given by:
\begin{equation} \label{eqn:gamma_choice1}
    \gamma_e^{[1]}(t) := \begin{cases}
    (1-t) \cdot X(v) + t \cdot X(w) &\text{ if } v \neq w \\
    \sqrt{3} \left( a_v t(t-1) + \frac{1}{\sqrt{3}} \right) \cdot X(v) &\text{ if } v = w
    \end{cases}
\end{equation}
\begin{equation} \label{eqn:gamma_choice2}
    \gamma_e^{[2]}(t) := \begin{cases}
    (1-t^2) \cdot X(v) + t^3 \cdot X(w) &\text{ if } v \neq w \\
    X(v) + 6t(t-1)(5t-3) \cdot a_{v,1} + 10t(t-1)(2t-1) \cdot a_{v,2}  &\text{ if } v = w
    \end{cases}
\end{equation}
\begin{equation} \label{eqn:gamma_choice3}
    \gamma_e^{[3]}(t) := \begin{cases}
    (1-t) \cdot X(v) + t \cdot X(w) + t(t-1) \cdot a_{e} &\text{ if } v \neq w \\
    X(v) + 3t(t-1) \cdot a_{v,3}  &\text{ if } v = w
    \end{cases}
\end{equation}
for some fixed choices of set of scalars $\{a_v\}_{v\in \mathcal{V}}$ and $h$-dimensional real vectors $\{a_{e}\}_{e \in \mathcal{E}}, \{a_{v,i}\}_{v \in \mathcal{V}} \subset \mathbb{R}^h$. Then given a pair of edges $e_1 := (v_1,w_1), e_2 := (v_2,w_2)$, the $(e_1,e_2)$-th entries of the tangent attention score matrices $e_{\mathcal{G},\gamma^{[i]}}$ are equal to:
\begin{equation} \label{eqn:gamma_result1}
    \left(e_{\mathcal{G},\gamma^{[1]}}\right)_{e_1,e_2} := \begin{cases}
    \left( X(w_1) - X(v_1) \right)^T \cdot S \cdot \left( X(w_2) - X(v_2) \right) &\text{ if } v_1 \neq w_1, \; v_2 \neq w_2 \\
    0 &\text{ if } v_1 \neq w_1, \; v_2 = w_2 \\
    0 &\text{ if } v_1 = w_1, \; v_2 \neq w_2 \\
    a_{v_1}a_{v_2} X(v_1)^T \cdot S \cdot X(v_2) &\text{ if } v_1 = w_1, \; v_2 = w_2
    \end{cases}
\end{equation}
\begin{equation} \label{eqn:gamma_result2}
    \left(e_{\mathcal{G},\gamma^{[2]}}\right)_{e_1,e_2} := \begin{cases}
    \left( \frac{4}{3} X(v_1) - \frac{3}{2} X(w_1) \right)^T \cdot S \cdot X(v_2) \\ \hspace{20pt} - \left( \frac{3}{2} X(v_1) - \frac{9}{5} X(w_1) \right)^T \cdot S \cdot X(w_2) &\text{ if } v_1 \neq w_1, \; v_2 \neq w_2 \\
    X(v_1)^T \cdot S \cdot a_{v_2,1} + X(w_1)^T \cdot S \cdot a_{v_2,2} &\text{ if } v_1 \neq w_1, \; v_2 = w_2 \\
    a_{v_1,1}^T \cdot S \cdot X(v_2) + a_{v_1,2}^T \cdot S \cdot X(w_2) &\text{ if } v_1 = w_1, \; v_2 \neq w_2 \\
    48 a_{v_1,1}^T \cdot S \cdot a_{v_2,1} + 30 a_{v_1,1}^T \cdot S \cdot a_{v_2,2} 
    \\ \hspace{20pt} + 30 a_{v_1,2}^T \cdot S \cdot a_{v_2,1} + 20 a_{v_1,2}^T \cdot S \cdot a_{v_2,2} &\text{ if } v_1 = w_1, \; v_2 = w_2
    \end{cases}
\end{equation}
\begin{equation} \label{eqn:gamma_result3}
    \left(e_{\mathcal{G},\gamma^{[3]}}\right)_{e_1,e_2} := \begin{cases}
    (X(w_1) - X(v_1))^T \cdot S \cdot (X(w_2) - X(v_2))
    \\ \hspace{20pt} + \frac{1}{3} a_{e_1}^T \cdot S \cdot a_{e_2} &\text{ if } v_1 \neq w_1, \; v_2 \neq w_2 \\
    a_{e_1}^T \cdot S \cdot a_{v_2,3} &\text{ if } v_1 \neq w_1, \; v_2 = w_2 \\
    a_{v_1,3}^T \cdot S \cdot a_{e_2} &\text{ if } v_1 = w_1, \; v_2 \neq w_2 \\
    3 a_{v_1,3}^T \cdot S \cdot a_{v_2,3} &\text{ if } v_1 = w_1, \; v_2 = w_2
    \end{cases}
\end{equation}
\end{theorem}
\begin{proof}
We prove the statement of the theorem for the curve $\gamma_e^{[1]}:[0,1] \to \mathcal{M}$. The remaining cases follow analogously. Let $e_1 = (v_1,w_1), e_2=(v_2,w_2)$ be a pair of directed edges of a finite directed complete graph $\mathcal{G}$.

Suppose $v_1 \neq w_1$ and $v_2 \neq w_2$. Then we obtain that
\begin{align}
\begin{split}
    \gamma_{e_1}^{'[1]}(t) &= X(w_1) - X(v_1) \\
    \gamma_{e_2}^{'[1]}(t) &= X(w_2) - X(v_2)
\end{split}
\end{align}
Because both derivatives are constant functions with respect to the variable $t$,
\begin{equation}
    e_{\gamma_{e_1,e_2}}(\gamma_{e_1}^{[1]},\gamma_{e_2}^{[1]}) = \left( X(w_1) - X(v_1) \right)^T \cdot S \cdot \left( X(w_2) - X(v_2) \right).
\end{equation}
If $v_1 = w_1$ whereas $v_2 \neq w_2$, then
\begin{align}
\begin{split}
    \gamma_{e_1}^{'[1]}(t) &= \sqrt{3}a_{v_1}(2t-1) \cdot X(v_1) \\
    \gamma_{e_2}^{'[1]}(t) &= X(w_2) - X(v_2)
\end{split}
\end{align}
which leads to
\begin{equation}
    e_{\gamma_{e_1,e_2}}(\gamma_{e_1}^{[1]},\gamma_{e_2}^{[1]}) = \left( \int_0^1 \sqrt{3}a_{v_1}(2t-1) dt \right) \cdot \left( X(v_1)^T \cdot S \cdot \left( X(w_2) - X(v_2) \right) \right) = 0.
\end{equation}
Lastly, if $v_1 = w_1$ and $v_2 = w_2$, then
\begin{align}
\begin{split}
    \gamma_{e_1}^{'[1]}(t) &= \sqrt{3}a_{v_1}(2t-1) \cdot X(v_1) \\
    \gamma_{e_2}^{'[1]}(t) &= \sqrt{3}a_{v_2}(2t-1) \cdot X(v_2)
\end{split}
\end{align}
which gives
\begin{equation}
\begin{split}
    e_{\gamma_{e_1,e_2}}(\gamma_{e_1}^{[1]},\gamma_{e_2}^{[1]}) &= \left( \int_0^1 3a_{v_1}a_{v_2}(2t-1)^2 dt \right) \cdot \left( X(v_1)^T \cdot S \cdot X(v_2) \right) \\
    &= a_{v_1}a_{v_2}X(v_1)^T \cdot S \cdot X(v_2).
\end{split}
\end{equation}
\end{proof}
Theorem \ref{theorem:main_theorem} demonstrates that previously researched attention scores correspond to computing the tangent attention score between two directed edges where at least one of them is a self-loop at each node of $\mathcal{G}$. The dot product attention score from the encoder layer of attention-transformers \cite{vaswani2017attention} with identical query and key weight matrices is equal to the tangent attention score between two self-loops represented by $\gamma_e^{[1]}$, with $S = W^T W$ a symmetric matrix obtained from a product of invertible square weight matrix $W \in \mathbb{R}^{h \times h}$. With the same symmetric matrix $S$, the attention score of the single-layer feedforward linear layer in \cite{vel2018graph} can be obtained from the tangent attention score between a self-loop and a directed edge represented by $\gamma_e^{[2]}$. Lastly, by setting $S$ to be the identity matrix $I$, and by choosing the fixed vector $a_e$ as $a_e = \text{LeakyRelu} \left(W[X(v) || X(w)] \right)$ for each edge $e = (v,w) \in \mathcal{E}$ and every $a_{v,3}$ as the shared attention vector $\mathbf{a} \in \mathbb{R}^{h}$ of GATv2, where $[X(v) || X(w)] \in \mathbb{R}^{2h}$ is a concatenation of two node feature vectors, and $W \in \mathbb{R}^{h \times 2h}$ is a weight matrix, the tangent attention score between a self-loop and a directed edge represented by $\gamma_e^{[3]}$ is equal to the dynamic graph attention score constructed from \cite{brody2022how}.

\section{Geometric Framework of PRAT}
\label{PRATframework}

\begin{figure}[t]
    \centering
    \begin{tikzcd}[
        row sep=1.2em,
        column sep=1.5em,
        ampersand replacement=\&
    ]
        \& \& \& \& (\mathcal{L}, g_{\mathcal{L}}) \cong \mathbb{R}^{|L|} \& \& \\
        \& \& \& \& \& \& \\
        \mathcal{V} \arrow[r]
        \& \mathcal{M} \arrow[r, "\psi_{\text{enc}}"]
        \& \mathcal{H}^{(0)} \arrow[r] \arrow[rruu, "\varphi^{(0)}"]
        \& \cdots \arrow[r]
        \& \mathcal{H}^{(l)} \arrow[r] \arrow[uu, "\varphi^{(l)}"]
        \& \cdots \arrow[r]
        \& \mathcal{H}^{(l_{\text{max}})} \arrow[lluu, "\varphi^{(l_{\text{max}})}"]
        \\
        v \arrow[r, mapsto]
        \& X(v) \arrow[r, mapsto]
        \& h_v^{(0)} \arrow[r, mapsto]
        \& \cdots \arrow[r, mapsto]
        \& h_v^{(l)} \arrow[r, mapsto]
        \& \cdots \arrow[r, mapsto]
        \& h_v^{(l_{\text{max}})}
    \end{tikzcd}
    \caption{Geometric interpretation of PRAT. Each hidden manifold $\mathcal{H}^{(l)}$ is mapped to the common latent manifold $\mathcal{L}$ via $\varphi^{(l)}$.}
    \label{PRATdiagram}
\end{figure}

In \cref{PRATdiagram}, $\mathcal{H}^{(l)} \cong \mathbb{R}^{h'}$ denotes the hidden manifold at layer $l$, and $\mathcal{L} \cong \mathbb{R}^{|L|}$ is the common latent manifold corresponding to the node labels. The encoder $\psi_{\text{enc}}: \mathcal{M} \to \mathcal{H}^{(0)}$ calibrates node features to the initial hidden feature.

For each layer $l$, we posit the existence of a linear map $\varphi^{(l)}: \mathcal{H}^{(l)} \to \mathcal{L}$, which projects the hidden representation to the label space. While the final layer's map $\varphi^{(l_{\text{max}})}$ acts as the explicit readout function for classification, the intermediate maps $\varphi^{(l)}$ (for $l < l_{\text{max}}$) provide a critical geometric link: they allow us to transport the intrinsic geometric structure of the latent manifold back to the hidden manifolds.

Let $g_{\mathcal{L}}$ be the pseudo-metric tensor defined on $\mathcal{L}$, represented by a symmetric matrix $S^{|L|}$. To ensure that the hidden representations at layer $l$ respect the global structural prior of the latent space, we set the metric on $\mathcal{H}^{(l)}$ to the \textit{pullback metric} induced by $\varphi^{(l)}$. Hence, the pseudo-metric on $\mathcal{H}^{(l)}$ is determined by pulling back $g_{\mathcal{L}}$ via $\varphi^{(l)}$:
\begin{equation}
    g_{\mathcal{H}^{(l)}} = (\varphi^{(l)})^* g_{\mathcal{L}}.
\end{equation}

Considering the multi-head attention mechanism where $k$ denotes the head index, the projection $\varphi^{(l)}$ is parameterized by a weight matrix $\mathbf{W}_{\varphi}^{[l,k]}$. The induced metric tensor $\mathbf{D}^{[l,k]}$ for the $k$-th head at layer $l$ is then derived as:

\begin{align}
    \mathbf{D}^{[l,k]} = (\mathbf{W}_{\varphi}^{[l,k]})^\top S^{|L|} \mathbf{W}_{\varphi}^{[l,k]},
\end{align}

which is identical to the expression for $\mathbf{D}^{[l,k]}$ in \cref{pullback_metric}.

\section{Benchmark Datasets}
\label{benchmarkdatasets}

We use fifteen real-world benchmarks in two groups: (i) eight \emph{standard} benchmarks spanning homophilic and heterophilic regimes, and (ii) seven \emph{Platonov benchmarks}~\citep{platonov2023critical} consisting of filtered Wikipedia networks and five large-scale heterophilic datasets.

\paragraph{Standard benchmarks.}
\begin{description}
    \item[Citation Networks (Cora, Citeseer, PubMed)] Standard benchmark datasets representing citation relationships between scientific publications \citep{sen2008collective, yang2016revisiting}. Nodes correspond to documents, and edges denote citation links. For each node, the features are typically sparse bag-of-words vectors derived from the document text, and the label indicates the subject category of the paper.

    \item[Wikipedia Networks (Chameleon, Squirrel)] These are page-page networks derived from Wikipedia \citep{rozemberczki2021multi}. Nodes represent Wikipedia articles, and edges indicate mutual hyperlinks between them. Node features are informative nouns extracted from the corresponding article, and the node label represents the average monthly traffic of the webpage.

    \item[Actor Co-occurrence Network (Actor)] The Actor dataset (often referred to as Film) \citep{tang2009social} is a subgraph of a film-director-actor-writer network. Nodes represent actors, and edges signify co-occurrence on the same Wikipedia page. Node features are derived from keywords on the actor's Wikipedia page, and node labels are determined by the words found on the webpage.

    \item[Webpage Networks (Texas, Cornell)] Subsets of the WebKB dataset\footnote{\url{http://www.cs.cmu.edu/afs/cs.cmu.edu/project/theo-11/www/wwkb}}. Nodes represent university webpages, and edges represent hyperlinks between them. For each node, the features are bag-of-words representations of the webpage content, and the node label denotes the category of the webpage.
\end{description}

\paragraph{Platonov benchmarks~\citep{platonov2023critical}.}
\begin{description}
    \item[Filtered Wikipedia networks (Chameleon-filtered, Squirrel-filtered)] \citet{platonov2023critical} identified that 57--61\% of nodes in the original Chameleon and Squirrel datasets are exact duplicates with identical neighborhoods and labels, causing severe train--test leakage. The \emph{filtered} versions remove these duplicate nodes, yielding 890 and 2,223 nodes respectively. Node features and class definitions follow the original \citet{rozemberczki2021multi} construction. These filtered benchmarks are widely adopted in subsequent heterophily research as a more reliable proxy for the underlying task.

    \item[roman-empire] Each node corresponds to a non-unique word in the English Wikipedia article ``Roman Empire,'' and edges connect words that are syntactically dependent or sequential in the article. Node features are GloVe-300 embeddings of the word, and the 18-class label is the syntactic role assigned by spaCy. With $h_E = 0.05$, this is the most extreme heterophilic dataset in the suite.

    \item[amazon-ratings] A product co-purchase network from Amazon, where nodes are products (books, music CDs, DVDs, VHS tapes) and edges connect frequently co-purchased products. Node features are the mean of fastText embeddings of the product description, and the 5-class label is the average rating, binned into five ranges.

    \item[minesweeper] A synthetic $100 \times 100$ grid graph from the Minesweeper game: nodes are cells, edges connect each cell to its eight surrounding neighbors, and 20\% of cells are randomly designated as mines (the binary class). Node features are 7-dimensional one-hot encodings of the number of neighboring mines for non-mine cells (with mine cells masked); the resulting graph has very low-dimensional node features and a regular topological structure.

    \item[tolokers] A worker-collaboration network from the Toloka crowdsourcing platform, where nodes are workers and edges connect workers who co-completed at least one task. Node features include task performance and profile statistics, and the binary label indicates whether a worker was banned. With over 519K edges over 11.8K nodes, tolokers is the densest graph in the suite.

    \item[questions] A user-interaction network on the Yandex.Q question-answering site, where nodes are users and edges connect users who answered each other's questions during a one-year window. Node features are FastText embeddings of the user description, and the binary label indicates account activity at the end of the window. With 48,921 nodes, this is the largest graph in the suite.
\end{description}

\begin{figure}[t]
    \centering
    \includegraphics[width=\textwidth]{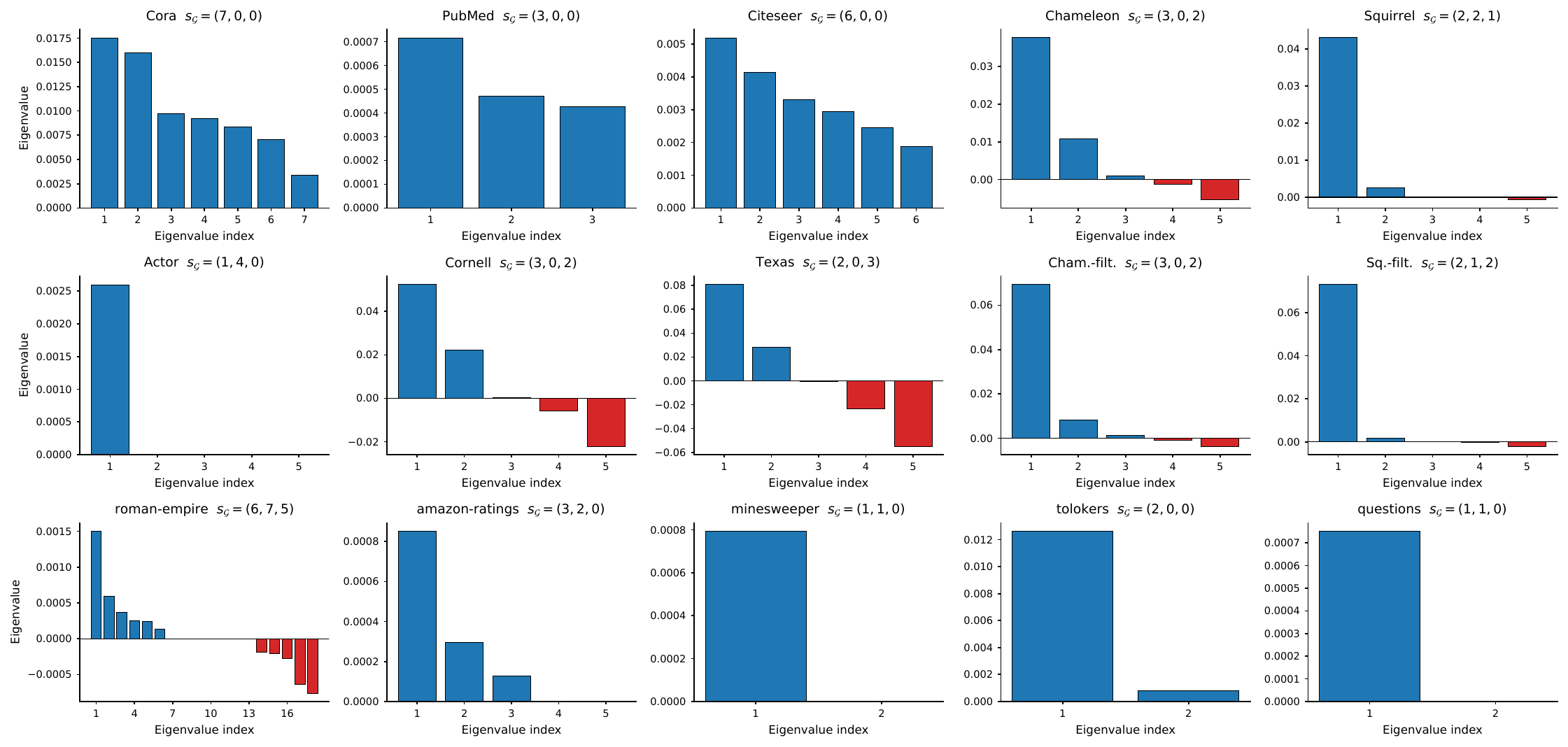}
    \caption{Eigenvalue distributions of the graph signature matrix of the full labeled graph (Definition~\ref{def:graph_signature_matrix}) for all fifteen benchmark datasets (eight standard + seven Platonov), which give the signatures of \cref{benchmark_stats} and \cref{benchmark_stats_platonov}. Eigenvalues with $|\lambda| < 1.0 \times 10^{-4}$ are rounded to zero for reporting.}
    \label{fig:graph_signature_appendix}
\end{figure}

\paragraph{Analysis on Graph Signature Index.} \cref{fig:graph_signature_appendix} illustrates the eigenvalue distributions of the graph signature matrices for all benchmark datasets, revealing distinct structural characteristics. Among the standard benchmarks, the \textbf{Actor} dataset stands out for its many zero eigenvalues.
%These zero eigenvalues denote a linear dependency among the rows of $S_{\mathcal{G}}$, implying that the underlying graph topology lacks sufficient discriminatory power along those dimensions. This explains the consistently lower GNN performance on Actor compared to other benchmarks (\cref{benchmark_results}).
Its signature $s_{\mathcal{G}} = (1, 4, 0)$ indicates connectivity of numerical rank one at the $10^{-4}$ threshold, where all 5 classes share nearly identical edge density patterns, with 4 out of 5 label-space dimensions carrying negligible discriminative signal. PRAT leverages this by learning $\Lambda$ values near zero for the corresponding dimensions during training, effectively projecting out uninformative directions and concentrating the metric on the single informative one. This ``feature selection via metric learning'' explains why PRAT achieves the best accuracy on Actor despite the absence of negative eigenvalues: the pseudo-Riemannian framework naturally accommodates zero eigenvalues, providing a mechanism to suppress noise that other methods aggregate indiscriminately.

\section{Synthetic Data Generation}
\label{syntheticdatagen}

Synthetic datasets are generated based on Model 1 proposed in Appendix B of \cite{ji2025rethinking}. According to this model, each node feature is initially sampled from a Bernoulli distribution, yielding binary values of 0 or 1. However, we observed that this formulation results in class labels being trivially predictable by GNNs, leading to near-perfect accuracy even in highly heterophilic regimes.

To address this limitation and make classification harder, we modify the original model's features. We add class-independent Gaussian noise of standard deviation $0.4$ to each node feature, thereby transforming the discrete binary features into continuous values and increasing the difficulty of class discrimination. The added noise prevents perfect predictions while preserving the synthetic dataset's underlying structure.

For graph construction, given a target number of edges, a desired homophily level $h_E \in [0,1]$, and a target graph signature index, we first compute the number of homophilic edges as $|\mathcal{E}_{homo}| = h_E \cdot |\mathcal{E}|$ and heterophilic edges as $|\mathcal{E}_{hetero}| = |\mathcal{E}| - |\mathcal{E}_{homo}|$. To ensure the resulting graph matches the specified signature index, these edge counts are strategically distributed across valid intra-class and inter-class pairs by a random search over at most $2{,}000$ symmetric candidates for a class-connectivity matrix that satisfies both the homophily and signature constraints. Homophilic edges are then generated by randomly sampling node pairs within the same class, while heterophilic edges are created by connecting nodes from different classes according to the determined connectivity matrix. Certain combinations of homophily levels and signature indices are mathematically incompatible; in such cases, synthetic data generation is restricted. This procedure precisely controls both the graph's homophily ratio and signature index, enabling systematic evaluation of model performance across diverse structural scenarios. Every graph is generated with seed $42$, the five random splits use seeds $42$ to $46$, and every model is trained with seed $42$.

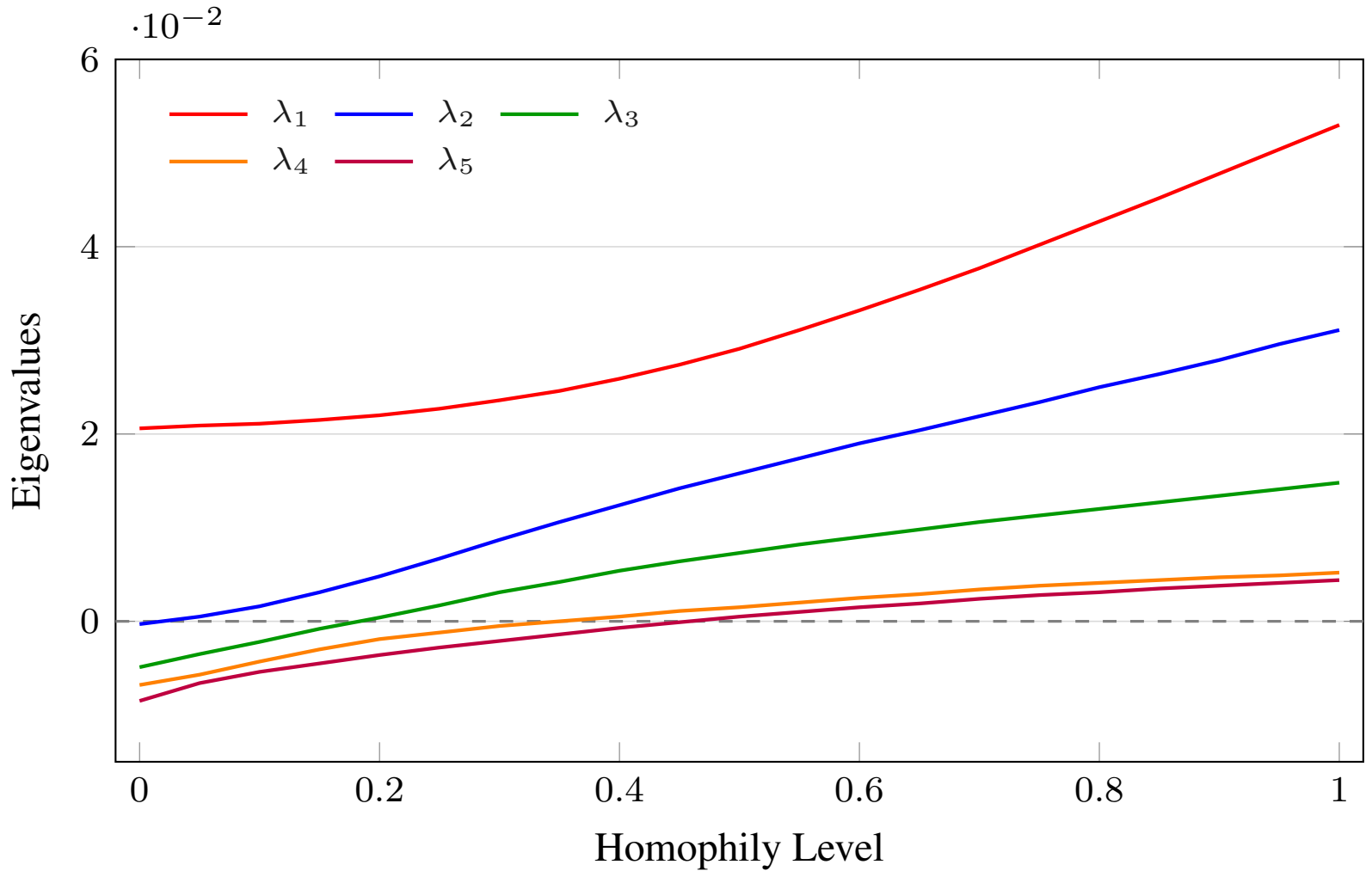
\begin{figure}[t]
    \centering
    \begin{tikzpicture}
        \begin{axis}[
            width=0.8\columnwidth,
            height=0.5\columnwidth,
            xlabel={Homophily Level},
            ylabel={Eigenvalues},
            xmin=-0.02, xmax=1.02,
            ymin=-0.015, ymax=0.06,
            xtick={0, 0.2, 0.4, 0.6, 0.8, 1.0},
            legend style={
                at={(0.03,0.97)},
                anchor=north west,
                legend columns=3,
                draw=none,
                fill=white,
                fill opacity=0.9,
                font=\scriptsize,
                column sep=0.1cm
            },
            ymajorgrids=true,
            grid style={line width=0.3pt, draw=gray!30},
            tick label style={font=\footnotesize},
            label style={font=\small},
        ]

        \draw[gray, dashed, line width=0.6pt] (axis cs:-0.02,0) -- (axis cs:1.02,0);

        \addplot[
            color=red,
            mark=*,
            mark size=0.0pt,
            line width=0.8pt,
        ] coordinates {
            (0.00,0.0206)(0.05,0.0209)(0.10,0.0211)(0.15,0.0215)(0.20,0.0220)
            (0.25,0.0227)(0.30,0.0236)(0.35,0.0246)(0.40,0.0259)(0.45,0.0274)
            (0.50,0.0291)(0.55,0.0311)(0.60,0.0332)(0.65,0.0354)(0.70,0.0377)
            (0.75,0.0402)(0.80,0.0427)(0.85,0.0452)(0.90,0.0478)(0.95,0.0504)(1.00,0.0530)
        };
        \addlegendentry{$\lambda_1$}

        \addplot[
            color=blue,
            mark=square,
            mark size=0.0pt,
            line width=0.8pt,
        ] coordinates {
            (0.00,-0.0003)(0.05,0.0005)(0.10,0.0016)(0.15,0.0031)(0.20,0.0048)
            (0.25,0.0067)(0.30,0.0087)(0.35,0.0106)(0.40,0.0124)(0.45,0.0142)
            (0.50,0.0158)(0.55,0.0174)(0.60,0.0190)(0.65,0.0204)(0.70,0.0219)
            (0.75,0.0234)(0.80,0.0250)(0.85,0.0264)(0.90,0.0279)(0.95,0.0296)(1.00,0.0311)
        };
        \addlegendentry{$\lambda_2$}

        \addplot[
            color=green!60!black,
            mark=triangle,
            mark size=0.0pt,
            line width=0.8pt,
        ] coordinates {
            (0.00,-0.0049)(0.05,-0.0035)(0.10,-0.0022)(0.15,-0.0008)(0.20,0.0004)
            (0.25,0.0017)(0.30,0.0031)(0.35,0.0042)(0.40,0.0054)(0.45,0.0064)
            (0.50,0.0073)(0.55,0.0082)(0.60,0.0090)(0.65,0.0098)(0.70,0.0106)
            (0.75,0.0113)(0.80,0.0120)(0.85,0.0127)(0.90,0.0134)(0.95,0.0141)(1.00,0.0148)
        };
        \addlegendentry{$\lambda_3$}

        \addplot[
            color=orange,
            mark=diamond,
            mark size=0.0pt,
            line width=0.8pt,
        ] coordinates {
            (0.00,-0.0068)(0.05,-0.0057)(0.10,-0.0043)(0.15,-0.0030)(0.20,-0.0019)
            (0.25,-0.0012)(0.30,-0.0005)(0.35,0.0000)(0.40,0.0005)(0.45,0.0011)
            (0.50,0.0015)(0.55,0.0020)(0.60,0.0025)(0.65,0.0029)(0.70,0.0034)
            (0.75,0.0038)(0.80,0.0041)(0.85,0.0044)(0.90,0.0047)(0.95,0.0049)(1.00,0.0052)
        };
        \addlegendentry{$\lambda_4$}

        \addplot[
            color=purple,
            mark=pentagon,
            mark size=0.0pt,
            line width=0.8pt,
        ] coordinates {
            (0.00,-0.0085)(0.05,-0.0066)(0.10,-0.0054)(0.15,-0.0045)(0.20,-0.0036)
            (0.25,-0.0028)(0.30,-0.0021)(0.35,-0.0014)(0.40,-0.0007)(0.45,-0.0001)
            (0.50,0.0005)(0.55,0.0010)(0.60,0.0015)(0.65,0.0019)(0.70,0.0024)
            (0.75,0.0028)(0.80,0.0031)(0.85,0.0035)(0.90,0.0038)(0.95,0.0041)(1.00,0.0044)
        };
        \addlegendentry{$\lambda_5$}

        \end{axis}
    \end{tikzpicture}
    \caption{Evolution of eigenvalues of graph signature matrix across varying homophily levels. As homophily increases, eigenvalues change from mixed signs to positive values, indicating a shift from pseudo-Riemannian to Riemannian geometry.}
    \label{fig:eigenvalue_evolution}
\end{figure}

\paragraph{Eigenvalue Evolution.} \cref{fig:eigenvalue_evolution} illustrates the evolution of eigenvalues in the graph signature matrix across synthetic datasets generated with varying homophily levels. At the heterophilic extreme ($h_E=0$), the graph signature index is $(1,0,4)$, indicating a predominance of negative eigenvalues. As $h_E$ increases, these eigenvalues exhibit a continuous upward trend, eventually transitioning to a strictly positive regime. This trajectory closely mirrors the sign evolution of the learned metric observed in \cref{fig:positive_ratio}. These findings provide empirical evidence that PRAT successfully leverages the underlying topological information regarding graph homophily.

\section{Computational Complexity Analysis}
\label{appendix:complexity}

We analyze the computational complexity of PRAT compared to other attention-based models. Let $N$ denote nodes, $E$ edges, $H$ hidden dimension, $C$ classes, $K$ attention heads, and $L$ layers. $H/K$ is the per-head dimension.

\subsection{Time Complexity}

\begin{theorem}[PRAT Layer Complexity]
The time complexity of a single PRAT layer is $\mathcal{O}(NH(H+KC) + E(H+KC))$.
\end{theorem}

\begin{proof}
PRAT has operations:
\begin{enumerate}[itemsep=0pt, topsep=0pt, leftmargin=*]
    \item Latent space projection ($\mathbf{X}\mathbf{W}$, $\mathbf{W} \in \mathbb{R}^{H \times KC}$): $\mathcal{O}(NHKC)$
    \item Value projection ($\mathbf{X}\mathbf{W}_V$, $\mathbf{W}_V \in \mathbb{R}^{H \times H}$): $\mathcal{O}(NH^2)$
    \item Attention scores (difference, square, weighted sum per edge): $\mathcal{O}(EKC)$
    \item Softmax and aggregation: $\mathcal{O}(EK + EH)$
\end{enumerate}
\end{proof}

\begin{theorem}[GAT/GATv2 Layer Complexity]
The time complexity of GAT and GATv2 layers is $\mathcal{O}(NH^2 + EH)$.
\end{theorem}

\begin{proof}
GAT computes $e_{ij} = \text{LeakyReLU}(\mathbf{a}^\top[\mathbf{W}\mathbf{h}_i \| \mathbf{W}\mathbf{h}_j])$:
\begin{enumerate}[itemsep=0pt, topsep=0pt, leftmargin=*]
    \item Feature transform ($\mathbf{X}\mathbf{W}$): $\mathcal{O}(NH^2)$
    \item Attention computation (concatenation + linear per edge): $\mathcal{O}(EK(H/K)) = \mathcal{O}(EH)$
    \item Softmax and aggregation: $\mathcal{O}(EK + EH)$
\end{enumerate}
GATv2 applies LeakyReLU before the attention vector but has identical complexity.

\end{proof}

\begin{theorem}[SDP Layer Complexity]
The time complexity of Scaled Dot-Product attention is $\mathcal{O}(3NH^2 + EH)$.
\end{theorem}

\begin{proof}
SDP computes $e_{ij} = (\mathbf{Q}_i \cdot \mathbf{K}_j)/\sqrt{d_k}$ with three projections:
\begin{enumerate}[itemsep=0pt, topsep=0pt, leftmargin=*]
    \item Query, key, and value projections: $\mathcal{O}(3NH^2)$
    \item Dot product per edge: $\mathcal{O}(EK(H/K)) = \mathcal{O}(EH)$
    \item Softmax and aggregation: $\mathcal{O}(EK + EH)$
\end{enumerate}
\end{proof}

\begin{table}[h]
\centering
\caption{Time complexity comparison per layer.}
\label{tab:time_complexity}
\begin{tabular}{@{}lc@{}}
\toprule
\textbf{Model} & \textbf{Complexity} \\
\midrule
PRAT & $\mathcal{O}(NH(H+KC) + E(H+KC))$ \\
GAT & $\mathcal{O}(NH^2 + EH)$ \\
GATv2 & $\mathcal{O}(NH^2 + EH)$ \\
SDP & $\mathcal{O}(3NH^2 + EH)$ \\
\bottomrule
\end{tabular}
\end{table}

\cref{tab:time_complexity} summarizes the time complexity of PRAT and attention-based models. PRAT shares asymptotic complexity $\mathcal{O}(NH^2 + EH)$ with GAT and GATv2 since $KC$ is typically small relative to $H$ (e.g., $K{=}4$, $C{=}5$, $H{=}64$ gives $KC{=}20 < 64$).

\subsection{Parameter Count}

For a model with encoder, $L$ attention layers, and decoder:

\begin{theorem}[Parameter Counts]
The total parameters for each model are:
\begin{align}
P_{\text{PRAT}} &= (F{+}1)H + L(H^2 + HKC + H + KC + K) + (H{+}1)C \\
P_{\text{GAT}} &= (F{+}1)H + L(H^2 + 3H) + (H{+}1)C \\
P_{\text{GATv2}} &= (F{+}1)H + L(H^2 + 3H) + (H{+}1)C \\
P_{\text{SDP}} &= (F{+}1)H + L(3H^2 + 3H) + (H{+}1)C
\end{align}
\end{theorem}

\begin{proof}
Per-layer parameters:
\begin{itemize}
    \item \textbf{PRAT}: $\mathbf{W} \in \mathbb{R}^{H \times KC}$ without bias, $\mathbf{W}_V \in \mathbb{R}^{H \times H}$ with bias, the $C$ diagonal entries of the learnable metric held for each of the $K$ heads, $\Lambda \in \mathbb{R}^{K \times C}$, self-loop scaling parameters $\{a\} \in \mathbb{R}^K$. Total: $HKC + H^2 + H + KC + K$.
    \item \textbf{GAT/GATv2}: $\mathbf{W} \in \mathbb{R}^{H \times H}$ and an output bias, with $\mathbf{a} \in \mathbb{R}^{2H}$ for GAT; GATv2 shares $\mathbf{W}$ between source and target and adds a bias to it, with $\mathbf{a} \in \mathbb{R}^{H}$. Total: $H^2 + 3H$.
    \item \textbf{SDP}: $\mathbf{W}_Q, \mathbf{W}_K, \mathbf{W}_V \in \mathbb{R}^{H \times H}$ each with bias. Total: $3H^2 + 3H$.
\end{itemize}
Single layer MLP encoder and decoder contribute $(F{+}1)H$ and $(H{+}1)C$, respectively, for all models.

\end{proof}

\begin{table}[h]
\centering
\caption{Parameter count per layer ($H{=}64$, $C{=}5$, $K{=}4$).}
\label{tab:param_count}
\begin{tabular}{@{}lcc@{}}
\toprule
\textbf{Model} & \textbf{Formula} & \textbf{Count} \\
\midrule
PRAT & $H^2 + HKC + H + KC + K$ & 5,464 \\
GAT & $H^2 + 3H$ & 4,288 \\
GATv2 & $H^2 + 3H$ & 4,288 \\
SDP & $3H^2 + 3H$ & 12,480 \\
\bottomrule
\end{tabular}
\end{table}

Table~\ref{tab:param_count} shows that PRAT has modestly more parameters than GAT and GATv2 because of its learnable metric components. Since $KC<H$ in practice, this overhead remains negligible. Moreover, PRAT is substantially more parameter-efficient than SDP attention, which requires separate query, key, and value projections.

\subsection{Measured cost} \label{app:cost}

The asymptotics above say that PRAT and GAT are in the same class. \cref{tab:cost} shows what that means in practice, measured on the two largest graphs with the same backbone, width, depth and number of steps, so that the only difference is the attention module.

\begin{table}[h]
    \caption{Training cost per step and peak memory, measured on one A100 80GB with identical backbone, width and depth.}
    \label{tab:cost}
    \centering
    \scriptsize
    \begin{tabular}{@{}llccc@{}}
        \toprule
        Dataset & Model & ms / step & peak memory (MiB) & parameters \\
        \midrule
        \multirow{3}{*}{tolokers (11{,}758 nodes)}
          & PRAT & 26.72 & 1084.4 & 336{,}689 \\
          & GAT  & 22.80 &  903.0 & 336{,}657 \\
          & GT   & 32.85 &  466.1 & 531{,}713 \\
        \midrule
        \multirow{3}{*}{questions (48{,}921 nodes)}
          & PRAT &  88.73 & 3791.2 & 2{,}830{,}201 \\
          & GAT  &  91.23 & 3696.9 & 2{,}830{,}121 \\
          & GT   & 175.17 & 5285.5 & 6{,}734{,}081 \\
        \bottomrule
    \end{tabular}
\end{table}

PRAT adds 80 parameters over GAT on questions, out of $2.83$ million, and 32 on tolokers. The step time is $0.97$ times GAT's on questions and $1.17$ on tolokers, and peak memory is $1.03$ and $1.20$ times GAT's. On questions, PRAT and GAT are indistinguishable at that scale: single-step timing has run-to-run spread of the same order, so their timings do not establish that PRAT is faster. GT, whose attention is also restricted to graph neighbors, costs $1.92$ times GAT on questions and needs $2.4$ times the parameters.

\section{Experimental Details}
\label{expdetail}

\paragraph{Signature estimation.} $S_{\mathcal{G},\text{train}}$ is Definition~\ref{def:graph_signature_matrix} applied to the subgraph induced by the training nodes, with self-loops removed: an edge counts only when both endpoints are training nodes, the class sizes in the denominator are training-node counts, and a class with no training node contributes a zero row and column. $\Lambda_{\text{init}}$ is computed from the unrounded eigenvalues, with $\text{std}(\cdot)$ the population standard deviation over the $|L|$ eigenvalues. When all eigenvalues are equal, so that this deviation is zero, the division is skipped and $\Lambda_{\text{init}} = \Lambda_{\mathcal{G},\text{train}}/\sqrt{|L|}$; the identity initialization of \cref{tab:labelfree} takes this path. The $10^{-4}$ rounding of \cref{fig:graph_signature_appendix} is applied to the unnormalized eigenvalues and only for reported inertias.

\subsection{Hyperparameters}
\label{hyperparameters}

\paragraph{Standard Benchmark Experiments.} For experiments on the eight standard benchmarks, the hyperparameters of every model (PRAT and baselines) were selected via a hyperparameter sweep over the search ranges and distributions listed in \cref{tab:hyperparams}, with the best configuration chosen by validation accuracy.

\begin{table}[h]
    \centering
    \caption{Hyperparameter search ranges used for the standard benchmark experiments.}
    \label{tab:hyperparams}
    \begin{tabular}{lcc}
        \toprule
        \textbf{Hyperparameter} & \textbf{Search Range} & \textbf{Distribution} \\
        \midrule
        \multicolumn{3}{l}{\textit{Model Architecture}} \\
        Hidden channels & 64 & Fixed \\
        Number of layers & \{1, 2, 3, 4\} & Discrete \\
        Attention heads & \{1, 2, 4, 8, 16\} & Discrete \\
        Input dropout & $[0.0, 0.9]$ & Uniform \\
        Output dropout & $[0.0, 0.9]$ & Uniform \\
        \midrule
        \multicolumn{3}{l}{\textit{Optimization}} \\
        Learning rate & $[10^{-4}, 10^{-1}]$ & Log-uniform \\
        Weight decay & $[10^{-5}, 10^{-2}]$ & Log-uniform \\
        Epochs & 1,000 & Fixed \\
        Patience & 100 & Fixed \\
        \bottomrule
    \end{tabular}
\end{table}

\paragraph{Platonov Benchmark Experiments.} For experiments on the seven \citet{platonov2023critical} benchmarks (filtered Chameleon/Squirrel and the five large-scale heterophilic datasets), PRAT's hyperparameters were selected via a sweep starting from the official PRAT-compatible defaults of~\citet{platonov2023critical} (hidden channels 512, layers $\in\{1,2,3,4,5\}$, lr $3{\times}10^{-5}$, dropout $0.2$, $1{,}000$ training steps),
%and following their validation-based selection protocol exactly: validation is evaluated at every optimization step, the best-validation checkpoint per split is retained,
and we report 10-split mean $\pm$ std on the official predefined splits. \cref{tab:platonov_configs} lists the selected configurations.

\begin{table}[h]
    \centering
    \caption{Selected PRAT configurations on the Platonov benchmarks, shared by \cref{tab:platonov_results} and the matched run of \cref{app:platonov_ablation}. Every configuration uses LayerNorm, AdamW and training seed $42$; warmup is a fraction of the training steps.}
    \label{tab:platonov_configs}
    \scriptsize
    \begin{tabular}{@{}lccccccccc@{}}
        \toprule
        Dataset & Hidden & Layers & Heads & FFN mult. & Dropout & Learning rate & Weight decay & Steps & Warmup \\
        \midrule
        Cham.-filt.    & 128 & 1 & 8 & 1   & 0.217 & $1.37{\times}10^{-4}$ & $3.28{\times}10^{-2}$ & 2{,}000 & 0.031 \\
        Sq.-filt.      & 512 & 1 & 8 & 0.5 & 0.627 & $1.59{\times}10^{-4}$ & $2.34{\times}10^{-2}$ & 500     & 0.042 \\
        roman-empire   & 512 & 5 & 8 & 1   & 0.2   & $3{\times}10^{-4}$    & 0                     & 1{,}000 & 0 \\
        amazon-ratings & 512 & 2 & 8 & 1   & 0.2   & $3{\times}10^{-4}$    & $10^{-3}$             & 2{,}000 & 0.1 \\
        minesweeper    & 128 & 4 & 8 & 0.5 & 0.1   & $7{\times}10^{-5}$    & $10^{-3}$             & 2{,}000 & 0 \\
        tolokers       & 128 & 4 & 4 & 2   & 0.1   & $3{\times}10^{-4}$    & 0                     & 2{,}000 & 0 \\
        questions      & 512 & 5 & 8 & 0.5 & 0.2   & $10^{-4}$             & $5{\times}10^{-4}$    & 1{,}000 & 0 \\
        \bottomrule
    \end{tabular}
\end{table}

% \begin{table}[h]
%     \centering
%     \caption{Hyperparameter search ranges for PRAT on the seven \citet{platonov2023critical} benchmarks. Architectural choices follow the official PRAT codebase and the search space used by~\citet{platonov2023critical} for their attention baselines.}
%     \label{tab:hyperparams_platonov}
%     \begin{tabular}{lcc}
%         \toprule
%         \textbf{Hyperparameter} & \textbf{Search Range} & \textbf{Distribution} \\
%         \midrule
%         \multicolumn{3}{l}{\textit{Model Architecture}} \\
%         Hidden dimension          & $\{64, 128, 256, 512\}$               & Discrete \\
%         Number of layers          & $\{1, 2, 3, 4, 5\}$                   & Discrete \\
%         Attention heads           & $\{1, 2, 4, 8\}$                      & Discrete \\
%         Hidden-dim multiplier     & $\{0.5, 1.0, 2.0\}$                   & Discrete \\
%         Normalization             & $\{$LayerNorm, BatchNorm$\}$          & Categorical \\
%         Dropout                   & $[0.0, 0.7]$                          & Uniform \\
%         \midrule
%         \multicolumn{3}{l}{\textit{Optimization}} \\
%         Learning rate             & $[10^{-5},\,5{\times}10^{-2}]$        & Log-uniform \\
%         Weight decay              & $[0,\,5{\times}10^{-2}]$              & Uniform \\
%         Number of training steps  & $\{500, 1000, 2000\}$                 & Discrete \\
%         Warmup proportion         & $[0.0, 0.1]$                          & Uniform \\
%         \bottomrule
%     \end{tabular}
% \end{table}

\paragraph{Synthetic Data Experiments.} When training models on synthetic data, we employed a fixed set of hyperparameters shown in \cref{tab:synthetic_config} across all synthetic datasets and models. This controlled setup ensures that any observed performance fluctuations are solely attributable to changes in the homophily level or graph signature index, thus facilitating a clear analysis of the behavioral trends and stability of each method.

\begin{table}[t]
    \centering
    \caption{Configuration for synthetic data experiments.}
    \label{tab:synthetic_config}
    \begin{tabular}{@{}lc@{\hspace{2em}}lc@{}}
        \toprule
        \multicolumn{2}{c}{\textbf{Dataset Configuration}} & \multicolumn{2}{c}{\textbf{Hyperparameters}} \\
        \cmidrule(r){1-2}\cmidrule(l){3-4}
        Nodes      & 1,000  & Hidden channels & 64 \\
        Edges    & 2,000  & Number of layers        & 2 \\
        Features   & 1,000  & Attention heads         & 8 \\
        Classes    & 5      & Input dropout           & 0.6 \\
        \multicolumn{2}{c}{}               & Output dropout           & 0.6 \\
        \multicolumn{2}{c}{}               & Learning rate           & 0.002 \\
        \multicolumn{2}{c}{}               & Weight decay            & 0.0005 \\
        \multicolumn{2}{c}{}               & Epochs              & 1,000 \\
        \multicolumn{2}{c}{}               & Patience                & 100 \\
        \bottomrule
    \end{tabular}
\end{table}

\subsection{Computing Resources}
\label{compute_resources}

All experiments we conducted for this paper were run on a single NVIDIA A100 80GB GPU. The baseline rows of \cref{tab:platonov_results} are quoted from~\citet{platonov2023critical}, and the Geom-GCN results from~\citet{yan2022}. No multi-GPU or multi-node training was used.
One full hyperparameter sweep per dataset takes between \(0.5\) and \(8\) GPU-hours depending on dataset size, after which the selected configuration is rerun across multiple splits for the reported test mean / standard deviation. Final 10-split evaluations on the Platonov benchmarks (\cref{tab:platonov_results}) require an additional \(\sim\) \(0.5\)--\(2\) GPU-hours per dataset.
The total compute budget across the project, including preliminary and discarded configurations not reported in the paper, is conservatively estimated at \(\sim\)200 A100-hours.

\subsection{Calibration of the Self-Loop Scaling Parameter}
\label{selfloopscaling}

From the tangent attention framework in Appendix~\ref{AppendixA:main_theorem}, self-loop scores involve a scaling coefficient applied to the quadratic form $h_v^\top \mathbf{D} h_v$. We derive a principled initialization for the self-loop scaling parameter $a$ by matching expected attention scores.

\paragraph{Pseudo-Label Representation.}
Assume hidden features encode noisy pseudo-labels in the latent space $\mathcal{L}$:
\begin{equation}
    \mathbf{W}h_v = e_{\ell(v)} + \epsilon_v, \quad \epsilon_v \sim \mathcal{N}(0, \sigma^2 I_{|L|}),
\end{equation}
where $\mathbf{W}$ is the weight matrix from \cref{pullback_metric}. $e_i \in \mathbb{R}^{|L|}$ is the one-hot encoding of label $i$ and $\sigma^2$ quantifies the noise level in the representations.

\paragraph{Matching Condition.}
Using the graph signature matrix $S_\mathcal{G}$ as the metric tensor and the proportional allocation model introduced in Appendix~\ref{AppendixA:critical_threshold}, matching the expected self-loop and edge scores yields, for $h_E > 0$:
\begin{equation}
    a(1 + \sigma^2|L|) = \frac{2(1-h_E)(h_E - h_E^*)}{h_E(1 - h_E^*)} + 2\sigma^2 |L|,
\end{equation}
where $h_E^* = \sum_{k=1}^{|L|} (n_k/N)^2$ is the critical homophily threshold from \cref{theorem:critical_threshold}, with $n_k$ denoting the number of nodes in class $k$. Solving for $a$:
\begin{equation}
\label{scorematching}
    a = \frac{\displaystyle\frac{2(1-h_E)(h_E - h_E^*)}{h_E(1 - h_E^*)} + 2\sigma^2 |L|}{1 + \sigma^2|L|}.
\end{equation}

\paragraph{Noise-Dominated Initialization.}
At the onset of training, before the network has learned label-discriminative representations, the hidden features are approximately uninformative about class labels, corresponding to the high-noise regime. In this regime, the structure-dependent term $\frac{2(1-h_E)(h_E - h_E^*)}{h_E(1 - h_E^*)}$ contributes negligibly compared to the noise-driven term $2\sigma^2|L|$. Consequently, \cref{scorematching} simplifies to:
\begin{equation}
    a_{\mathrm{init}} \simeq \frac{2\sigma^2 |L|}{\sigma^2|L|}=2.
\end{equation}
This initial value is independent of both the homophily level $h_E$ and the class distribution $\{n_k\}$, providing a universal, \textit{structure-agnostic} initialization.

\paragraph{Separation of Concerns.}
This derivation reveals a principled division of roles between PRAT's two learnable components:
\begin{itemize}
    \item \textbf{Metric tensor $\Lambda$}: Encodes the \emph{geometric structure} of the graph. Initialized from the graph signature eigenvalues $\Lambda_{\mathcal{G},\text{train}}$, it captures whether the underlying manifold is Riemannian or pseudo-Riemannian, directly reflecting the homophilic or heterophilic nature of the graph.
    \item \textbf{Self-loop scaling $a$}: Controls the \emph{signal-to-noise balance} in attention computation. Initialized at $2$ corresponding to uninformative representations, it adapts during training as the network learns discriminative features.
\end{itemize}

\section{Statistical reliability}
\label{app:reliability}

How the reported margins are tested, and how the spread divides between seeds and splits.

\subsection{Statistical significance} \label{app:significance}

Several margins in \cref{benchmark_results} and \cref{tab:platonov_results} are small relative to the split deviation, so we test them at two levels.

A Friedman test over all models of each table rejects equality of mean ranks on both suites, and PRAT holds the best mean rank on each: $2.63$ against $2.88$ for BernNet on the standard suite, with $\chi^2 = 54.46$, $p = 4.4 \times 10^{-6}$, and $1.86$ against $4.57$ for GT-sep on the Platonov suite, with $\chi^2 = 67.59$, $p = 1.2 \times 10^{-8}$. With eight and seven datasets the Nemenyi critical difference is $8.73$ and $8.72$ rank units at $\alpha = 0.05$, so this post-hoc test separates PRAT from the four and five lowest-ranked models, and the per-dataset tests carry the comparison with the strongest baselines.

Per dataset we use Welch's $t$ where only a mean and a standard deviation are available, and an exact paired Wilcoxon signed-rank test where per-split values exist on both sides. PRAT beats GAT-sep on roman-empire at $p = 3.9 \times 10^{-5}$ and on tolokers at $p = 8.2 \times 10^{-6}$. We re-ran GAT-sep, the strongest baseline on roman-empire, in the identical pipeline, reproducing its $88.75$ there, which allows a paired test on the shared splits that PRAT wins on all ten at $p = 0.002$.

In four comparisons PRAT has the higher mean but the difference is not significant at $\alpha = 0.05$, and we count these as ties: amazon-ratings against SAGE at $p = 0.55$, Citeseer against Geom-GCN at $p = 0.94$, filtered Chameleon against FAGCN at $p = 0.52$, and Actor against FAGCN at $p = 0.20$. The paired tests behind the ablation are given with \cref{tab:platonov_ablation} and \cref{tab:init_controls}.

\subsection{Variation across training seeds} \label{app:seeds}

The tables vary data splits with the training seed fixed, so that a split is reproducible. To check that fixing the seed hides no variance, we reran the ten-split evaluation on the five large Platonov datasets with three training seeds, in one matched run that reproduces \cref{tab:platonov_results} to within $0.11$ points.

\begin{table}[h]
    \caption{Ten-split mean under three training seeds, against the split-to-split deviation of the reported run.}
    \label{tab:seeds}
    \centering
    \scriptsize
    \begin{tabular}{@{}lcccccc@{}}
        \toprule
        Dataset & seed 42 & seed 7 & seed 123 & seed-to-seed sd & split-to-split sd \\
        \midrule
        roman-empire   & 90.16 & 90.08 & 89.98 & 0.09 & 0.56 \\
        amazon-ratings & 53.79 & 53.56 & 53.39 & 0.20 & 0.59 \\
        minesweeper    & 92.25 & 92.02 & 91.34 & 0.47 & 0.68 \\
        tolokers       & 84.93 & 85.09 & 85.14 & 0.11 & 0.65 \\
        questions      & 78.06 & 78.08 & 77.96 & 0.07 & 1.00 \\
        \bottomrule
    \end{tabular}
\end{table}

In \cref{tab:seeds} each ten-split mean moves by under one point across the three seeds. On a fixed split, the deviation across the three seeds, pooled over the ten splits, is $0.25$, $0.40$, $0.50$, $0.37$ and $0.47$ points in the order of the table, below the split-to-split deviation of a single run on every dataset, so the error bars we report capture the dominant source of variability.

\section{Graph Transformer and geometric baselines}
\label{app:outside}

\label{app:attention_geometric}

Four rows of \cref{benchmark_results} answer two further questions: how the neighbor-only Graph Transformer used by~\citet{platonov2023critical} behaves on these splits (GT and GT-sep), and how PRAT compares against models that already carry an indefinite or curved geometry (Q-GCN on a pseudo-hyperboloid, HAT on the Poincar\'e ball).

Two readings follow. First, GT is close on the two Wikipedia graphs, at $68.71$ against $68.73$ on Chameleon and $56.83$ against $56.96$ on Squirrel, and an exact paired test on the shared splits finds no significant difference ($p = 0.16$ and $p = 0.70$). GT differs from PRAT in its projections and layer architecture as well as in the metric, so it is not a geometry-matched control; the causal claim in this paper rests on the paired ablation of \cref{app:platonov_ablation}, not on this table.

Second, among the geometric models the pattern is cleaner. On the three homophilic graphs, where no negative direction can be exploited, everything sits within two points. On the heterophilic half Q-GCN trails by $7.61$ to $26.21$ points and HAT, curved but positive definite, trails on every dataset here. Q-GCN also trails PRAT by $3.87$ and $4.48$ points on filtered Chameleon and filtered Squirrel, at $38.82 \pm 2.28$ and $35.12 \pm 1.36$.

\section{The sign-conditional ablation}
\label{app:signablation}

What happens when the metric is forbidden to be indefinite, and what that does and does not license.

\subsection{Three-way ablation on the Platonov suite} \label{app:platonov_ablation}

\cref{ablation_result} runs the three-way ablation on the standard benchmarks. \cref{tab:platonov_ablation} repeats it on the Platonov suite and the filtered pair, inside one matched run whose PRAT rows reproduce \cref{tab:platonov_results} to within $0.24$ points. Because the three variants share every split, an exact paired Wilcoxon signed-rank test over the ten splits is available, which a comparison of means alone is not.

\begin{table}[h]
    \caption{Three-way ablation on the Platonov benchmarks, grouped as in \cref{tab:platonov_results}, with the graph signature index of each dataset. $p$ is an exact paired Wilcoxon signed-rank test of PRAT against $\text{PRAT}_{\text{pos}}$ over the ten shared splits; \textbf{bold} $p$ marks significance at $\alpha = 0.05$. Among the three variants the best mean is in \textbf{bold}, and where means tie the smaller deviation takes it.}
    \label{tab:platonov_ablation}
    \centering
    \scriptsize
    \begin{tabular}{@{}lccccc@{}}
        \toprule
        Dataset & $s_{\mathcal{G}}$ & PRAT & $\text{PRAT}_{\text{pos}}$ & $\text{PRAT}_{\text{fixed}}$ & $p$ \\
        \midrule
        Cham.-filt.    & (3,0,2) & \textbf{42.45\scriptsize{$\pm$2.20}} & 40.39\scriptsize{$\pm$2.73} & 42.00\scriptsize{$\pm$2.66} & \textbf{0.039} \\
        Sq.-filt.      & (2,1,2) & 39.58\scriptsize{$\pm$1.71} & 39.39\scriptsize{$\pm$2.16} & \textbf{40.22\scriptsize{$\pm$1.59}} & 0.770 \\
        \midrule
        roman-empire   & (6,7,5) & \textbf{90.16\scriptsize{$\pm$0.56}} & 87.56\scriptsize{$\pm$0.57} & 90.02\scriptsize{$\pm$0.51} & \textbf{0.002} \\
        amazon-ratings & (3,2,0) & \textbf{53.79\scriptsize{$\pm$0.59}} & 53.72\scriptsize{$\pm$0.48} & 53.70\scriptsize{$\pm$0.63} & 0.492 \\
        minesweeper    & (1,1,0) & 92.25\scriptsize{$\pm$0.68} & \textbf{92.28\scriptsize{$\pm$0.56}} & 92.09\scriptsize{$\pm$0.70} & 0.846 \\
        tolokers       & (2,0,0) & 84.93\scriptsize{$\pm$0.65} & 84.83\scriptsize{$\pm$0.72} & \textbf{84.93\scriptsize{$\pm$0.57}} & 0.131 \\
        questions      & (1,1,0) & \textbf{78.06\scriptsize{$\pm$1.00}} & 77.92\scriptsize{$\pm$1.21} & 77.91\scriptsize{$\pm$1.07} & 0.625 \\
        \bottomrule
    \end{tabular}
\end{table}

The gap between PRAT and $\text{PRAT}_{\text{pos}}$ appears only where the signature has negative eigenvalues, and there it is significant: $+2.60$ points on roman-empire on all ten splits, and $+2.06$ on filtered Chameleon on eight of ten. On the four datasets with $n_- = 0$ the two variants differ by at most $0.14$ points at $p \geq 0.13$, and the single nominal win for the constraint is minesweeper by $0.03$ points at $p = 0.85$. Filtered Squirrel is the boundary case. Its training-split spectrum is nearly flat beyond $\lambda_1 = 0.1375$, the next three eigenvalues carrying $3.2\%$, $0.2\%$ and $-0.7\%$ of it, so what the initialization supplies there is near-zero suppression rather than sign, and the positive-definite constraint costs $0.19$ points at $p = 0.77$. A constraint that mattered for generic reasons would not localize this way.

$\text{PRAT}_{\text{fixed}}$ matches PRAT on six of the seven, the exception being minesweeper by $0.16$ points at $p = 0.002$. Freezing $\Lambda$ freezes the inertia and not the metric, since $\mathbf{D} = (\mathbf{W}')^\top \Lambda \mathbf{W}'$ keeps adapting through $\mathbf{W}'$, so these rows say that on large graphs the initialization is what has to supply the spectrum while the pullback is still learned. Where the training-split signature is noisy, on the small graphs of \cref{ablation_result}, learning $\Lambda$ as well is what recovers $16.7$ points on Cornell.

\subsection{The signature as a diagnostic rather than a predictor} \label{app:correlation}

It is natural to ask whether $n_-$ predicts how much a signed or high-pass model gains. Rank-correlating $n_-$ from \cref{benchmark_stats} against what each such baseline of \cref{benchmark_results} gains over GCN is positive throughout, at $\rho = +0.23$ for FAGCN, $+0.33$ for GPR-GNN, $+0.60$ for BernNet and $+0.33$ for their mean gain; none is significant at eight datasets, and Squirrel is the exception, where the three models lose $2.82$ points on average despite $n_- = 1$.

We therefore present $s_{\mathcal{G}}$ as a diagnostic of the condition under which an indefinite mechanism should pay, not as a predictor of accuracy. The direct evidence for the diagnostic reading is the sign-conditional ablation of \cref{app:platonov_ablation}, which manipulates the admissible sign structure with everything else held fixed and finds the effect exactly where the condition holds. Predicting the gain is hard for measures of this kind in general: \citet{zheng2024missing} report their own composite homophily score correlating with the graph-aware advantage at $p \geq 0.11$ over thirty-one graphs.

\section{What the initialization has to supply}
\label{app:initialization}

Which part of the signature carries the effect, and whether ground-truth labels are needed to obtain it.

\subsection{What the initialization has to supply} \label{app:init_controls}

The ablation above shows that the freedom to be indefinite is worth something. It does not say which part of the signature carries it, the sign pattern or the magnitudes. \cref{tab:init_controls} separates the two with four controls that keep the architecture fixed and change the initial diagonal; \texttt{abs\_pos} also holds $\Lambda$ positive definite during training, as $\text{PRAT}_{\text{pos}}$ does.

\begin{table}[h]
    \caption{Initialization controls on the two indefinite benchmarks. The signature row is the reference: $\Delta$ is the loss against it and $p$ an exact paired Wilcoxon over ten shared splits, with \textbf{bold} $p$ marking significance at $\alpha = 0.05$. \texttt{abs\_pos} keeps the magnitudes and deletes the negative directions; \texttt{sign} keeps the signs at a uniform magnitude; \texttt{flip} keeps the magnitudes and inverts the signs; \texttt{random} draws an indefinite diagonal of the same inertia at the signature's scale.}
    \label{tab:init_controls}
    \centering
    \scriptsize
    \begin{tabular}{@{}lcccccc@{}}
        \toprule
        & \multicolumn{3}{c}{roman-empire (6,7,5)} & \multicolumn{3}{c}{Cham.-filt. (3,0,2)} \\
        \cmidrule(lr){2-4}\cmidrule(lr){5-7}
        Initialization & Accuracy & $\Delta$ & $p$ & Accuracy & $\Delta$ & $p$ \\
        \midrule
        signature (PRAT)   & 90.16\scriptsize{$\pm$0.56} & --   & --    & 42.45\scriptsize{$\pm$2.20} & --   & --    \\
        \texttt{abs\_pos}  & 87.91\scriptsize{$\pm$0.61} & 2.26 & \textbf{0.002} & 41.64\scriptsize{$\pm$2.50} & 0.81 & 0.275 \\
        \texttt{sign}      & 89.86\scriptsize{$\pm$0.62} & 0.30 & 0.065 & 39.90\scriptsize{$\pm$2.22} & 2.55 & \textbf{0.008} \\
        \texttt{flip}      & 89.96\scriptsize{$\pm$0.64} & 0.21 & 0.084 & 39.44\scriptsize{$\pm$3.20} & 3.01 & \textbf{0.027} \\
        \texttt{random} 1  & 89.53\scriptsize{$\pm$0.31} & 0.63 & \textbf{0.010} & 40.81\scriptsize{$\pm$2.77} & 1.63 & \textbf{0.037} \\
        \texttt{random} 2  & 90.16\scriptsize{$\pm$0.47} & 0.00 & 1.000 & 39.83\scriptsize{$\pm$2.40} & 2.62 & \textbf{0.037} \\
        \texttt{random} 3  & 89.58\scriptsize{$\pm$0.42} & 0.58 & \textbf{0.002} & 40.11\scriptsize{$\pm$3.52} & 2.34 & 0.055 \\
        \midrule
        identity           & 87.61\scriptsize{$\pm$0.69} & 2.55 & --    & 40.18\scriptsize{$\pm$2.39} & 2.27 & --    \\
        positive definite  & 87.56\scriptsize{$\pm$0.57} & 2.60 & \textbf{0.002} & 40.39\scriptsize{$\pm$2.73} & 2.06 & \textbf{0.039} \\
        \bottomrule
    \end{tabular}
\end{table}

The controls give different results on the two graphs, and the difference is informative. On roman-empire the only variant that loses decisively is \texttt{abs\_pos}, which keeps the magnitude profile and removes the negative directions: it costs $2.26$ points on all ten splits. Everything that retains negative directions, including sign-only, sign-flipped and all three random draws of matching inertia, stays within $0.7$ points. The reason is visible in the spectrum: eleven of roman-empire's eighteen training-split eigenvalues are below a sixth of $\lambda_1$ in magnitude, so the profile is close to flat and the only information the signature carries there is that negative directions exist. On filtered Chameleon the spectrum is not flat, its second eigenvalue carrying $10.8\%$ of the first against $3.2\%$ on filtered Squirrel, and there the specific values matter: every random draw of matching inertia loses $1.63$ to $2.62$ points and sign-only loses $2.55$.

An uninformed initialization is the other end of the scale. An identity initialization that is free to become indefinite but carries no signature information reaches $87.61$ on roman-empire, level with the positive-definite variant at $87.56$. Freedom alone therefore buys nothing; what has to be supplied is the negative directions and, where the spectrum is informative, their relative sizes as well.

The initialization also survives training.
\cref{fig:positive_ratio} reports the converged positive-eigenvalue ratio on synthetic graphs. On real graphs we logged $\Lambda$ throughout training. Between $97.2\%$ and $100\%$ of its entries keep their initial sign at convergence, the cosine to the initialization stays above $0.985$, and roman-empire's negative-entry fraction holds between $0.38$ and $0.42$ while the magnitudes adapt. Every sign change falls on an entry initialized from a training-split eigenvalue within $10^{-4}$ of zero, which the convention of \cref{fig:graph_signature_appendix} rounds to zero, so every entry whose training-split eigenvalue counts as positive or negative keeps its sign. On tolokers, whose signature is strictly positive, no entry becomes negative. The model does not drift back to a Riemannian solution when it is free to, which is what the diagnose-then-learn reading predicts.

\subsection{Estimating the signature without labels} \label{app:labelfree}

The signature is computed from training-node labels, but nothing in the forward pass reads a label. To test whether the labels are the essential ingredient or merely a convenient one, we re-estimate the signature from $k$-means pseudo-labels on node features, with the number of clusters set to the number of classes and no ground-truth label used in the estimate, and train as usual.

\begin{table}[h]
    \caption{Label-free initialization. The inertia is that of the eigenvalue vector each initialization is built from, before normalization, at the $10^{-4}$ rounding convention of \cref{fig:graph_signature_appendix}; the two signature estimates use the training nodes.}
    \label{tab:labelfree}
    \centering
    \scriptsize
    \begin{tabular}{@{}lcccc@{}}
        \toprule
        & \multicolumn{2}{c}{roman-empire} & \multicolumn{2}{c}{Cham.-filt.} \\
        \cmidrule(lr){2-3}\cmidrule(lr){4-5}
        Initialization & Accuracy & inertia & Accuracy & inertia \\
        \midrule
        label-based signature & \textbf{90.16\scriptsize{$\pm$0.56}} & (8,4,6) & 42.45\scriptsize{$\pm$2.20} & (3,0,2) \\
        $k$-means pseudo-labels & 90.10\scriptsize{$\pm$0.48} & (7,4,7) & \textbf{42.70\scriptsize{$\pm$2.00}} & (1,3,1) \\
        uninformed identity & 87.61\scriptsize{$\pm$0.69} & (18,0,0) & 40.18\scriptsize{$\pm$2.39} & (5,0,0) \\
        \bottomrule
    \end{tabular}
\end{table}

In \cref{tab:labelfree} recovery is complete on both indefinite benchmarks, while the uninformed initialization loses $2.55$ and $2.27$ points. The feature-only estimate is not a reconstruction of the label signature: its inertia is $(7,4,7)$ against $(8,4,6)$ on roman-empire and $(1,3,1)$ against $(3,0,2)$ on filtered Chameleon, and it still recovers the accuracy. Taken with \cref{app:init_controls}, the picture is that what has to be supplied is the presence of negative directions together with near-zero suppression, not a faithful copy of the label spectrum, and that node features alone can supply it. A complete self-supervised treatment, the training objective as well as the initialization, is outside the scope of this paper.

\section{Behavior under sparse and noisy labels}
\label{app:sparse}

How far the signature and the accuracy survive when training labels are few or wrong.

\subsection{Stability of the signature under label sparsity and noise} \label{app:stability}

\cref{tab:training_fraction} varies how many labels the model trains on. A separate question is how quickly the signature estimate itself degrades. We keep a random fraction $p$ of the training nodes, or flip a fraction $\eta$ of the training labels to a random other class, recompute the signature and report the mean sign agreement of its eigenvalues against the full-training-label estimate over 100 repeats.

\begin{table}[h]
    \caption{Mean eigenvalue sign agreement of the estimated signature against the full-training-label signature, over 100 repeats.}
    \label{tab:sig_stability}
    \centering
    \scriptsize
    \begin{tabular}{@{}lcccccc@{}}
        \toprule
        & \multicolumn{3}{c}{fraction of training nodes kept} & \multicolumn{3}{c}{fraction of training labels flipped} \\
        \cmidrule(lr){2-4}\cmidrule(lr){5-7}
        Dataset & $p{=}10\%$ & $p{=}25\%$ & $p{=}50\%$ & $\eta{=}10\%$ & $\eta{=}20\%$ & $\eta{=}30\%$ \\
        \midrule
        roman-empire   & 0.88 & 0.88 & 0.93 & 0.97 & 0.93 & 0.87 \\
        amazon-ratings & 0.74 & 0.85 & 0.91 & 1.00 & 0.87 & 0.80 \\
        minesweeper    & 0.69 & 0.82 & 0.98 & 1.00 & 1.00 & 1.00 \\
        tolokers       & 0.99 & 1.00 & 1.00 & 1.00 & 1.00 & 0.96 \\
        questions      & 0.93 & 1.00 & 1.00 & 1.00 & 1.00 & 1.00 \\
        \midrule
        Chameleon      & 0.92 & 0.93 & 0.91 & 0.94 & 0.93 & 0.92 \\
        Squirrel       & 0.93 & 0.92 & 0.93 & 0.95 & 0.93 & 0.90 \\
        Actor          & 0.45 & 0.53 & 0.67 & 0.86 & 0.83 & 0.84 \\
        Cornell        & 0.11 & 0.53 & 0.85 & 0.93 & 0.90 & 0.87 \\
        Texas          & 0.34 & 0.67 & 0.94 & 0.86 & 0.81 & 0.81 \\
        \bottomrule
    \end{tabular}
\end{table}

\cref{tab:sig_stability} reports the agreement. Large graphs are stable at modest label fractions. Actor's low agreement sits in near-zero eigenvalues at the rounding threshold, where the dominant directions keep their sign, and the 183-node WebKB graphs lose the signature below a quarter of an already small training set. That is exactly where \cref{tab:training_fraction} shows Cornell losing its lead, with all three models inside $0.6$ points at $10\%$, and it is why the learnable $\Lambda$ matters most on small graphs.

What matters in the end is accuracy. Training PRAT from the corrupted initialization, with training labels left clean and only the initialization degraded, costs at most $0.11$ points on roman-empire ($90.16 \to 90.05$ at $\eta = 10\%$) and $1.46$ on filtered Chameleon ($42.45 \to 40.99$ at $p = 25\%$), both inside the split deviation, and every corrupted variant stays at or above the positive-definite ablation.

\subsection{Training Fraction Sensitivity} \label{appendix:training_fraction}

We evaluate how PRAT's advantage depends on label availability by training on 10\%, 20\%, 40\%, 60\%, and 100\% of the original training split on Cornell, Texas, Actor, and Cora, using each model's best hyperparameters and 10 data splits.

\begin{table}[h]
    \caption{Accuracy (\%) at varying training fractions. Bold indicates the best model per dataset and fraction.}
    \centering
    \setlength{\tabcolsep}{3pt}
    \scriptsize
    \begin{tabular}{@{}llccccc@{}}
        \toprule
        {Dataset} & {Model} & {10\%} & {20\%} & {40\%} & {60\%} & {100\%} \\
        \midrule
        \multirow{3}{*}{Cornell}
        & PRAT & 44.05\scriptsize{$\pm$4.53} & \textbf{49.46\scriptsize{$\pm$9.37}} & \textbf{57.03\scriptsize{$\pm$8.59}} & \textbf{68.38\scriptsize{$\pm$6.40}} & \textbf{71.89\scriptsize{$\pm$4.71}} \\
        & GAT & 44.32\scriptsize{$\pm$5.16} & 41.62\scriptsize{$\pm$8.04} & 45.68\scriptsize{$\pm$9.00} & 46.76\scriptsize{$\pm$9.37} & 50.54\scriptsize{$\pm$7.55} \\
        & GATv2 & \textbf{44.59\scriptsize{$\pm$5.44}} & 41.35\scriptsize{$\pm$4.02} & 43.51\scriptsize{$\pm$3.91} & 46.49\scriptsize{$\pm$10.38} & 54.32\scriptsize{$\pm$9.70} \\
        \midrule
        \multirow{3}{*}{Texas}
        & PRAT & \textbf{63.24\scriptsize{$\pm$6.42}} & \textbf{67.57\scriptsize{$\pm$9.82}} & \textbf{74.05\scriptsize{$\pm$6.42}} & \textbf{78.65\scriptsize{$\pm$5.33}} & \textbf{84.32\scriptsize{$\pm$5.10}} \\
        & GAT & 60.54\scriptsize{$\pm$6.07} & 59.73\scriptsize{$\pm$9.16} & 61.62\scriptsize{$\pm$5.51} & 61.08\scriptsize{$\pm$7.27} & 64.59\scriptsize{$\pm$6.67} \\
        & GATv2 & 57.57\scriptsize{$\pm$6.95} & 60.54\scriptsize{$\pm$8.65} & 61.62\scriptsize{$\pm$4.80} & 62.16\scriptsize{$\pm$6.16} & 66.76\scriptsize{$\pm$6.63} \\
        \midrule
        \multirow{3}{*}{Actor}
        & PRAT & \textbf{27.40\scriptsize{$\pm$1.61}} & \textbf{28.93\scriptsize{$\pm$1.20}} & \textbf{33.72\scriptsize{$\pm$1.14}} & \textbf{35.61\scriptsize{$\pm$1.57}} & \textbf{37.26\scriptsize{$\pm$1.44}} \\
        & GAT & 26.32\scriptsize{$\pm$1.11} & 27.53\scriptsize{$\pm$1.75} & 28.79\scriptsize{$\pm$1.64} & 29.56\scriptsize{$\pm$1.31} & 31.42\scriptsize{$\pm$0.69} \\
        & GATv2 & 25.60\scriptsize{$\pm$1.33} & 27.46\scriptsize{$\pm$1.10} & 30.36\scriptsize{$\pm$1.39} & 32.28\scriptsize{$\pm$1.33} & 34.24\scriptsize{$\pm$1.10} \\
        \midrule
        \multirow{3}{*}{Cora}
        & PRAT & 74.24\scriptsize{$\pm$14.97} & 83.09\scriptsize{$\pm$2.12} & 85.27\scriptsize{$\pm$1.16} & 86.61\scriptsize{$\pm$1.14} & \textbf{88.42\scriptsize{$\pm$1.17}} \\
        & GAT & \textbf{80.11\scriptsize{$\pm$1.99}} & 82.98\scriptsize{$\pm$1.76} & 84.88\scriptsize{$\pm$1.17} & 86.54\scriptsize{$\pm$1.32} & 88.32\scriptsize{$\pm$1.09} \\
        & GATv2 & 78.88\scriptsize{$\pm$1.75} & \textbf{83.26\scriptsize{$\pm$1.98}} & \textbf{85.62\scriptsize{$\pm$1.26}} & \textbf{86.74\scriptsize{$\pm$1.06}} & 87.99\scriptsize{$\pm$0.93} \\
        \bottomrule
    \end{tabular}
    \label{tab:training_fraction}
\end{table}

On heterophilic datasets, PRAT consistently outperforms GAT and GATv2, but the gap narrows with fewer labels (Texas: +19.7pp at 100\% $\to$ +2.7pp at 10\%). This reveals the mechanism behind PRAT's advantage: as $S_{\mathcal{G}}$ becomes noisier with fewer labels, initialization quality degrades and performance approaches that of standard attention.

On homophilic Cora, all models perform comparably at fractions $\geq 20\%$, as expected, since the positive-definite structure of Cora's $S_\mathcal{G}$ initializes PRAT's metric as positive semi-definite, offering no additional geometric advantage over GAT/GATv2. At 10\%, PRAT shows higher variance (74.24±14.97\%) due to one split where $S_\mathcal{G}$ computed from $\sim$163 training nodes across 7 classes produces a severely noisy initialization. This is consistent with our analysis: the $S_\mathcal{G}$ quality degrades when training labels are extremely sparse relative to the number of classes, but performance recovers rapidly at 20\% (83.09±2.12\%).

\newpage
\section*{NeurIPS Paper Checklist}

\begin{enumerate}

\item {\bf Claims}
    \item[] Question: Do the main claims made in the abstract and introduction accurately reflect the paper's contributions and scope?
    \item[] Answer: \answerYes{} % Replace by \answerYes{}, \answerNo{}, or \answerNA{}.
    \item[] Justification: The main claims in the abstract and introduction are supported by the theoretical analysis in Section 2, the proposed PRAT method in Section 3, and empirical results in Section 4. The scope is limited to node classification on homophilic and heterophilic graph benchmarks.
    \item[] Guidelines:
    \begin{itemize}
        \item The answer \answerNA{} means that the abstract and introduction do not include the claims made in the paper.
        \item The abstract and/or introduction should clearly state the claims made, including the contributions made in the paper and important assumptions and limitations. A \answerNo{} or \answerNA{} answer to this question will not be perceived well by the reviewers. 
        \item The claims made should match theoretical and experimental results, and reflect how much the results can be expected to generalize to other settings. 
        \item It is fine to include aspirational goals as motivation as long as it is clear that these goals are not attained by the paper. 
    \end{itemize}

\item {\bf Limitations}
    \item[] Question: Does the paper discuss the limitations of the work performed by the authors?
    \item[] Answer: \answerYes{} % Replace by \answerYes{}, \answerNo{}, or \answerNA{}.
    \item[] Justification: The Limitations paragraph after the conclusion states four: the signature is initialized from training-node labels, which ties the method as presented to (semi-)supervised node classification; signature shift between training and test graphs is not studied; graph-level tasks are out of reach because $S_{\mathcal{G}}$ is undefined without node labels; and $\mathrm{rank}(\mathbf{D}) \leq |L|$ bounds the structure the metric can carry when classes are few. Appendix~\ref{app:labelfree} reports the label-free variant that weakens the first of these.
    \item[] Guidelines:
    \begin{itemize}
        \item The answer \answerNA{} means that the paper has no limitation while the answer \answerNo{} means that the paper has limitations, but those are not discussed in the paper. 
        \item The authors are encouraged to create a separate ``Limitations'' section in their paper.
        \item The paper should point out any strong assumptions and how robust the results are to violations of these assumptions (e.g., independence assumptions, noiseless settings, model well-specification, asymptotic approximations only holding locally). The authors should reflect on how these assumptions might be violated in practice and what the implications would be.
        \item The authors should reflect on the scope of the claims made, e.g., if the approach was only tested on a few datasets or with a few runs. In general, empirical results often depend on implicit assumptions, which should be articulated.
        \item The authors should reflect on the factors that influence the performance of the approach. For example, a facial recognition algorithm may perform poorly when image resolution is low or images are taken in low lighting. Or a speech-to-text system might not be used reliably to provide closed captions for online lectures because it fails to handle technical jargon.
        \item The authors should discuss the computational efficiency of the proposed algorithms and how they scale with dataset size.
        \item If applicable, the authors should discuss possible limitations of their approach to address problems of privacy and fairness.
        \item While the authors might fear that complete honesty about limitations might be used by reviewers as grounds for rejection, a worse outcome might be that reviewers discover limitations that aren't acknowledged in the paper. The authors should use their best judgment and recognize that individual actions in favor of transparency play an important role in developing norms that preserve the integrity of the community. Reviewers will be specifically instructed to not penalize honesty concerning limitations.
    \end{itemize}

\item {\bf Theory assumptions and proofs}
    \item[] Question: For each theoretical result, does the paper provide the full set of assumptions and a complete (and correct) proof?
    \item[] Answer: \answerYes{} % Replace by \answerYes{}, \answerNo{}, or \answerNA{}.
    \item[] Justification: The paper states the assumptions and formal results in Section 2, including Theorems 2.2–2.5, and provides detailed proofs in Appendix A.
    \item[] Guidelines:
    \begin{itemize}
        \item The answer \answerNA{} means that the paper does not include theoretical results. 
        \item All the theorems, formulas, and proofs in the paper should be numbered and cross-referenced.
        \item All assumptions should be clearly stated or referenced in the statement of any theorems.
        \item The proofs can either appear in the main paper or the supplemental material, but if they appear in the supplemental material, the authors are encouraged to provide a short proof sketch to provide intuition. 
        \item Inversely, any informal proof provided in the core of the paper should be complemented by formal proofs provided in appendix or supplemental material.
        \item Theorems and Lemmas that the proof relies upon should be properly referenced. 
    \end{itemize}

    \item {\bf Experimental result reproducibility}
    \item[] Question: Does the paper fully disclose all the information needed to reproduce the main experimental results of the paper to the extent that it affects the main claims and/or conclusions of the paper (regardless of whether the code and data are provided or not)?
    \item[] Answer: \answerYes{} % Replace by \answerYes{}, \answerNo{}, or \answerNA{}.
    \item[] Justification: Section 4.2 specifies the train/validation/test split, use of fixed and random splits, self-loop preprocessing, random seed control, and hyperparameter tuning protocol. Detailed configurations are provided in Appendix F.1.
    \item[] Guidelines:
    \begin{itemize}
        \item The answer \answerNA{} means that the paper does not include experiments.
        \item If the paper includes experiments, a \answerNo{} answer to this question will not be perceived well by the reviewers: Making the paper reproducible is important, regardless of whether the code and data are provided or not.
        \item If the contribution is a dataset and\slash or model, the authors should describe the steps taken to make their results reproducible or verifiable. 
        \item Depending on the contribution, reproducibility can be accomplished in various ways. For example, if the contribution is a novel architecture, describing the architecture fully might suffice, or if the contribution is a specific model and empirical evaluation, it may be necessary to either make it possible for others to replicate the model with the same dataset, or provide access to the model. In general. releasing code and data is often one good way to accomplish this, but reproducibility can also be provided via detailed instructions for how to replicate the results, access to a hosted model (e.g., in the case of a large language model), releasing of a model checkpoint, or other means that are appropriate to the research performed.
        \item While NeurIPS does not require releasing code, the conference does require all submissions to provide some reasonable avenue for reproducibility, which may depend on the nature of the contribution. For example
        \begin{enumerate}
            \item If the contribution is primarily a new algorithm, the paper should make it clear how to reproduce that algorithm.
            \item If the contribution is primarily a new model architecture, the paper should describe the architecture clearly and fully.
            \item If the contribution is a new model (e.g., a large language model), then there should either be a way to access this model for reproducing the results or a way to reproduce the model (e.g., with an open-source dataset or instructions for how to construct the dataset).
            \item We recognize that reproducibility may be tricky in some cases, in which case authors are welcome to describe the particular way they provide for reproducibility. In the case of closed-source models, it may be that access to the model is limited in some way (e.g., to registered users), but it should be possible for other researchers to have some path to reproducing or verifying the results.
        \end{enumerate}
    \end{itemize}

\item {\bf Open access to data and code}
    \item[] Question: Does the paper provide open access to the data and code, with sufficient instructions to faithfully reproduce the main experimental results, as described in supplemental material?
    \item[] Answer: \answerYes{} % Replace by \answerYes{}, \answerNo{}, or \answerNA{}.
    \item[] Justification: We provide a repository at \url{https://github.com/voltwin-dev/PRAT_NeurIPS_2026}, containing the implementation of PRAT, scripts for reproducing the experiments, configuration files, license information, and instructions for the reported benchmarks. The released code is provided for research and non-commercial use only due to patent-related restrictions, while the datasets used in this work are publicly available benchmark datasets and cited in Section 4.1.
    \item[] Guidelines:
    \begin{itemize}
        \item The answer \answerNA{} means that paper does not include experiments requiring code.
        \item Please see the NeurIPS code and data submission guidelines (\url{https://neurips.cc/public/guides/CodeSubmissionPolicy}) for more details.
        \item While we encourage the release of code and data, we understand that this might not be possible, so \answerNo{} is an acceptable answer. Papers cannot be rejected simply for not including code, unless this is central to the contribution (e.g., for a new open-source benchmark).
        \item The instructions should contain the exact command and environment needed to run to reproduce the results. See the NeurIPS code and data submission guidelines (\url{https://neurips.cc/public/guides/CodeSubmissionPolicy}) for more details.
        \item The authors should provide instructions on data access and preparation, including how to access the raw data, preprocessed data, intermediate data, and generated data, etc.
        \item The authors should provide scripts to reproduce all experimental results for the new proposed method and baselines. If only a subset of experiments are reproducible, they should state which ones are omitted from the script and why.
        \item At submission time, to preserve anonymity, the authors should release anonymized versions (if applicable).
        \item Providing as much information as possible in supplemental material (appended to the paper) is recommended, but including URLs to data and code is permitted.
    \end{itemize}

\item {\bf Experimental setting/details}
    \item[] Question: Does the paper specify all the training and test details (e.g., data splits, hyperparameters, how they were chosen, type of optimizer) necessary to understand the results?
    \item[] Answer: \answerYes{} % Replace by \answerYes{}, \answerNo{}, or \answerNA{}.
    \item[] Justification: The paper specifies dataset splits, preprocessing, random seeds, hyperparameter tuning, and evaluation metrics in Section 4.2, with additional configuration details in Appendix F.1.
    \item[] Guidelines:
    \begin{itemize}
        \item The answer \answerNA{} means that the paper does not include experiments.
        \item The experimental setting should be presented in the core of the paper to a level of detail that is necessary to appreciate the results and make sense of them.
        \item The full details can be provided either with the code, in appendix, or as supplemental material.
    \end{itemize}

\item {\bf Experiment statistical significance}
    \item[] Question: Does the paper report error bars suitably and correctly defined or other appropriate information about the statistical significance of the experiments?
    \item[] Answer: \answerYes{} % Replace by \answerYes{}, \answerNo{}, or \answerNA{}.
    \item[] Justification: All benchmark, ablation and appendix tables report the mean over the ten fixed evaluation splits with the standard deviation across those splits, so the error bars capture split variability at a fixed seed. Appendix~\ref{app:significance} adds a Friedman test across models on each suite and, per dataset, Welch's $t$-test and the exact paired Wilcoxon signed-rank test over the ten splits, and reports which comparisons are ties. Appendix~\ref{app:seeds} separates seed variability from split variability by repeating the ten-split evaluation under three training seeds.
    \item[] Guidelines:
    \begin{itemize}
        \item The answer \answerNA{} means that the paper does not include experiments.
        \item The authors should answer \answerYes{} if the results are accompanied by error bars, confidence intervals, or statistical significance tests, at least for the experiments that support the main claims of the paper.
        \item The factors of variability that the error bars are capturing should be clearly stated (for example, train/test split, initialization, random drawing of some parameter, or overall run with given experimental conditions).
        \item The method for calculating the error bars should be explained (closed form formula, call to a library function, bootstrap, etc.)
        \item The assumptions made should be given (e.g., Normally distributed errors).
        \item It should be clear whether the error bar is the standard deviation or the standard error of the mean.
        \item It is OK to report 1-sigma error bars, but one should state it. The authors should preferably report a 2-sigma error bar than state that they have a 96\% CI, if the hypothesis of Normality of errors is not verified.
        \item For asymmetric distributions, the authors should be careful not to show in tables or figures symmetric error bars that would yield results that are out of range (e.g., negative error rates).
        \item If error bars are reported in tables or plots, the authors should explain in the text how they were calculated and reference the corresponding figures or tables in the text.
    \end{itemize}

\item {\bf Experiments compute resources}
    \item[] Question: For each experiment, does the paper provide sufficient information on the computer resources (type of compute workers, memory, time of execution) needed to reproduce the experiments?
    \item[] Answer: \answerYes{}
    \item[] Justification: All experiments we conducted were run on a single NVIDIA A100 80GB GPU; the used computing resources are reported in Appendix~\ref{compute_resources}.
    \item[] Guidelines:
    \begin{itemize}
        \item The answer \answerNA{} means that the paper does not include experiments.
        \item The paper should indicate the type of compute workers CPU or GPU, internal cluster, or cloud provider, including relevant memory and storage.
        \item The paper should provide the amount of compute required for each of the individual experimental runs as well as estimate the total compute. 
        \item The paper should disclose whether the full research project required more compute than the experiments reported in the paper (e.g., preliminary or failed experiments that didn't make it into the paper). 
    \end{itemize}
    
\item {\bf Code of ethics}
    \item[] Question: Does the research conducted in the paper conform, in every respect, with the NeurIPS Code of Ethics \url{https://neurips.cc/public/EthicsGuidelines}?
    \item[] Answer: \answerYes{} % Replace by \answerYes{}, \answerNo{}, or \answerNA{}.
    \item[] Justification: The research appears to conform to the NeurIPS Code of Ethics and does not involve human subjects, private data, or high-risk deployment.
    \item[] Guidelines:
    \begin{itemize}
        \item The answer \answerNA{} means that the authors have not reviewed the NeurIPS Code of Ethics.
        \item If the authors answer \answerNo, they should explain the special circumstances that require a deviation from the Code of Ethics.
        \item The authors should make sure to preserve anonymity (e.g., if there is a special consideration due to laws or regulations in their jurisdiction).
    \end{itemize}

\item {\bf Broader impacts}
    \item[] Question: Does the paper discuss both potential positive societal impacts and negative societal impacts of the work performed?
    \item[] Answer: \answerNo{} % Replace by \answerYes{}, \answerNo{}, or \answerNA{}.
    \item[] Justification: The paper does not include a separate discussion of societal impacts. PRAT is a foundational method for node classification on graphs and is not tied to a particular application or deployment, and we do not see a direct path from it to negative applications.
    \item[] Guidelines:
    \begin{itemize}
        \item The answer \answerNA{} means that there is no societal impact of the work performed.
        \item If the authors answer \answerNA{} or \answerNo, they should explain why their work has no societal impact or why the paper does not address societal impact.
        \item Examples of negative societal impacts include potential malicious or unintended uses (e.g., disinformation, generating fake profiles, surveillance), fairness considerations (e.g., deployment of technologies that could make decisions that unfairly impact specific groups), privacy considerations, and security considerations.
        \item The conference expects that many papers will be foundational research and not tied to particular applications, let alone deployments. However, if there is a direct path to any negative applications, the authors should point it out. For example, it is legitimate to point out that an improvement in the quality of generative models could be used to generate Deepfakes for disinformation. On the other hand, it is not needed to point out that a generic algorithm for optimizing neural networks could enable people to train models that generate Deepfakes faster.
        \item The authors should consider possible harms that could arise when the technology is being used as intended and functioning correctly, harms that could arise when the technology is being used as intended but gives incorrect results, and harms following from (intentional or unintentional) misuse of the technology.
        \item If there are negative societal impacts, the authors could also discuss possible mitigation strategies (e.g., gated release of models, providing defenses in addition to attacks, mechanisms for monitoring misuse, mechanisms to monitor how a system learns from feedback over time, improving the efficiency and accessibility of ML).
    \end{itemize}
    
\item {\bf Safeguards}
    \item[] Question: Does the paper describe safeguards that have been put in place for responsible release of data or models that have a high risk for misuse (e.g., pre-trained language models, image generators, or scraped datasets)?
    \item[] Answer: \answerNA{} % Replace by \answerYes{}, \answerNo{}, or \answerNA{}.
    \item[] Justification: The paper does not release a high-risk pretrained model, scraped dataset, or system with a clear high-risk misuse scenario requiring special safeguards.
    \item[] Guidelines:
    \begin{itemize}
        \item The answer \answerNA{} means that the paper poses no such risks.
        \item Released models that have a high risk for misuse or dual-use should be released with necessary safeguards to allow for controlled use of the model, for example by requiring that users adhere to usage guidelines or restrictions to access the model or implementing safety filters. 
        \item Datasets that have been scraped from the Internet could pose safety risks. The authors should describe how they avoided releasing unsafe images.
        \item We recognize that providing effective safeguards is challenging, and many papers do not require this, but we encourage authors to take this into account and make a best faith effort.
    \end{itemize}

\item {\bf Licenses for existing assets}
    \item[] Question: Are the creators or original owners of assets (e.g., code, data, models), used in the paper, properly credited and are the license and terms of use explicitly mentioned and properly respected?
    \item[] Answer: \answerYes{} % Replace by \answerYes{}, \answerNo{}, or \answerNA{}.
    \item[] Justification: The paper cites the original sources of the benchmark datasets and baseline methods used in the experiments. The repository documents the licenses and terms of use for the datasets and baseline code assets, and states that the released PRAT implementation is provided under a research-only, non-commercial license due to patent-related restrictions.
    \item[] Guidelines:
    \begin{itemize}
        \item The answer \answerNA{} means that the paper does not use existing assets.
        \item The authors should cite the original paper that produced the code package or dataset.
        \item The authors should state which version of the asset is used and, if possible, include a URL.
        \item The name of the license (e.g., CC-BY 4.0) should be included for each asset.
        \item For scraped data from a particular source (e.g., website), the copyright and terms of service of that source should be provided.
        \item If assets are released, the license, copyright information, and terms of use in the package should be provided. For popular datasets, \url{paperswithcode.com/datasets} has curated licenses for some datasets. Their licensing guide can help determine the license of a dataset.
        \item For existing datasets that are re-packaged, both the original license and the license of the derived asset (if it has changed) should be provided.
        \item If this information is not available online, the authors are encouraged to reach out to the asset's creators.
    \end{itemize}

\item {\bf New assets}
    \item[] Question: Are new assets introduced in the paper well documented and is the documentation provided alongside the assets?
    \item[] Answer: \answerYes{} % Replace by \answerYes{}, \answerNo{}, or \answerNA{}.
    \item[] Justification:  We release an implementation of PRAT with documentation, experiment scripts, configuration files, license information, and instructions for reproducing the reported results. The released PRAT code is made available for research and non-commercial use only due to patent-related restrictions.
    \item[] Guidelines:
    \begin{itemize}
        \item The answer \answerNA{} means that the paper does not release new assets.
        \item Researchers should communicate the details of the dataset\slash code\slash model as part of their submissions via structured templates. This includes details about training, license, limitations, etc. 
        \item The paper should discuss whether and how consent was obtained from people whose asset is used.
        \item At submission time, remember to anonymize your assets (if applicable). You can either create an anonymized URL or include an anonymized zip file.
    \end{itemize}

\item {\bf Crowdsourcing and research with human subjects}
    \item[] Question: For crowdsourcing experiments and research with human subjects, does the paper include the full text of instructions given to participants and screenshots, if applicable, as well as details about compensation (if any)? 
    \item[] Answer: \answerNA{} % Replace by \answerYes{}, \answerNo{}, or \answerNA{}.
    \item[] Justification: The paper does not involve crowdsourcing experiments or research with human subjects.
    \item[] Guidelines:
    \begin{itemize}
        \item The answer \answerNA{} means that the paper does not involve crowdsourcing nor research with human subjects.
        \item Including this information in the supplemental material is fine, but if the main contribution of the paper involves human subjects, then as much detail as possible should be included in the main paper. 
        \item According to the NeurIPS Code of Ethics, workers involved in data collection, curation, or other labor should be paid at least the minimum wage in the country of the data collector. 
    \end{itemize}

\item {\bf Institutional review board (IRB) approvals or equivalent for research with human subjects}
    \item[] Question: Does the paper describe potential risks incurred by study participants, whether such risks were disclosed to the subjects, and whether Institutional Review Board (IRB) approvals (or an equivalent approval/review based on the requirements of your country or institution) were obtained?
    \item[] Answer: \answerNA{} % Replace by \answerYes{}, \answerNo{}, or \answerNA{}.
    \item[] Justification: The paper does not involve crowdsourcing experiments or research with human subjects, so IRB approval or equivalent review is not applicable.
    \item[] Guidelines:
    \begin{itemize}
        \item The answer \answerNA{} means that the paper does not involve crowdsourcing nor research with human subjects.
        \item Depending on the country in which research is conducted, IRB approval (or equivalent) may be required for any human subjects research. If you obtained IRB approval, you should clearly state this in the paper. 
        \item We recognize that the procedures for this may vary significantly between institutions and locations, and we expect authors to adhere to the NeurIPS Code of Ethics and the guidelines for their institution. 
        \item For initial submissions, do not include any information that would break anonymity (if applicable), such as the institution conducting the review.
    \end{itemize}

\item {\bf Declaration of LLM usage}
    \item[] Question: Does the paper describe the usage of LLMs if it is an important, original, or non-standard component of the core methods in this research? Note that if the LLM is used only for writing, editing, or formatting purposes and does \emph{not} impact the core methodology, scientific rigor, or originality of the research, declaration is not required.
    %this research? 
    \item[] Answer: \answerNA{} % Replace by \answerYes{}, \answerNo{}, or \answerNA{}.
    \item[] Justification: LLMs are not used as an important, original, or non-standard component of the core method, experiments, or scientific contributions of this work.
    \item[] Guidelines:
    \begin{itemize}
        \item The answer \answerNA{} means that the core method development in this research does not involve LLMs as any important, original, or non-standard components.
        \item Please refer to our LLM policy in the NeurIPS handbook for what should or should not be described.
    \end{itemize}

\end{enumerate}

\end{document}